\documentclass{article}
\PassOptionsToPackage{authoryear,square}{natbib}
\usepackage{amsmath,amssymb,amsthm,mathtools}
\usepackage[utf8]{inputenc}
\usepackage[T1]{fontenc}
\usepackage[final]{microtype}
\usepackage{graphicx,booktabs,array,multirow}
\usepackage{float}
\usepackage{placeins}
\usepackage{needspace}
\graphicspath{{figures/}}

\usepackage{xcolor}
\definecolor{MITOrange}{HTML}{C65D26}
\usepackage{url,bm,pifont,alphalph}
\usepackage{algorithmicx}
\usepackage{algpseudocode}
\usepackage{caption}
\usepackage[compact]{titlesec}
\makeatletter
\newcommand{\appendixsectionnumbering}{\renewcommand{\thesection}{\AlphAlph{\value{section}}}}
\makeatother
\allowdisplaybreaks

\floatstyle{ruled}
\newfloat{algorithm}{tbp}{loa}
\floatname{algorithm}{Algorithm}
\newtheorem{lemma}{Lemma}
\newtheorem{proposition}{Proposition}
\newtheorem{corollary}{Corollary}
\newtheorem{example}{Example}
\newtheorem{remark}{Remark}
\newtheorem{assumption}{Assumption}
\newcommand{\KL}{\mathrm{KL}}
\newcommand{\Ber}{\mathrm{Ber}}
\newcommand{\Z}{\mathbb{Z}}
\newcommand{\Var}{\mathrm{Var}}
\newcommand{\Cov}{\mathrm{Cov}}
\usepackage{enumitem}
\usepackage[hyperfootnotes=false,colorlinks=true,linkcolor=MITOrange,citecolor=MITOrange,urlcolor=MITOrange]{hyperref}
\usepackage[final,main]{neurips_2026}
\renewcommand{\acksection}{\section*{Acknowledges}}
\hypersetup{
  colorlinks=true,
  linkcolor=MITOrange,
  citecolor=MITOrange,
  urlcolor=MITOrange,
  pdftitle={Boundary Mass in Feature Space as a Label-Budget Diagnostic for Representations},
  pdfauthor={Siming Zhang, Zhehui Shen, Shijie Chen, Yansen Yu, Hang Yu}
}
\setcitestyle{authoryear,square,citesep={,},aysep={,}}
\newtheorem{theorem}{Theorem}
\newtheorem{definition}{Definition}
\newcommand{\R}{\mathbb{R}}
\newcommand{\E}{\mathbb{E}}
\newcommand{\Prob}{\mathbb{P}}
\newcommand{\1}{\mathbf{1}}
\newcommand{\Hd}{\mathcal H^{d-1}}
\newcommand{\mup}{\mu^+}
\newcommand{\psieff}{\psi_{\mathrm{eff}}}
\newcommand{\Dker}{D_{\mathrm{ker}}}
\newcommand{\Dhard}{D_{\mathrm{hard}}}
\newcommand{\psiker}{\widehat\psi_{\mathrm{eff}}^{\mathrm{ker}}}
\newcommand{\gauss}{\varphi}
\newcommand{\upgood}{\ensuremath{\uparrow}}
\newcommand{\downgood}{\ensuremath{\downarrow}}
\newcommand{\psimethod}[1]{\begingroup\boldmath\textbf{#1}\endgroup}
\titlespacing*{\section}{0pt}{1.0ex plus .2ex minus .2ex}{0.55ex}
\titlespacing*{\subsection}{0pt}{0.75ex plus .2ex minus .2ex}{0.35ex}
\titlespacing*{\subsubsection}{0pt}{0.55ex plus .15ex minus .15ex}{0.25ex}
\titlespacing*{\paragraph}{0pt}{0.45ex plus .1ex minus .1ex}{0.55em}
\setlist[itemize]{leftmargin=*,topsep=0pt,itemsep=0pt,parsep=0pt,partopsep=0pt}
\setlist[enumerate]{leftmargin=*,topsep=0pt,itemsep=0pt,parsep=0pt,partopsep=0pt}
\title{Boundary Mass in Feature Space as a Label-Budget Diagnostic for Representations}
\author{%
  Siming Zhang\thanks{Equal contribution.}\quad
  Zhehui Shen\footnotemark[1]\quad
  Shijie Chen\quad
  Yansen Yu\quad
  Hang Yu\thanks{Corresponding author: \href{mailto:yuhang@shu.edu.cn}{\texttt{yuhang@shu.edu.cn}}.}\\
  Shanghai University
}
\begin{document}
% Use a uniform color for all author footnote markers.
\hypersetup{linkcolor=MITOrange}
\maketitle
\hypersetup{linkcolor=MITOrange}
\begin{abstract}
Low-label transfer is often decided by a practical question: after a representation looks useful, how many labels are still needed to learn the downstream boundary? Transferability and probe-quality scores characterize global alignment, likelihood, margins, and calibration. Boundary-near target mass complements these diagnostics by measuring the ambiguous regions where scarce labels are most costly. We propose \textbf{effective boundary complexity}, a label-budget diagnostic for frozen representations and linear probes. It measures mass in a thin strip around the learned boundary and normalizes it by class balance, so dense ambiguous regions and predicted-class imbalance are both counted. We prove that, under a smooth representation map, density change and tangential surface change cancel at first order, leaving boundary-normal stretch or compression as the governing factor. This yields a cached-feature estimator that fits a probe on one split and counts independent held-out features near its boundary. Experiments support the diagnostic at three levels. Controlled transformations match the predicted law across 74 maps with $0.38\%$ median relative error. On real frozen features, normal-direction changes move both measured boundary mass and label demand, while tangent controls preserve the first-order coefficient. A differentiable surrogate reduces measured complexity by $16$--$36\%$ at matched top-1 utility. Finally, CIFAR-100 and Tiny-ImageNet experiments support boundary mass as a second-stage diagnostic for near-tied representations. Five-way experiments show strong OvR--interface backbone-ranking agreement (Spearman $0.872$), while acquisition experiments show high correlations between initial boundary mass and realized label budgets across three sampling policies ($0.779$--$0.830$). Code is available at \url{https://github.com/qqaqq-code/boundary_mass}.
\end{abstract}
\section{Introduction}\label{sec:introduction}
A powerful feature extractor can make a few-shot task look easier than it really is. With suitable pretrained embeddings, even simple classifiers can achieve competitive performance in few-shot recognition~\citep{chen2019closer}. However, support--query distribution alignment also plays an important role in few-shot generalization~\citep{tian2023aligning}, motivating a closer look at how target examples are arranged around the learned decision boundary.

Most existing diagnostics answer a different question. Transferability scores such as NCE, LEEP, H-score, LogME, TransRate, and PACTran estimate whether a source representation is globally aligned with a target task~\citep{tran2019transferability,nguyen2020leep,bao2019hscore,you2021logme,huang2022transrate,ding2022pactran}. Probe diagnostics such as likelihood, calibration, margins, and effective rank summarize how well the fitted head behaves~\citep{guo2017calibration}. These metrics identify useful representations and shortlist plausible candidates. Boundary-near target probability adds a local geometric measurement after the representation and probe are fixed. This complementary signal is particularly useful when labels are scarce and candidates receive similar global scores.

We study this local burden through effective boundary complexity, which measures boundary-near feature mass relative to predicted-class balance. For a binary classifier $f=\1_A$ in ambient dimension $d$, let $A=\{z\in\R^d:f(z)=1\}$ have boundary $\partial A$ and complement $A^c$. Let $P$ have density $p$, and let $\Hd$ denote $(d-1)$-dimensional Hausdorff measure. We define
\begin{equation}\label{eq:def}
\psieff(f,P):=\frac{\mup_P(f)}{\min\{P(A),P(A^c)\}},\qquad
\mup_P(f):=2\int_{\partial A}p(x)\,d\Hd(x).
\end{equation}
In our empirical evaluations, $f$ is the downstream linear probe fitted under the specified training protocol. Its empirical estimate is denoted $\hat\psieff$. Effective boundary complexity measures mass around this realized fitted probe under the induced feature distribution. The numerator is the first-order tube-probability coefficient; the denominator is the smaller predicted-class probability. A larger $\psieff$ indicates more boundary-near examples relative to that class, connecting local ambiguity to downstream label demand.

\begin{figure}[H]
\centering
\includegraphics[width=1\linewidth]{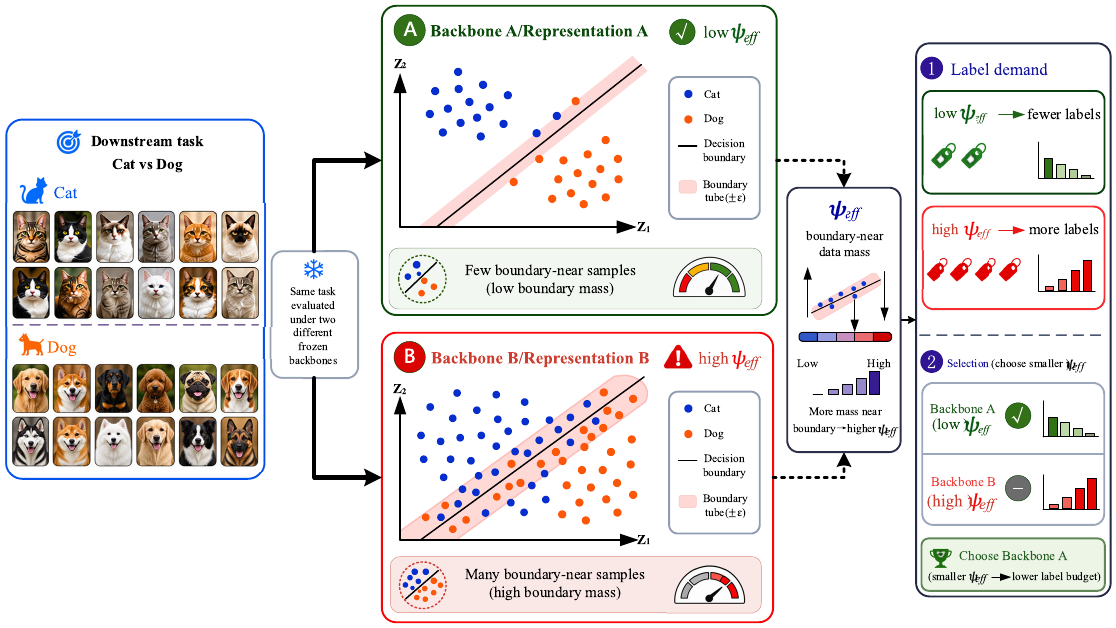}
\caption{Operational label-budget evaluation for frozen representations. For a fixed downstream task, different feature maps place different probability mass inside the learned boundary tube. Lower effective boundary complexity means fewer boundary-near examples for the low-label learner to resolve.}
\label{fig:operational}
\end{figure}
\vspace{-0.75em}

Given a fitted linear probe, we measure boundary-near target probability by counting independent diagnostic features inside a narrow tube and normalizing by predicted-class balance. Section~\ref{sec:frozen_estimator} formalizes this procedure through signed distances. A smooth surrogate supports interventions that reduce boundary mass while maintaining downstream utility.

Our pushforward law explains how representations change this quantity: density and tangential surface-area changes cancel at first order, leaving boundary-normal stretch or compression. Normal expansion thins the boundary tube; tangent-only changes preserve the first-order mass. Thus boundary mass measures the local ambiguity remaining after a representation induces a fitted downstream separator, complementing global transferability scores.

\Needspace{4\baselineskip}
\textbf{Contributions.}
\vspace{-0.25em}
\begin{itemize}[leftmargin=*,itemsep=0.35em,topsep=0.25em,parsep=0pt,partopsep=0pt]
\item \textbf{A local boundary diagnostic for label demand:} we introduce effective boundary complexity, a class-balanced measure of feature mass near a fitted downstream boundary, and prove that its first-order change under representation maps is governed by boundary-normal geometry.
\item \textbf{A practical low-label evaluation:} the theory yields a cached-feature estimator and differentiable surrogate that measure the remaining label burden after a probe is fitted, using held-out training features and empirical tube counts.
\item \textbf{Empirical evidence:} Tests on 74 controlled maps validate the pushforward law, normal/tangent interventions confirm its geometric mechanism, and surrogate optimization reduces measured complexity by $16$--$36\%$ at matched utility. Backbone, multiclass, and acquisition experiments further support near-tie selection, OvR--interface ranking agreement, and consistent boundary-mass--budget associations across sampling policies.
\end{itemize}

\section{Effective Boundary Complexity}
\begin{definition}[Effective boundary complexity]
Let $P$ have a density $p$ continuous near the $C^1$ boundary of $f=\1_A$, with finite density-weighted boundary integral and $0<P(A)<1$. The effective boundary complexity is Eq.~\eqref{eq:def}.
\end{definition}
We use boundary mass to refer broadly to probability concentrated near the decision boundary, while effective boundary complexity denotes its class-balanced first-order coefficient, $\psieff$.

The numerator is the first-order probability content of a thin boundary tube; the denominator expresses the same boundary mass relative to the smaller class. Thus the quantity is large when the decision surface passes through dense feature regions, when the harder side has little class mass, or both.

This interpretation is what makes the quantity useful for representation comparison. Density weighting assigns small contributions to complex boundary regions containing sparse feature mass. Conversely, a visually simple hyperplane can still be statistically costly if many target examples lie close to it. The density factor distinguishes these cases. The balance normalization serves a different purpose: it prevents a minority class from looking easy merely because its total probability is small. In binary and one-vs-rest transfer tasks, these two effects are exactly the ones a small-label learner experiences as boundary ambiguity.

\begin{theorem}[Boundary-tube constant]\label{thm:tube}
For $X\sim P$, if $\partial A$ is compact $C^2$ and $p$ is $C^1$ near $\partial A$, then, as the tube half-width $t\to0^+$,
\[
P\{\operatorname{dist}(X,\partial A)\le t\}=\mup_P(f)t+O(t^2).
\]
\end{theorem}
The theorem gives the local reading of $\mup_P(f)$: it is the coefficient of boundary-near probability mass under the actual feature distribution. In boundary-estimation language, $\mup_P(f)$ is the representation-dependent multiplier attached to the boundary tube. The controlled interventions and downstream learning curves evaluate the associated relationship between this geometric coefficient and label demand.

The factor 2 comes from the two-sided boundary tube. In signed-distance coordinates, the leading term is $2t\int_{\partial A}p\,d\Hd$; curvature and density variation enter through the $O(t^2)$ remainder under the stated regularity.

For a Gaussian halfspace $h_{w,b}(x)=\1\{w^\top x+b\ge0\}$, $w\in\R^d\setminus\{0\}$ is the head weight and $b\in\R$ its intercept. With $X\sim\mathcal N(0,I_d)$, where $I_d$ is the identity matrix, $\psieff(h_{w,b},P)=2\gauss(b/\|w\|)/\Phi(-|b|/\|w\|)$; $\gauss$ and $\Phi$ are the standard Gaussian density and cumulative distribution function. The minimum occurs at a balanced cut. Direct integration of the one-dimensional Gaussian marginal establishes the tube expansion for this noncompact boundary. For multiclass heads, our primary operational diagnostic sums one-vs-rest (OvR) coefficients, matching the classwise surrogate used for measurement and intervention. Appendix~\ref{app:multi} presents the interface-counted argmax-geometry estimator and evaluates its agreement with the OvR construction in multiclass backbone ranking.

\section{Related Work}
\paragraph{Transfer diagnostics.}
Our setting follows frozen-feature transfer: a pretrained representation is fixed and a simple target classifier is fit from limited labels. Transferability scores such as negative conditional entropy (NCE), Log Expected Empirical Prediction (LEEP), H-score, Log Maximum Evidence (LogME), TransRate, and PAC-Bayesian transferability (PACTran) rank candidate representations by global feature-label alignment~\citep{tran2019transferability,nguyen2020leep,bao2019hscore,you2021logme,huang2022transrate,ding2022pactran}. Fitted-head diagnostics such as likelihood, calibration, margins, and effective rank summarize confidence, separation, or feature spread~\citep{vapnik1998statistical,shalev2014understanding,neyshabur2015norm,bartlett2017spectrally,golowich2018size,dziugaite2017computing,guo2017calibration}. These quantities identify plausible transfer candidates. Boundary mass complements them by measuring target probability near the learned separator.

\paragraph{Boundary theory.}
The closest mathematical neighbors are boundary-content and boundary-rate quantities. Gaussian surface area, noise sensitivity, and robustness analyses connect boundary geometry to perturbations under a reference measure~\citep{borell1975brunn,sudakov1978extremal,ledoux2001concentration,talagrand1995concentration,gilmer2018adversarial,madry2018towards,bubeck2020law,bubeck2021universal}. Boundary-rate results in classification similarly express excess risk through probability mass in a tube around the decision boundary~\citep{tsybakov2009introduction,wainwright2019high,shamir2018distribution,fefferman2016testing}. Our quantity uses the empirical feature distribution induced by the representation and normalizes by class balance, so it measures the boundary burden seen by the learned head under the downstream protocol.

\paragraph{Representation geometry.}
Prior work on manifold learning, optimal transport, and expressivity studies how feature maps reshape data and decision regions~\citep{roweis2000nonlinear,tenenbaum2000global,peyre2019computational,cohen2016expressive,telgarsky2016benefits,raghu2017expressive}. In cross-domain few-shot segmentation, task-specific refinement of pretrained feature spaces has also been explored to mitigate representation mismatch, using lightweight feature enrichment and metric-based prediction~\citep{tian2025specificity}. Our pushforward law isolates the first-order boundary-normal factor: density and tangential-area changes cancel, leaving the inverse-transpose Jacobian applied to the boundary normal. Measuring held-out mass in the fitted boundary tube complements transferability, calibration, margin, and perturbation-stability diagnostics.

\section{Representation Pushforward of Boundary Mass}
Why does a representation change how many examples lie near a decision boundary? The crucial effect is how it reshapes space across the boundary, rather than along it. For a smooth invertible map, density and tangential-area changes cancel, leaving a boundary-normal factor that controls the first-order tube mass. We derive this law and show how to estimate boundary complexity using only a fitted linear probe and held-out cached features.

\subsection{Pushforward law}
Let $T$ be a $C^1$ diffeomorphism, let $Q=T_\#P$ be the feature distribution obtained by applying $T$ to samples from $P$, and let $g=f\circ T^{-1}=\1_{T(A)}$ be the pushed classifier. Write $DT(x)$ for the Jacobian and $n(x)$ for the outward unit normal to $\partial A$.
\begin{theorem}[Pushforward law]\label{thm:push}
If $A$ has $C^1$ boundary, $P$ has a continuous density $p$ near that boundary, and the displayed weighted surface integral is finite, then
\begin{equation}\label{eq:push}
\mup_Q(g)=2\int_{\partial A}p(x)\,\|DT(x)^{-\top}n(x)\|\,d\Hd(x).
\end{equation}
\end{theorem}
The area-formula calculation gives an operational transformation law. On $T(\partial A)$, the pushed density contributes $q(Tx)=p(x)/|\det DT(x)|$. The hypersurface area element contributes $|\det DT(x)|\|DT(x)^{-\top}n(x)\|\,d\Hd(x)$. The determinant cancels, so the only first-order representation factor visible to boundary mass is the inverse-transpose Jacobian applied to the boundary normal.

Tangential stretching preserves first-order boundary mass after density cancellation. Boundary-normal stretching changes $\psieff$: expanding the normal coordinate reduces the tube coefficient, while contracting it increases the coefficient. This predicts the normal-vs-tangent experiment: changing features along the learned boundary normal should move boundary mass and label demand, while tangent-only changes should have little first-order effect.

\begin{figure}[tbp]
\centering
\includegraphics[width=1\linewidth]{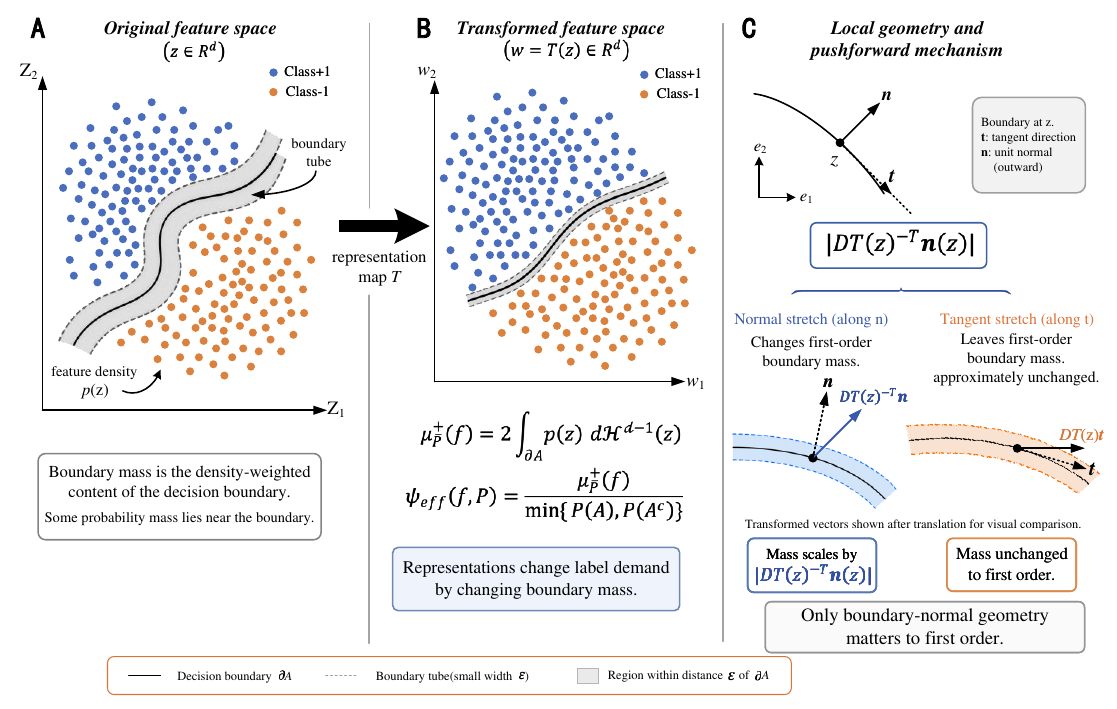}
\caption{Boundary mass and the pushforward mechanism. A boundary tube measures probability mass near the learned decision boundary. Under a smooth feature map $T$, the pushed density and tangential surface Jacobian cancel, leaving the inverse-transpose normal factor $\|DT(x)^{-\top}n(x)\|$ in the transformed boundary mass. Boundary-normal changes affect the first-order tube coefficient; tangent-only changes preserve it.}
\label{fig:mechanism}
\end{figure}

\paragraph{General smooth-boundary interpretation.}
The pushforward law covers general smooth decision surfaces, including nonlinear boundaries. At each boundary point $x$, the local unit normal $n(x)$ identifies the perpendicular direction, and $\|DT(x)^{-\top}n(x)\|$ determines the first-order change in boundary mass. A pure normal stretch by $\alpha$ gives $\|DT(x)^{-\top}n(x)\|=1/\alpha$. This local geometry describes how transformations move samples toward or away from the decision surface. Figure~\ref{fig:mechanism} visualizes the density--surface cancellation and normal factor.

For invertible representations the class-balance denominator is unchanged: $Q(T(A))=P(A)$ and $Q(T(A)^c)=P(A^c)$. Hence the entire change in $\psieff$ comes from the numerator in Eq.~\eqref{eq:push}. This gives an exact prediction for the controlled scaling maps used later. For a normal scale $s>0$ and unit normal $u$ to a linear decision boundary, define
\[
T_s z=z+(s-1)\langle z,u\rangle u .
\]
If the boundary normal is $u$, then $DT_s^{-\top}u=u/s$, so $\psieff$ is multiplied by $1/s$. Doubling the normal coordinate halves the first-order boundary-tube coefficient. For the same scaling applied in a unit tangent direction $v\perp u$, then $DT_s^{-\top}u=u$ and the first-order coefficient is unchanged. These factors are direct evaluations of Eq.~\eqref{eq:push}.

The law identifies a concrete intervention target: feature mass near the decision surface, governed by separation in the local normal direction. Tangential deformations and rotations describe complementary changes in feature geometry. This general smooth-boundary characterization connects representation transformations to boundary mass. The frozen-feature linear-head setting developed next provides a direct computational realization through sample-to-boundary distances.

\subsection{Frozen-feature estimator}\label{sec:frozen_estimator}
\paragraph{Operational procedure.}
The estimator follows three steps: (1) fit feature standardization statistics and a linear probe on the probe-training split; (2) compute signed distances on an independent diagnostic-heldout split; (3) aggregate tube membership and predicted-class balance. For a fixed backbone $\phi$, let $z_i=\phi(x_i)\in\R^d$, $i=1,\ldots,N$, be the $N$ diagnostic features. The fitted head $h_{w,b}$ gives signed distances, positive on its predicted class 1,
\[
s_i=\frac{w^\top z_i+b}{\|w\|}.
\]
For the fitted head $h_{w,b}$ and induced feature distribution $Q$, the population target is $G(w,b;Q)=\psieff(h_{w,b},Q)$ (Proposition~\ref{prop:bridge}). The head remains fixed during diagnostic evaluation, reducing tube membership to a scalar signed-distance calculation.

\paragraph{Hard tube estimator.}
For tube width $\varepsilon_0>0$, empirical tube mass and predicted-class balance are
\[
\widehat B_{\varepsilon_0}
=\frac1N\sum_{i=1}^N\1\{|s_i|\le\varepsilon_0\},
\qquad
\widehat u=\frac1N\sum_{i=1}^N\1\{s_i\ge0\},
\qquad
\widehat\beta=\min\{\widehat u,1-\widehat u\}.
\]
For $\widehat\beta>0$, the finite-tube estimator is
\begin{equation}\label{eq:hardtube}
\widehat{\psi}_{\mathrm{eff},\varepsilon_0}^{\mathrm{hard}}
=\frac{\widehat B_{\varepsilon_0}}{\varepsilon_0\,\widehat\beta}.
\end{equation}
Theorem~\ref{thm:tube} connects the surface integral to tube probability. The Gaussian-kernel estimator $\psiker=2N^{-1}\sum_i\varphi(s_i/\varepsilon_0)/(\varepsilon_0\widehat\beta)$ estimates twice the signed-distance density at zero, normalized by predicted-class balance. Its population counterpart converges to $\psieff$ as bandwidth shrinks under a continuous signed-distance density.

\paragraph{Smooth approximation.}
For differentiable interventions, we replace indicators and the minimum by smooth weights. With sigmoid $\sigma(t)=1/(1+e^{-t})$, tube temperature $\tau>0$, and balance smoothing $\rho>0$, define
\begin{equation}
\hat\psi_{\tau,\rho}:=\frac{N^{-1}\sum_i\sigma((\varepsilon_0-|s_i|)/\tau)}{\varepsilon_0\,m_\rho(N^{-1}\sum_i\sigma(s_i/\tau))},\quad
m_\rho(u)=\frac{1-\sqrt{(2u-1)^2+\rho^2}+\rho}{2}.
\end{equation}
For $u\in[0,1]$, $m_\rho(u)$ smooths class balance; the numerator softens tube membership. At fixed width, the hard limit recovers Eq.~\eqref{eq:hardtube} for sample distances off the indicator thresholds. Increasing sample size in the Hard estimate recovers the population finite-tube quantity. A continuous signed-distance marginal identifies its shrinking-width limit with $\psieff$ (Appendix~\ref{app:surrogate}).

\paragraph{Sampling and approximation.}
At fixed hyperparameters and positive class balance, bounded empirical weights yield $O_{\mathbb P}(N^{-1/2})$ sampling error. Finite-width and smoothing effects are separate approximation terms (Appendix~\ref{app:compute}); smaller widths reduce tube bias while increasing variance as fewer samples enter the tube. The fixed-grid evaluation uses $(\varepsilon_0,\tau,\rho)=(0.1,0.05,10^{-6})$. Backbone-selection measurements use $\varepsilon_0=\max(0.05,q_{0.2}(|s_i|))$, where $q_a$ is the empirical quantile at probability $a$. Appendix~\ref{app:score_comparison} reports kernel and tube estimates for label-budget prediction. Appendix~\ref{app:surrogate-short} gives consistency and smoothness details.

\paragraph{Multiclass aggregation.}
A $K$-class head sums $K$ classwise tube coefficients through one-vs-rest (OvR) aggregation, matching the independently parameterized classwise surrogate. The interface-counted estimator measures active boundaries of the argmax partition across $O(K^2)$ candidate pairs. Appendix~\ref{app:multi} compares Hard and Smooth versions of both constructions for backbone ranking.

\paragraph{Learner geometry.}
The default metric is standardized Euclidean distance, matching probe training. For a positive-definite learner metric $M$, the signed distance becomes
\[
s_M(z)=\frac{w^\top z+b}{\sqrt{w^\top M^{-1}w}}.
\]
Raw, normalized, and whitened variants in Appendix~\ref{app:expdetails} correspond to distinct learner geometries.

\paragraph{Smooth nonlinear heads.}
A regular smooth nonlinear surface also has a local unit normal and a boundary tube. Closest-point computations supply distances for the same tube-membership and class-balance calculation, providing a computational route to smooth nonlinear separators.

\paragraph{Computational complexity.}
Given cached features $Z\in\R^{N\times d}$ and a fitted head, binary distances cost $O(Nd)$, followed by scalar weighting and normalization. For $K$-class OvR, let $W\in\R^{K\times d}$ stack the head weights as rows. Products $ZW^\top$ cost $O(NKd)$; cached logits and head norms reduce additional diagnostic work to $O(NK)$. Batchwise accumulation supports memory-efficient evaluation; Appendix~\ref{app:compute} gives the streaming analysis.

All diagnostics are fixed before test labels evaluate downstream learning curves, preserving prospective selection. During end-to-end training, the representation distribution, feature density, and boundary evolve jointly; evaluation during training tracks this evolution (Appendix~\ref{app:ext_surrogate}), extending boundary-normal analysis to adaptive representations.

\section{Experiments}
We begin with synthetic experiments to validate the pushforward law in Eq.~\eqref{eq:push}. We then examine normal and tangential scaling on real frozen features and test whether surrogate-based interventions can reduce boundary mass while preserving accuracy. Finally, we evaluate the usefulness of boundary mass for backbone selection and label-budget prediction, including multiclass and label-acquisition settings.

\subsection{Experimental setting}
We use four evaluation regimes. The synthetic pushforward study uses known two-dimensional densities and 74 controlled representation transformations. The controlled real-feature evaluation uses cached CIFAR-100 features~\citep{krizhevsky2009learning} extracted from ResNet-50~\citep{he2016resnet} and ViT-B/16~\citep{dosovitskiy2021vit}, covering 8 binary class pairs per backbone and normal/tangent scaling interventions on the full-data linear probe direction. The intervention evaluation applies the smooth surrogate on frozen ImageNet-pretrained features~\citep{deng2009imagenet,russakovsky2015imagenet} and a top-2 boundary-pressure proxy in ResNet-18 end-to-end checks. The backbone-selection evaluation uses 40 CIFAR-100 binary class pairs and four frozen backbones: ResNet-50~\citep{he2016resnet}, ViT-B/16~\citep{dosovitskiy2021vit}, DINO-ViT-B/16~\citep{caron2021emerging}, and SigLIP-B/16~\citep{zhai2023sigmoid}. Appendix Table~\ref{tab:assets_app} lists the datasets, model families, and their original references. Tiny-ImageNet~\citep{le2015tiny} supplies binary replication and a 5-way aggregation evaluation under the same cached-feature protocol. We additionally evaluate multiclass estimator agreement on 40 five-way Tiny-ImageNet tasks, using four frozen backbones and five random splits. Acquisition-policy sensitivity is evaluated on 40 CIFAR-100 binary tasks across four frozen backbones, initializing Random, Uncertainty, and Hybrid acquisition from the same class-balanced two-samples-per-class (2-spc) labeled set.

For the main backbone-selection evaluation, standardization and one linear probe per task--backbone pair are fitted on the probe-training split; diagnostics use an independent diagnostic-heldout split. The binary scaling, backbone-selection, acquisition, and pilot-refitting experiments report the Gaussian-kernel score $\Dker=N^{-1}\sum_i\varphi(s_i/\varepsilon_0)/(\varepsilon_0\widehat\alpha_y)$, where $y_i\in\{0,1\}$ are diagnostic labels and $\widehat\alpha_y=\min\{N^{-1}\sum_i\1\{y_i=0\},N^{-1}\sum_i\1\{y_i=1\}\}$ is observed diagnostic-label balance. Its conversion is $\psiker=(2\widehat\alpha_y/\widehat\beta)\Dker$. These class-balanced tasks have $\widehat\alpha_y=1/2$. Appendix~\ref{app:score_comparison} reports label-budget prediction with kernel and hard scores under both balance normalizations. Comparison diagnostics include margins, negative log likelihood (NLL), expected calibration error (ECE)~\citep{guo2017calibration}, rank, and transferability. Acquisition and pilot experiments follow their initialization protocols; Appendix~\ref{app:pilot_refitting} reports refitting stability. Test labels evaluate downstream learning curves.

For the main backbone-selection experiments, the label budget $B_{p,b}$ is the interpolated number of samples per class (spc) required for backbone $b$ to reach 90\% of the best full-shot test accuracy for class pair $p$. For task set $\mathcal P$ and a policy $\pi$ selecting backbone $\pi(p)$, regret is
\[
\mathrm{Regret}(\pi):=\frac1{|\mathcal P|}\sum_{p\in\mathcal P}\bigl(B_{p,\pi(p)}-\min_b B_{p,b}\bigr),
\]
reported in spc; relative regret divides each difference by $\min_bB_{p,b}$. Area under the learning curve (AULC) uses the same label grid. Binary regression entries ``+$\Dker$'' add $\log(1+\Dker)$; $R^2$ measures explained variance and $\Delta R^2$ its increase. $p$-values use class-pair permutation tests. Selection policies use leave-one-class-pair-out prediction and report regret, AULC, and paired budget reductions. Appendix~\ref{app:expdetails} gives the protocols and supplementary results.

\subsection{Results}
\subsubsection{Finite-sample validation of the pushforward law}
This experiment directly tests the pushforward identity in Eq.~\eqref{eq:push}. We use synthetic distributions where the density, boundary, and representation map are known. The transformations include rotations, isotropic and anisotropic scalings, shears, diagonal stretching, and piecewise-affine maps. For each transformation we compute the predicted boundary coefficient from the inverse-transpose normal factor and compare it with a measured finite-tube estimate.

\begin{figure}[!htbp]
\centering
\begin{minipage}{.49\linewidth}
\includegraphics[width=\linewidth]{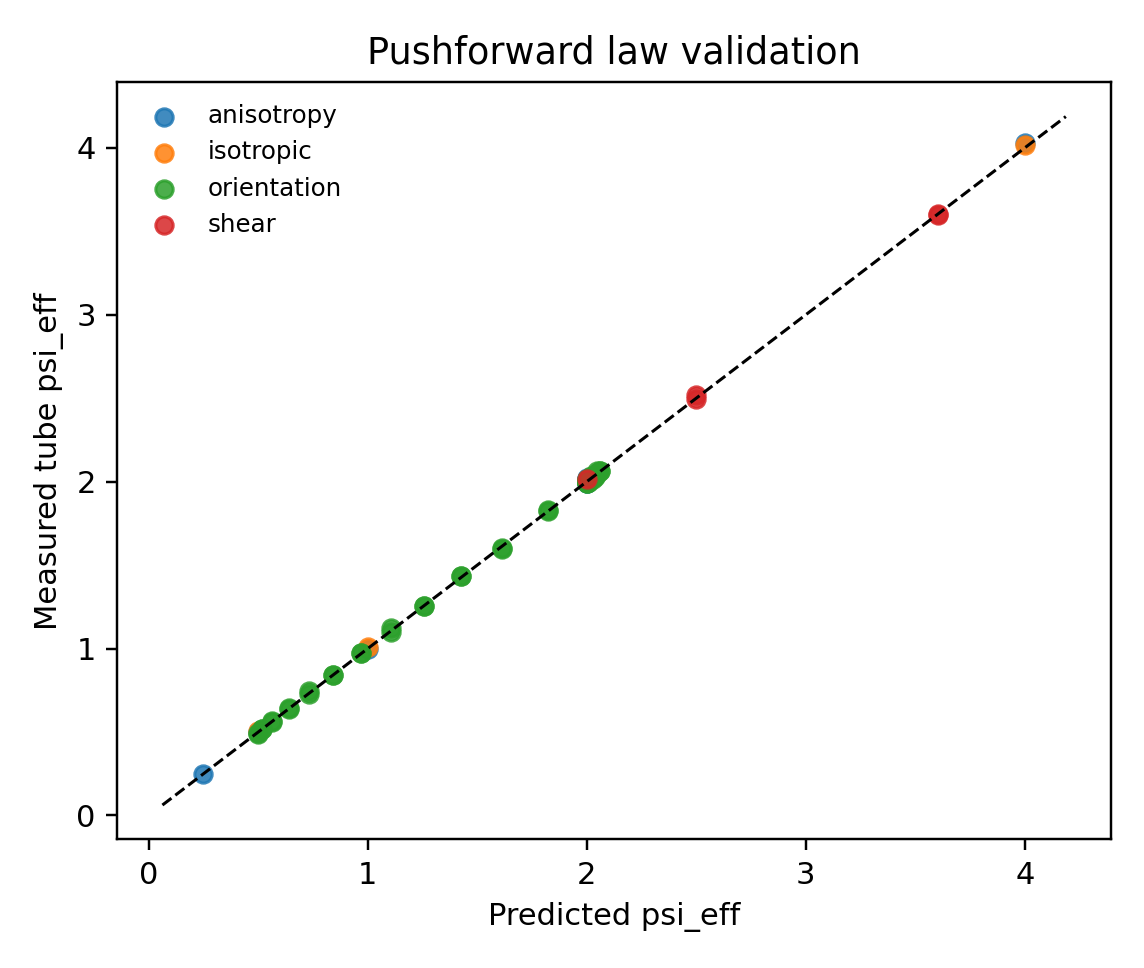}
\end{minipage}\hfill
\begin{minipage}{.49\linewidth}
\includegraphics[width=\linewidth]{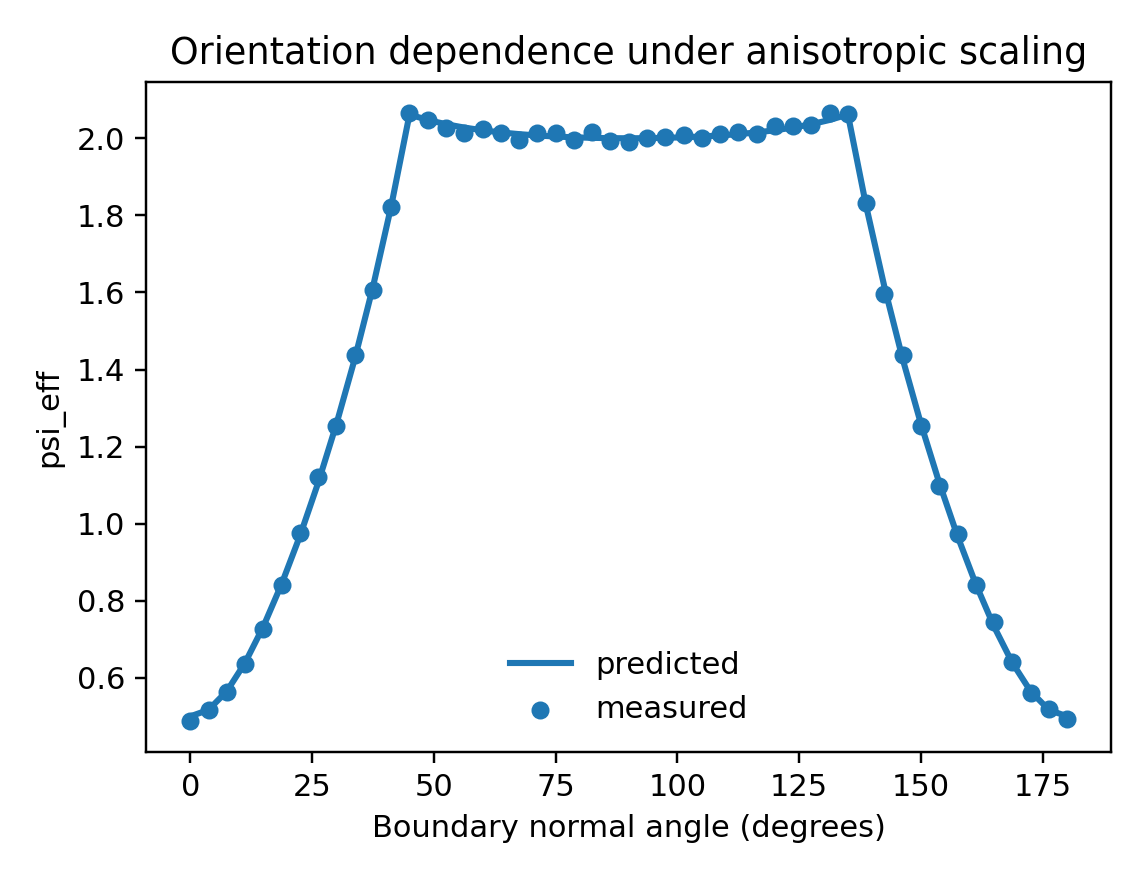}
\end{minipage}
\caption{Finite-sample validation of the pushforward law. Predicted and measured boundary mass agree across controlled transformations; the orientation sweep shows the boundary-normal factor $\lVert DT^{-\top}n\rVert$ directly. Across 74 transformations, median relative error is $0.38\%$ and mean relative error is $0.50\%$.}
\label{fig:pushforward_main}
\end{figure}

The measured finite-tube estimates closely track the predicted coefficients across the tested transformations. The orientation curve is the main stress test: the same anisotropic map produces different boundary mass depending on the boundary normal, exactly as the inverse-transpose term predicts. Appendix Figure~\ref{fig:pushforward_validation} reports the full four-panel sweep, including anisotropy and tube-width convergence.
The same constant also behaves like a sample-efficiency coefficient in the controlled plug-in study: fitting log error against raw sample size gives $R^2=0.698$, while using $n/\psieff$, with $n$ training samples, raises the fit to $R^2=0.895$ with log-log slope $-0.503$ (magnitude $0.503$), close to the two-dimensional boundary-rate exponent $1/2$.

\subsubsection{Controlled normal-direction changes in cached features}\label{sec:normal_tangent}
The second test moves from synthetic maps to real feature distributions. We use cached CIFAR-100 ResNet-50 and ViT-B/16 features, fit the full-data linear normal $u$ for each binary task, and apply
\[
T_s z=z+(s-1)\langle z,u\rangle u,\qquad s\in\{1/2,1,2,4\}.
\]
For each transformed feature set, we fit full-data and low-label linear heads. The full-data head defines $\Dker$, evaluated at the bandwidth chosen for $s=1$; low-label heads give the learning curves. Each budget uses three label-draw seeds, shared between directions at each scale. Images, labels, and downstream splits are fixed. Appendix~\ref{app:scale_conventions} gives the bandwidth relation under normal scaling.

\begin{figure}[!htbp]
\centering
\begin{minipage}{.5\linewidth}
\includegraphics[width=\linewidth]{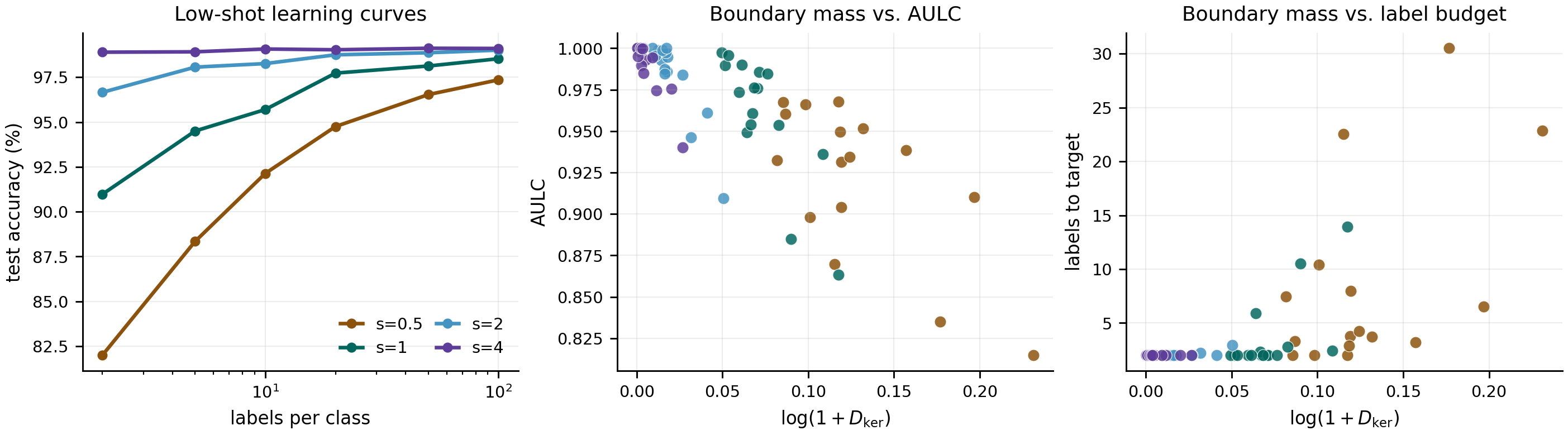}
\end{minipage}\hfill
\begin{minipage}{.5\linewidth}
\includegraphics[width=\linewidth]{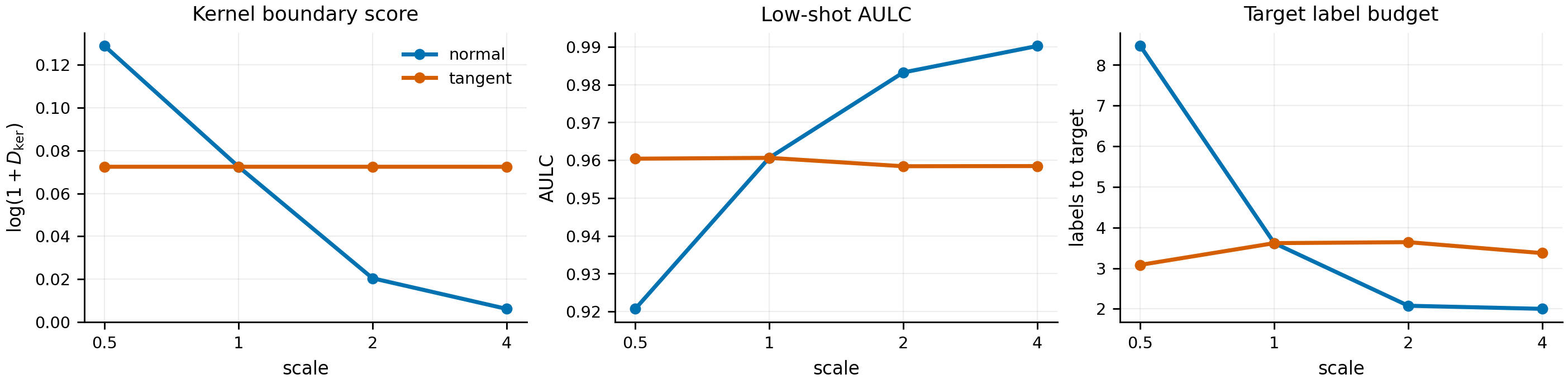}
\end{minipage}
\caption{Normal and tangent scaling of frozen features. Full-data heads provide the fixed-bandwidth kernel score $\Dker$; low-label heads provide learning curves and budgets. From $s=1$ to $s=2$, normal expansion lowers mean budget by $1.54$ spc, while the tangent control changes it by $+0.02$ spc.}
\label{fig:controlled_real}
\end{figure}

\begin{table}[!hbp]
\centering\scriptsize
\setlength{\tabcolsep}{2.5pt}
\caption{Normal and tangent scaling on 8 tasks and two backbones, with three label-draw seeds per budget. Left: mean normal-scaling score, accuracy at 5 spc, AULC on a 0--100 scale, and budget in spc. Right: paired changes from $s=1$ to $s=2$. Kernel scores use full-data heads and the $s=1$ bandwidth.}
\label{tab:controlled_normal}
\begin{minipage}[t]{.48\linewidth}\centering
\begin{tabular}{ccccc}\toprule
$s$ & $\log(1+\Dker)$\downgood & 5-spc acc.\upgood & AULC\upgood & budget\downgood\\
\midrule
0.5 & $0.129$ & $88.34$ & $92.072$ & $8.47$\\
1.0 & $0.072$ & $94.50$ & $96.066$ & $3.62$\\
2.0 & $0.020$ & $98.06$ & $98.323$ & $2.08$\\
4.0 & $0.006$ & $98.92$ & $99.023$ & $2.00$\\
\bottomrule\end{tabular}\end{minipage}\hfill
\begin{minipage}[t]{.48\linewidth}\centering
\begin{tabular}{lccc}\toprule
Dir. & $\Delta\log(1+\Dker)$\downgood & $\Delta$ AULC\upgood & $\Delta$ budget\downgood\\
\midrule
normal & $-0.052$ & $+2.26$ & $-1.54$\\
tangent & $+0.000$ & $-0.22$ & $+0.02$\\
\bottomrule\end{tabular}\end{minipage}
\end{table}

\paragraph{Controlled boundary-normal separation.}
Normal expansion increases separation along the original fitted normal. From $s=1$ to $s=2$, mean budget decreases by $1.54$ spc and AULC rises by $2.26$ points; tangent changes are $+0.02$ spc and $-0.22$ points (Table~\ref{tab:controlled_normal}, Figure~\ref{fig:controlled_real}). Across the 64 normal-scaling conditions, task--backbone-demeaned Pearson/Spearman correlations of $\log(1+\Dker)$ are $-0.889/-0.890$ with AULC and $0.659/0.670$ with budget. Normal separation therefore accompanies lower boundary-near mass and greater label efficiency in these experiments.

\subsubsection{Smooth interventions move the measured state}
We next test whether boundary mass can be reduced by optimization. On frozen ImageNet-pretrained features, we apply the differentiable surrogate to the linear head while monitoring held-out top-1 and $\hat\psieff$. The intervention penalizes soft mass near the fitted boundary.

\begin{table}[tbp]
\centering\scriptsize
\setlength{\tabcolsep}{3.5pt}
\caption{Smooth surrogate intervention. Frozen-feature rows compare standard linear-head training with a $\hat\psi$ intervention at the same samples per class; their low-label utility is the Top-1 column, so the probe column is reserved for the end-to-end rows. End-to-end rows use the top-2 boundary-pressure proxy and report mean$\pm$standard deviation over three CIFAR-100 ResNet-18 training seeds, each evaluated with three downstream probe seeds. Header arrows mark preferred directions.}
\label{tab:controllability_main}
\begin{tabular}{llccc}
\toprule
Setting & Method & Top-1\upgood & $\hat\psi$\downgood & Low-label probe\upgood\\
\midrule
CIFAR-100 10 spc & standard / $\hat\psi$-int. & $53.97/53.84$ & $201.68/169.22$ & --\\
CIFAR-100 50 spc & standard / $\hat\psi$-int. & $65.82/65.74$ & $246.23/185.57$ & --\\
Tiny-ImageNet 10 spc & standard / $\hat\psi$-int. & $62.85/62.73$ & $507.04/387.09$ & --\\
Tiny-ImageNet 50 spc & standard / $\hat\psi$-int. & $72.15/71.98$ & $659.58/420.08$ & --\\
\midrule
CIFAR-100 R18 & standard & $53.68\pm0.25$ & $23.57\pm0.33$ & $44.46/47.47$\\
CIFAR-100 R18 & weight decay & $54.55\pm0.20$ & $23.61\pm0.23$ & $45.53/48.43$\\
CIFAR-100 R18 & margin reg. & $54.55\pm0.18$ & $23.82\pm0.52$ & $44.96/47.96$\\
CIFAR-100 R18 & entropy reg. & $53.98\pm0.13$ & $24.01\pm0.41$ & $44.49/47.43$\\
CIFAR-100 R18 & \psimethod{boundary pressure} & \psimethod{$57.09\pm0.42$} & \psimethod{$22.07\pm0.35$} & \psimethod{$48.26/51.07$}\\
\bottomrule
\end{tabular}
\end{table}

Table~\ref{tab:controllability_main} shows the controllability pattern. In the frozen-feature intervention, top-1 changes by at most $0.17$ percentage points while measured boundary complexity drops by $16.1$--$36.3\%$. In the ResNet-18 end-to-end check, a top-2 boundary-pressure proxy lowers the held-out hard/kernel boundary estimates by $6.4/7.8\%$ relative to standard training and improves test accuracy and 20/50-spc frozen probes by $3.41$, $3.80$, and $3.60$ points. The boundary-mass reduction exceeds those of weight decay, margin regularization, and entropy regularization. Appendix~\ref{app:ext_endtoend} gives the lambda sweep, CIFAR-10 repeat, and anti-gaming controls.

\subsubsection{Second-stage selection in real backbone experiments}\label{sec:second_stage}
We evaluate 40 CIFAR-100 binary class pairs over ResNet-50, ViT-B/16, DINO-ViT-B/16, and SigLIP-B/16. The target is the 90\%-full-shot label budget in samples per class. Diagnostics use held-out training features; test labels evaluate the downstream learning curves.

\begin{figure}[!b]
\centering
\includegraphics[width=.96\linewidth]{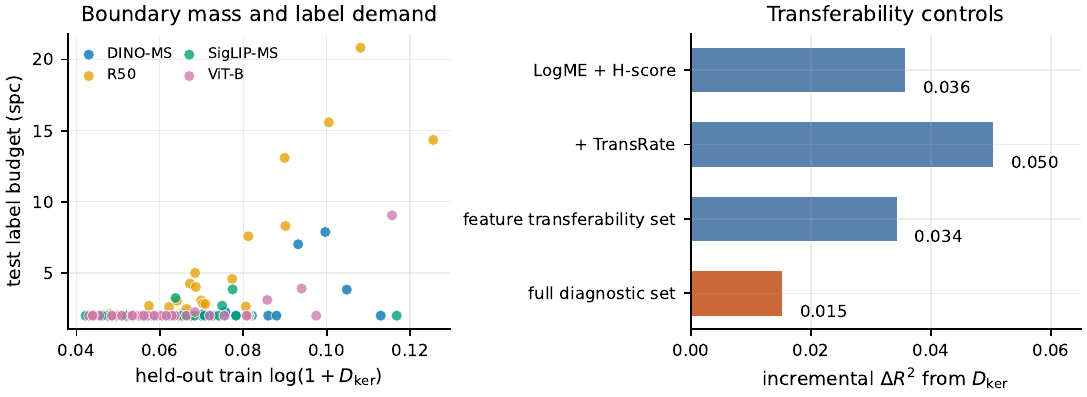}
\caption{Backbone-selection evaluation. Left: held-out boundary mass and realized label budget. Right: incremental $\Delta R^2$ from boundary mass after transferability controls. The full diagnostic set improves from $R^2=0.734$ to $0.749$, supporting complementary second-stage selection among near-tied candidates.}
\label{fig:modelselect_main}
\end{figure}

Figure~\ref{fig:modelselect_main} reports label demand and incremental predictive gains after transferability controls. Adding boundary mass yields $\Delta R^2=0.036$ for LogME + H-score, $0.050$ after adding TransRate, and $0.034$ for the feature-transferability set. With all conventional and transferability diagnostics, $R^2$ rises from $0.734$ to $0.749$: $\Delta R^2=0.015$, paired sign-flip $p=0.039$ (Table~\ref{tab:cifar_transfer_regression_app}). Using the reported rounded values, this explains about $5.6\%$ of the remaining variance, $0.015/(1-0.734)$. Boundary mass contributes local geometric information alongside accuracy, transferability, margin, and calibration.

The full baseline and boundary-augmented regression policies share a rounded mean regret of $0.06$ spc (Table~\ref{tab:transfer_policy_app}). The second-stage rule uses leave-one-class-pair-out prediction from conventional diagnostics to shortlist two backbones, then selects the lower $\log(1+\Dker)$, reducing regret from $0.058$ to $0.030$ spc. Appendix Figure~\ref{fig:neartie_app} supports retaining the conventional choice as diagnostic gaps widen; pairwise near ties yield a $0.02[0.00,0.06]$-spc reduction. CIFAR-100 and Tiny-ImageNet budget-floor effects concentrate gains on remaining budget differences (Appendix~\ref{app:external_audits}).

The retrospective matched-utility evaluation contains 176 and 227 of the 240 within-task backbone pairs at full-shot accuracy gaps of at most $1\%$ and $2\%$, respectively. Lower-boundary-mass choices achieve win/tie scores of $54.3\%$ and $54.6\%$, assigning scores of $1$, $0.5$, and $0$ to lower, equal, and higher label budgets (Table~\ref{tab:matchedutility_app}). These counts characterize near-tie prevalence; prospective shortlists use conventional diagnostic scores.

\subsubsection{Multiclass fidelity and acquisition-policy robustness}\label{sec:additional_audits}
\paragraph{Multiclass estimator agreement.}
On 40 five-way Tiny-ImageNet tasks, four frozen backbones, and five random splits, Hard OvR and Hard Interface achieve mean within-task Spearman correlation $0.872$, pairwise concordance $90.0\%$, and selected-backbone agreement $89.5\%$ (Table~\ref{tab:multiclass_agreement_main}). The Smooth comparison yields $0.868$, $89.7\%$, and $89.5\%$, respectively. Smooth estimators also closely match their Hard counterparts, with Spearman correlations of $0.988$--$0.992$ and concordance of $99.0$--$99.3\%$. These results support OvR as an efficient estimator that closely preserves interface-based backbone rankings. Appendix~\ref{app:multi} gives the complete comparison.

\begin{table}[!htbp]
\centering\small
\caption{Multiclass estimator agreement on 40 five-way Tiny-ImageNet tasks, four frozen backbones, and five random splits with shared semantic class identities across backbones. The metrics quantify agreement in backbone rankings.}
\label{tab:multiclass_agreement_main}
\begin{tabular}{lccc}
\toprule
Comparison & Spearman\upgood & Concordance\upgood & Backbone agreement\upgood\\
\midrule
Hard OvR vs.\ Hard Interface & $0.872$ & $90.0\%$ & $89.5\%$\\
Smooth OvR vs.\ Smooth Interface & $0.868$ & $89.7\%$ & $89.5\%$\\
\bottomrule
\end{tabular}
\end{table}

\paragraph{Sensitivity to label-acquisition policies.}
On 40 CIFAR-100 binary tasks and four frozen backbones, we fit the initial probe using the shared class-balanced 2-spc labeled set and compute Gaussian-kernel $\Dker$ on diagnostic-heldout features. Subsequent acquisition uses class-balanced Random sampling, Uncertainty sampling from the current probe, or an equal mixture (Hybrid). Mean realized budgets are $7.05$, $5.57$, and $5.47$ spc, while the corresponding Spearman correlations with initial kernel $\Dker$ are $0.779$, $0.830$, and $0.824$ (Table~\ref{tab:acquisition_main}). Selection regret is $1.08$, $0.58$, and $0.42$ spc, respectively. Thus acquisition changes the absolute budget while preserving a strong label-demand ranking signal. Appendix~\ref{app:acquisition} provides the protocol and full results.

\begin{table}[!htbp]
\centering\small
\caption{Acquisition-policy sensitivity on 40 CIFAR-100 binary tasks and four frozen backbones. Initial Gaussian-kernel $\Dker$ uses a probe fit on the shared class-balanced 2-spc labeled set and diagnostic-heldout features. Budgets and selection regrets are in spc.}
\label{tab:acquisition_main}
\begin{tabular}{lccc}
\toprule
Policy & Realized budget\downgood & Spearman\upgood & Selection regret\downgood\\
\midrule
Random & $7.05$ & $0.779$ & $1.08$\\
Uncertainty & $5.57$ & $0.830$ & $0.58$\\
Hybrid & $5.47$ & $0.824$ & $0.42$\\
\bottomrule
\end{tabular}
\end{table}

\section{Conclusion}
Effective boundary complexity measures representation-dependent label difficulty through boundary-near feature mass. Its pushforward law captures local normal geometry for general smooth decision surfaces; frozen representations and fitted linear heads provide efficient estimation from cached signed distances.

\vspace{0.7\baselineskip}
Synthetic maps validate the law, normal-versus-tangent interventions link boundary separation to label efficiency, and surrogate interventions reduce boundary mass at matched utility. Backbone experiments support complementary second-stage selection among near-tied representations. Multiclass experiments show strong OvR--interface ranking agreement, while acquisition experiments preserve strong associations with label demand across three sampling policies.

\newpage
\begin{ack}
This work was supported by the National Key R\&D Program of China (Grant No. 2025YFF0514701).
\end{ack}

\bibliographystyle{plainnat}
\bibliography{references}
\clearpage
\appendix
\appendixsectionnumbering
\clearpage
\thispagestyle{plain}
\begingroup
\hypersetup{linkcolor=MITOrange}
\begin{center}
{\Large\bfseries Appendix}
\end{center}
\vspace{0.9em}
\noindent{\Large\bfseries Contents}\par
\vspace{0.85em}
\begingroup
\normalsize
\newcommand{\appdotleader}{\leaders\hbox to .55em{\hss.\hss}\hfill}
\newcommand{\appsec}[3]{\noindent\makebox[0.055\linewidth][l]{\textbf{#1}}\makebox[0.875\linewidth][l]{\textbf{#2}\appdotleader}\makebox[0.06\linewidth][r]{#3}\par\vspace{0.38em}}
\newcommand{\appsubsec}[3]{\noindent\hspace{0.055\linewidth}\makebox[0.10\linewidth][l]{#1}\makebox[0.775\linewidth][l]{#2\appdotleader}\makebox[0.06\linewidth][r]{#3}\par\vspace{0.24em}}
\appsec{A}{Proof Sketches and Geometric Interpretation}{\pageref{app:proofs}}
\appsubsec{A.1}{Tube, pushforward, and characterization sketches}{\pageref{app:proofs}}
\appsubsec{A.2}{Sample-complexity reading}{\pageref{app:proofs}}
\appsec{B}{Surrogate Details}{\pageref{app:surrogate-short}}
\appsec{C}{Supplementary Experimental Tables and Evaluation Diagnostics}{\pageref{app:expdetails}}
\appsubsec{C.1}{Estimator reliability, complexity, and robustness}{\pageref{app:estimator_audits}}
\appsubsec{C.2}{Backbone-selection and transferability experiments}{\pageref{app:selection_audits}}
\appsubsec{C.3}{External validation and acquisition-policy experiments}{\pageref{app:external_audits}}
\appsubsec{C.4}{Objective-separation and foundation-backbone experiments}{\pageref{app:control_audits}}
\appsec{D}{Multiclass OvR--Interface Estimator Comparison}{\pageref{app:multi}}
\appsec{E}{Supplementary Technical Appendix}{\pageref{app:extended_tech}}
\appsubsec{\ref{app:ext_background}}{Background and notation}{\pageref{app:ext_background}}
\appsubsec{\ref{app:ext_boundary}}{Boundary-complexity foundations}{\pageref{app:ext_boundary}}
\appsubsec{\ref{app:ext_pushforward}}{Pushforward geometry and extensions}{\pageref{app:ext_pushforward}}
\appsubsec{\ref{app:ext_surrogate}}{Surrogate analysis and optimization notes}{\pageref{app:ext_surrogate}}
\appsubsec{\ref{app:ext_experiments}}{Supplementary experiments}{\pageref{app:ext_experiments}}
\appsubsec{\ref{app:margin}}{Complete proof collection}{\pageref{app:margin}}
\appsubsec{\ref{app:ext_endtoend}}{End-to-end controllability check}{\pageref{app:ext_endtoend}}
\endgroup
\vfill
\endgroup
\clearpage

\section{Proof Sketches and Geometric Interpretation}\label{app:proofs}
\paragraph{Tube constant.} The tubular-neighborhood theorem gives a signed-distance chart near $\partial A$. Integrating $p$ over the two sides of the tube yields $P(\operatorname{dist}(X,\partial A)\le t)=2t\int_{\partial A}p\,d\Hd+O(t^2)$ when $\partial A$ is compact $C^2$ and $p$ is $C^1$ near the boundary. For locally bounded density, weighted Minkowski content provides the first-order interpretation whenever the tube-content limit exists.
\paragraph{Pushforward.} Apply the area formula to $T(\partial A)$. The pushed density contributes $|\det DT|^{-1}$ and the tangential Jacobian contributes $|\det DT|\|DT^{-\top}n\|$, leaving Eq.~\eqref{eq:push}. Piecewise-affine and ReLU extensions hold cellwise away from fold sets under the stated rank and boundary-injectivity conditions.
\paragraph{Supplementary characterization.} Locality and density linearity give a boundary integral with kernel $\kappa$, namely $\int_{\partial A}\kappa(x,n(x),p(x))\,d\Hd$. Isotropy removes dependence on the orientation $n$; density linearity gives $cp(x)$ for a constant $c$; the balanced Gaussian halfspace normalization fixes $c=2$ within this restricted functional class.
\paragraph{Sample-complexity reading.} In plug-in nearest-boundary rules, the error region is a tube around $\partial A$. Replacing the abstract tube constant by $\psieff$ gives the representation-dependent constant in the rate. This is why the synthetic collapse uses the normalized sample size $n/\psieff$: two representations with the same Bayes rule can require different label budgets because they present different boundary mass to the learner.

\section{Surrogate Details}\label{app:surrogate-short}
The estimator in the main text is smooth on every open parameter region where every diagnostic signed distance satisfies $s_i\ne0$; globally it is locally Lipschitz and differentiable almost everywhere because of the absolute value in $|s_i|$. A globally $C^\infty$ variant replaces $|s_i|$ by $(s_i^2+\nu^2)^{1/2}$ for fixed $\nu>0$. For multi-class one-vs-rest heads we sum the per-class surrogates. At fixed positive class balance, the approximation terms have separate origins: finite tube width $O(\varepsilon_0)$, sigmoid shell smoothing $O(\tau/\varepsilon_0)$, and class-balance smoothing $O(\rho)$. The implementation uses held-out training features and differentiates the sigmoid terms directly.
\paragraph{Notation guide.}
The main symbols used throughout the paper have fixed roles: $\mup_P(f)$ is the unnormalized two-sided tube coefficient; $\psieff(f,P)$ is the class-balanced population coefficient; $\widehat{\psi}_{\mathrm{eff},\varepsilon_0}^{\mathrm{hard}}$ is the finite hard-tube estimate; $\hat\psi_{\tau,\rho}$ is the differentiable surrogate with tube smoothing $\tau$ and balance smoothing $\rho$; $\widehat B_{\varepsilon_0,\tau}$ is the smoothed tube fraction, $\hat u_\tau$ is the smoothed positive-class fraction; and $s_i$ is the signed distance from diagnostic feature $z_i$ to the fitted probe boundary. We use $N$ diagnostic features and $n$ labeled training observations. Binary diagnostic tables use $\Dker$ for the Gaussian score with observed-label balance, $\Dhard$ for its hard counterpart, and $\psiker$ for the two-sided Gaussian estimate with predicted balance. Intervention and multiclass tables identify their respective readouts. Entries written as $a/b$ report Pearson/Spearman correlations.
\section{Supplementary Experimental Tables and Evaluation Diagnostics}\label{app:expdetails}
This appendix presents estimator reliability, computational cost, backbone selection, and label-acquisition results. Table~\ref{tab:senscost} reports surrogate-training costs, and Appendix~\ref{app:compute} describes cached-feature scoring. Correlation entries written as $a/b$ mean Pearson/Spearman, and header arrows mark the preferred direction.
\clearpage
\subsection{Estimator reliability, complexity, and robustness}\label{app:estimator_audits}
\begin{table}[h!]
\centering\small
\caption{Estimator reliability on the strict diagnostic-heldout split. Split-half entries summarize 100 random stratified halves. Hard-tube agreement compares $\log(1+\Dhard)$ and $\log(1+\Dker)$ with the same quantile width. Correlations are computed across 160 heads.}
\label{tab:estimatorreliability}
\begin{tabular}{lcccc}
\toprule
Check & heads & Pearson\upgood & Spearman\upgood & error / diff\downgood\\
\midrule
split-half $\log(1+\Dker)$ & 160 & $0.960$ & $0.970$ & $3.2\%$\\
hard-tube vs. kernel & 160 & $0.981$ & $0.982$ & $0.026$\\
\bottomrule
\end{tabular}
\end{table}

\begin{table}[h!]
\centering\small
\caption{Estimator sensitivity and cost. Rank stability is Spearman correlation of $\hat\psieff$ under a hyperparameter setting against the default $(\varepsilon_0,\tau,\rho)=(0.1,0.05,10^{-6})$. Cost is mean seconds per epoch from saved training logs.}
\label{tab:senscost}
\begin{tabular}{lccc}
\toprule
Setting & n & rank stability\upgood & median shift\downgood\\
\midrule
eps0.05/tau0.025/rho1e-06 & 54 & $0.993$ & $1.0\%$\\
eps0.05/tau0.05/rho1e-06 & 54 & $0.999$ & $23.6\%$\\
eps0.1/tau0.025/rho1e-06 & 54 & $0.998$ & $5.2\%$\\
eps0.1/tau0.05/rho1e-06 & 54 & $1.000$ & $0.0\%$\\
eps0.1/tau0.1/rho1e-06 & 54 & $0.993$ & $21.9\%$\\
eps0.2/tau0.05/rho1e-06 & 54 & $0.991$ & $5.9\%$\\
eps0.1/tau0.05/rho1e-05 & 54 & $1.000$ & $0.0\%$\\
\bottomrule
\end{tabular}\quad
\begin{tabular}{lccc}
\toprule
Benchmark & CE sec.\downgood & $\psi$ sec.\downgood & overhead\downgood\\
\midrule
CIFAR-100/R50 & $1.00$ & $1.39$ & $39.8\%$\\
CIFAR-100/ViT-B & $0.13$ & $0.13$ & $2.8\%$\\
Tiny-IN/R50 & $0.38$ & $0.37$ & $-0.6\%$\\
\bottomrule
\end{tabular}
\end{table}

\FloatBarrier

\paragraph{Computational complexity and scalability.}\label{app:compute}
Feature extraction and probe fitting are shared with frozen-feature evaluation. The marginal diagnostic cost starts with cached features and a fitted head; Table~\ref{tab:senscost} separately reports surrogate-regularization training costs.

Let $Z\in\R^{N\times d}$ be the diagnostic feature matrix. For a binary head $(w,b)$, the estimator computes
\[
s_i=\frac{w^\top z_i+b}{\|w\|},
\]
in $O(Nd)$ time, followed by scalar Smooth weights and balance normalization. For $K$-class OvR with $W\in\R^{K\times d}$, linear score terms $ZW^\top$ cost $O(NKd)$, head norms cost $O(Kd)$, and weighting and accumulation cost $O(NK)$. Cached logits and head norms reduce additional work to $O(NK)$.

For batch size $B$ and class block size $K_b$, streaming uses $O(Bd+BK_b+K)$ working memory beyond the fitted head. Per-class running sums accumulate tube weights and predicted-class masses in one feature-cache pass. Reading dense features costs $\Omega(Nd)$, so the binary estimator is input-linear. Linear scores, scalar weights, and reductions provide a lightweight complement to covariance- and spectrum-based diagnostics.

At fixed feature dimension and downstream head, pretrained model size affects shared feature extraction, while diagnostic cost depends on the cache. The computation also applies to pooled language-model embeddings or hidden states with a downstream linear classifier.

\paragraph{Diagnostic sample size and estimation error.}
Theorem~\ref{thm:tube} relates the population surface integral to first-order tube probability. Conditional on the fitted probe, independent diagnostic samples estimate tube probability and predicted-class mass by empirical averages. Their ratio in Eq.~\eqref{eq:hardtube} estimates the population finite-width quantity. A locally Lipschitz signed-distance density gives $O(\varepsilon_0)$ approximation bias at fixed positive class balance. Smooth weights yield the differentiable implementation.

At fixed tube width and smoothing parameters and positive smaller predicted-class mass $\beta$, standard concentration and the delta method give $O_{\mathbb P}(N^{-1/2})$ sampling error. Combining this with Proposition~\ref{prop:surrogate_bias} gives
\[
\hat\psi_{\tau,\rho}-\psieff
=O_{\mathbb P}(N^{-1/2})+O\!\left(\frac{\varepsilon_0}{\beta}
 +\frac{\tau}{\varepsilon_0\beta}
 +\frac{\tau+\rho}{\beta^2}\right).
\]
The terms describe sampling, width, sigmoid, and balance effects; sampling constants depend on fixed width and balance. Smaller widths reduce bias and increase variance, while larger samples improve precision. At fixed positive $\beta$, taking the sample-size limit and then shrinking width with $\tau/\varepsilon_0\to0$ and $\rho\to0$ recovers the population coefficient. Tables~\ref{tab:estimatorreliability}--\ref{tab:senscost} assess reliability, estimator agreement, and sensitivity.

\begin{table}[h!]
\centering\scriptsize
\caption{Tube-width robustness on the strict CIFAR-100 evaluation. Rank stability is Spearman correlation against the default standardized $q_{0.2}$ kernel estimate. Nearby quantile widths and the hard-tube estimator preserve rankings; a globally fixed width is less stable because signed-distance scales differ across heads.}
\label{tab:tubewidth_robustness}
\begin{tabular}{lccccc}
\toprule
Setting & heads & rank stability\upgood & mean regret\downgood & rel. regret\downgood & AULC\upgood\\
\midrule
fixed $\varepsilon_0=0.05$ & 160 & $0.671$ & $0.21$ & $10.7$ & $0.975$\\
$q_{0.1}$ tube & 160 & $0.939$ & $0.09$ & $4.4$ & $0.975$\\
\psimethod{$q_{0.2}$ tube} & \psimethod{160} & \psimethod{$1.000$} & \psimethod{$0.10$} & \psimethod{$5.0$} & \psimethod{$0.977$}\\
$q_{0.3}$ tube & 160 & $0.998$ & $0.08$ & $4.0$ & $0.978$\\
margin-scaled tube & 160 & $0.942$ & $0.02$ & $0.9$ & $0.980$\\
hard $q_{0.2}$ tube & 160 & $0.982$ & $0.11$ & $5.6$ & $0.978$\\
\bottomrule
\end{tabular}
\end{table}

\begin{table}[!htbp]
\centering\scriptsize
\caption{Feature-normalization evaluation on the strict CIFAR-100 backbone selection task. The diagnostic is stable under the paper's standardized protocol and nearby normalized variants, while raw and principal component analysis (PCA)-whitened protocols change the ranking substantially. This supports comparing absolute $\Dker$ values under a fixed protocol.}
\label{tab:normalization_robustness}
\begin{tabular}{lccccc}
\toprule
Protocol & heads & rank stability\upgood & mean regret\downgood & rel. regret\downgood & AULC\upgood\\
\midrule
raw features & 160 & $0.369$ & $0.33$ & $16.3$ & $0.973$\\
\psimethod{standardized} & \psimethod{160} & \psimethod{$1.000$} & \psimethod{$0.10$} & \psimethod{$5.0$} & \psimethod{$0.977$}\\
L2-normalized & 160 & $0.683$ & $0.05$ & $2.4$ & $0.979$\\
PCA-whitened & 160 & $0.232$ & $0.33$ & $16.6$ & $0.972$\\
\bottomrule
\end{tabular}
\end{table}

\begin{table}[!htbp]
\centering\small
\caption{Benchmarks included in the completed proxy evaluation. The evaluation covers both CIFAR-100 and Tiny-ImageNet cached-feature settings.}
\begin{tabular}{lccc}\toprule Benchmark & runs & classes & feature dim.\\\midrule
CIFAR-100/R50 & 106 & 100 & 2048\\
CIFAR-100/ViT-B & 27 & 100 & 768\\
Tiny-IN/R50 & 45 & 200 & 2048\\
\bottomrule\end{tabular}
\end{table}

\paragraph{Existing assets and sources.}
Table~\ref{tab:assets_app} lists the datasets and pretrained backbones used in the main experiments, their roles, and the corresponding references.
\begin{table}[!htbp]
\centering\small
\caption{Datasets and pretrained backbones used in the main experiments, with their roles and original references.}
\label{tab:assets_app}
\begin{tabular}{@{}p{.25\linewidth}p{.34\linewidth}p{.34\linewidth}@{}}
\toprule
Asset & Used for & Citation\\
\midrule
CIFAR-100 & binary class-pair experiments, interventions & \citep{krizhevsky2009learning}\\
Tiny-ImageNet & external replication & \citep{le2015tiny}\\
ImageNet / ILSVRC & pretraining source for frozen backbones & \citep{deng2009imagenet,russakovsky2015imagenet}\\
ResNet-50 / ResNet-18 & backbone / end-to-end evaluation & \citep{he2016resnet}\\
ViT-B/16 & frozen backbone & \citep{dosovitskiy2021vit}\\
DINO-ViT-B/16 & frozen backbone & \citep{caron2021emerging}\\
SigLIP-B/16 & frozen backbone & \citep{zhai2023sigmoid}\\
\bottomrule
\end{tabular}
\end{table}

\begin{table}[!htbp]
\centering\small
\caption{Boundary--Utility final distance to the empirical envelope. The intervention changes boundary geometry at comparable final utility, and this table visualizes the resulting diagnostic state. The expansion+compression column is the fraction of runs whose final point improves both boundary compression and utility expansion relative to the empirical envelope; all modes are zero under this strict joint criterion.}
\begin{tabular}{lcccc}\toprule Mode & runs & median dist.\downgood & mean dist.\downgood & expansion+compression\upgood\\\midrule
\psimethod{psieff} & \psimethod{118} & \psimethod{$0.417$} & \psimethod{$0.516$} & \psimethod{$0.0\%$}\\
standard & 54 & $0.951$ & $0.875$ & $0.0\%$\\
weight decay & 78 & $0.929$ & $0.855$ & $0.0\%$\\
\bottomrule\end{tabular}
\end{table}

\FloatBarrier
\paragraph{Kernel and tube estimates.}\label{app:score_comparison}
Gaussian weighting and tube counting estimate the signed-distance density at the learned boundary. Normalizing by the smaller observed class frequency $\widehat\alpha_y$ gives
\[
\Dker=\frac{N^{-1}\sum_i\varphi(s_i/\varepsilon_0)}{\varepsilon_0\widehat\alpha_y},
\qquad
\Dhard=\frac{\widehat B_{\varepsilon_0}}{2\varepsilon_0\widehat\alpha_y}.
\]
The binary experiments use $\Dker$ to measure boundary concentration relative to observed class balance. For the smaller predicted-class frequency $\widehat\beta$, the two-sided coefficients in Section~\ref{sec:frozen_estimator} are $\psiker=(2\widehat\alpha_y/\widehat\beta)\Dker$ and $\widehat\psi_{\mathrm{eff},\varepsilon_0}^{\mathrm{hard}}=(2\widehat\alpha_y/\widehat\beta)\Dhard$.

\paragraph{Label-budget prediction.}
Kernel and tube scores provide additional predictive information beyond conventional diagnostics ($\Delta R^2=0.012$--$0.023$) and correlate with acquisition budgets across three sampling policies (Spearman $0.779$--$0.872$; Table~\ref{tab:score_conventions}). We compute all four scores from the same fitted head and bandwidth for each task--backbone pair. Regressions add $\log(1+\mathrm{score})$ to conventional diagnostics, and the top-2 policy chooses the lower-scoring candidate from their shortlist. Acquisition correlations use mean log-scores and mean budgets over five seeds for each of 160 task--backbone pairs.

\begin{table}[!htbp]\centering\small
\caption{Kernel and tube estimates for label-budget prediction. Full-probe regressions use 160 task--backbone pairs and a conventional baseline with $R^2=0.734$. Top-2 regret is in spc. Acquisition columns report Spearman correlations between seed-averaged log-score and realized label budget.}
\label{tab:score_conventions}
\begin{tabular}{lccccc}\toprule
& \multicolumn{2}{c}{Full-probe evaluation} & \multicolumn{3}{c}{Acquisition correlation}\\
\cmidrule(lr){2-3}\cmidrule(lr){4-6}
Score & $\Delta R^2$ & Top-2 regret & Random & Uncertainty & Hybrid\\\midrule
$\Dker$ & $0.015$ & $0.030$ & $0.779$ & $0.830$ & $0.824$\\
$\psiker$ & $0.012$ & $0.030$ & $0.809$ & $0.855$ & $0.849$\\
$\Dhard$ & $0.023$ & $0.081$ & $0.814$ & $0.860$ & $0.855$\\
$\widehat\psi_{\mathrm{eff},\varepsilon_0}^{\mathrm{hard}}$ & $0.022$ & $0.034$ & $0.829$ & $0.872$ & $0.867$\\
\bottomrule\end{tabular}\end{table}

\paragraph{Bandwidth under normal scaling.}\label{app:scale_conventions}
For $T_s$, the transported head $h\circ T_s^{-1}$ has signed distances $s$ times the original distances and preserves class balance. Let $D_{\rm ker}^{\rm base}(\varepsilon)$ denote the original score at bandwidth $\varepsilon$, and $D_{\rm ker}^{\rm push}(s;\varepsilon)$ the transported score. Substitution in the kernel gives
\[
D_{\rm ker}^{\rm push}(s;\varepsilon)
=\frac1sD_{\rm ker}^{\rm base}(\varepsilon/s),\qquad
D_{\rm ker}^{\rm push}(s;s\varepsilon)
=\frac1sD_{\rm ker}^{\rm base}(\varepsilon).
\]
The identities also hold for hard scores and predicted-balance normalization. The first relates normal scaling to bandwidth; the second gives finite-width equivariance when distances and bandwidth scale together. The population coefficient follows $1/s$. Section~\ref{sec:normal_tangent} evaluates refitted heads at a common bandwidth, measuring boundary-near mass in fixed distance units. Its label-budget target is 90\% of the best full-data accuracy over the transformations within each task--backbone configuration.

\paragraph{Intervention readouts.}
Sigmoid tube weights define the differentiable intervention surrogate. End-to-end tables average classwise hard and triangular-kernel scores using observed class frequencies with floor $0.01$; these monitor training. Multiclass comparisons retain their classwise OvR and interface definitions.
\FloatBarrier

\subsection{Backbone-selection and transferability experiments}\label{app:selection_audits}
\begin{samepage}
\paragraph{Incremental predictive signal.}
Accuracy, transferability, margin, calibration, and boundary quality capture related aspects of strong representations, creating partial statistical overlap. In Table~\ref{tab:cifar_transfer_regression_app}, boundary mass raises the full baseline's $R^2$ from $0.734$ to $0.749$, with $\Delta R^2=0.015$, and paired sign-flip $p=0.039$. Using the displayed values, the increment explains $0.015/(1-0.734)\approx5.6\%$ of residual variance. This measures additional predictive information, while Tables~\ref{tab:transfer_policy_app} and~\ref{tab:matchedutility_app} separately evaluate selection outcomes. The top-2 rule in Section~\ref{sec:second_stage} targets candidates with closely matched conventional scores, and the normal-versus-tangent intervention in Section~\ref{sec:normal_tangent} connects the local boundary signal to a controlled geometric mechanism.
\end{samepage}

\begingroup
\setlength{\intextsep}{3pt plus 1pt minus 1pt}
\begin{figure}[H]
\centering
\includegraphics[width=.92\linewidth]{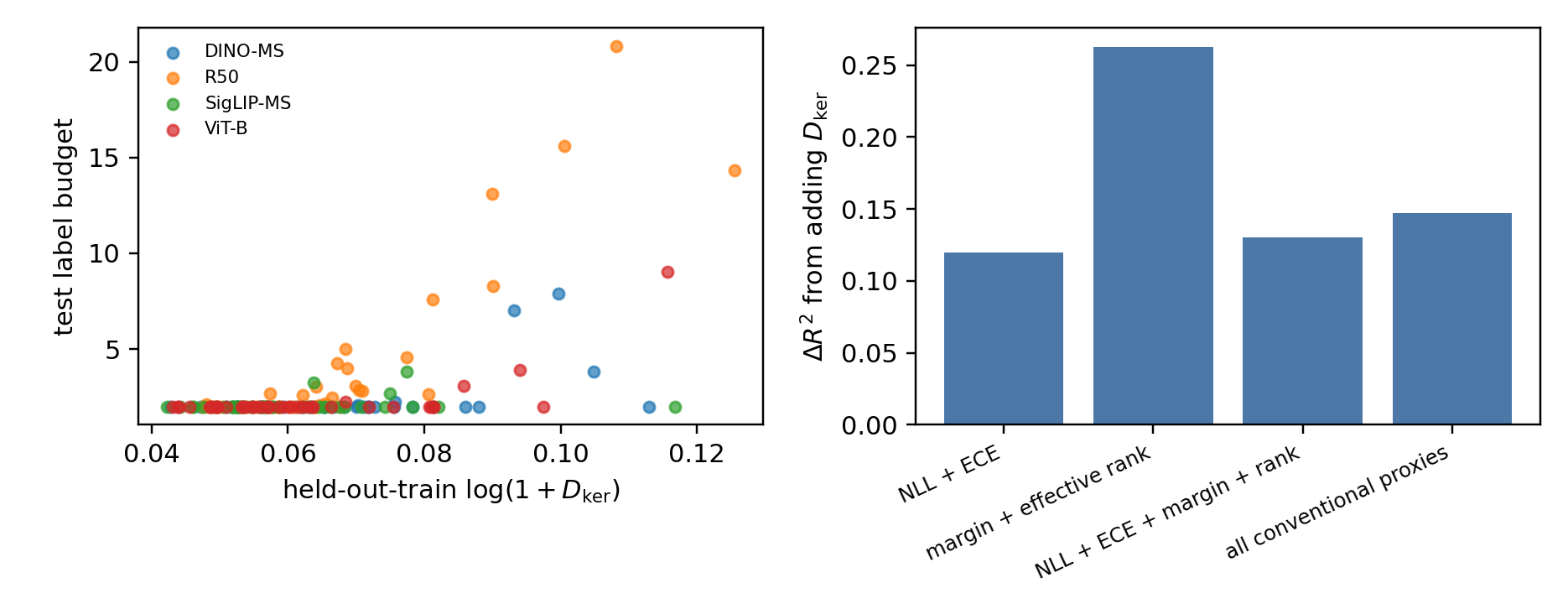}
\caption{Strict backbone-selection label-budget evaluation across 40 CIFAR-100 class pairs and four cached backbones, including DINO and SigLIP. Diagnostics are computed on held-out training features; test labels are reserved for the target learning curves, preserving the prospective selection protocol.}
\label{fig:modelselect_app}
\end{figure}
\begin{table}[!htbp]
\centering\scriptsize
\caption{Main CIFAR-100 four-backbone regression evaluation. $R^2$ predicts pair-demeaned label budget from the listed non-$\Dker$ covariates, then adds $\log(1+\Dker)$. $p$ is a paired sign-flip test for the incremental $\Delta R^2$.}
\label{tab:cifar_main_regression_app}
\setlength{\tabcolsep}{3pt}
\begin{tabular}{p{.42\linewidth}cccc}
\toprule
Baseline set & base $R^2$\upgood & +$\Dker$ $R^2$\upgood & $\Delta R^2$\upgood & $p$\downgood\\
\midrule
NLL + ECE & $0.343$ & $0.462$ & $0.120$ & $0.007$\\
margin + effective rank & $0.171$ & $0.434$ & $0.263$ & $<0.001$\\
NLL + ECE + margin + rank & $0.399$ & $0.530$ & $0.130$ & $<0.001$\\
all non-$\Dker$ proxies & $0.513$ & $0.660$ & $0.147$ & $<0.001$\\
\bottomrule
\end{tabular}
\end{table}

\begin{table}[!htbp]
\centering\scriptsize
\caption{CIFAR-100 four-backbone regression evaluation with transferability controls. Feature-transferability metrics are computed only on diagnostic-heldout training features.}
\label{tab:cifar_transfer_regression_app}
\setlength{\tabcolsep}{3pt}
\begin{tabular}{p{.36\linewidth}ccccc}
\toprule
Baseline set & base $R^2$\upgood & +$\Dker$ $R^2$\upgood & $\Delta R^2$\upgood & $p$\downgood & partial corr.\\
\midrule
LogME + H-score & $0.509$ & $0.545$ & $0.036$ & $0.027$ & $0.270/0.180$\\
LogME + H-score + TransRate & $0.538$ & $0.589$ & $0.050$ & $0.004$ & $0.330/0.258$\\
feature transferability set & $0.612$ & $0.646$ & $0.034$ & $0.009$ & $0.297/0.260$\\
all non-$\Dker$ + transferability & $0.734$ & $0.749$ & $0.015$ & $0.039$ & $0.239/0.233$\\
\bottomrule
\end{tabular}
\end{table}

\begin{table}[!htbp]
\centering\scriptsize
\caption{Strict single-metric backbone-selection results for label budget over ResNet-50, ViT-B/16, DINO, and SigLIP. Diagnostics are computed on held-out training features only.}
\setlength{\tabcolsep}{2.5pt}
\begin{tabular}{lcccc}
\toprule
Metric & label-budget corr. & selection acc.\upgood & mean regret\downgood & rel. regret\downgood\\
\midrule
\psimethod{$\log(1+\Dker)$} & \psimethod{0.493/0.266} & \psimethod{$25.0$} & \psimethod{$0.10$} & \psimethod{$5.0$}\\
\psimethod{$\Dker$} & \psimethod{0.499/0.266} & \psimethod{$25.0$} & \psimethod{$0.10$} & \psimethod{$5.0$}\\
Gaussian-area proxy & 0.240/0.036 & $30.0$ & $0.38$ & $19.0$\\
noise flip proxy & 0.207/0.157 & $50.0$ & $0.84$ & $42.1$\\
median margin & -0.293/-0.239 & $22.5$ & $0.13$ & $6.6$\\
held-out NLL & 0.525/0.404 & $12.5$ & $0.18$ & $8.9$\\
held-out ECE & 0.309/0.373 & $15.0$ & $0.24$ & $11.9$\\
held-out acc. & -0.532/-0.459 & $35.0$ & $0.18$ & $8.8$\\
effective rank & 0.402/0.412 & $47.5$ & $2.09$ & $104.4$\\
\bottomrule
\end{tabular}
\end{table}

\begin{table}[!htbp]
\centering\small
\caption{Feature-transferability single-metric correlations with pair-demeaned label budget under the strict diagnostic protocol. These metrics use diagnostic-heldout training features and labels. Higher LogME/H-score/TransRate/NCE-style values are conventionally better, so negative correlations mean lower label demand.}
\label{tab:transfer_single_app}
\begin{tabular}{lc}
\toprule
Metric & label-budget corr.\\
\midrule
LogME & $-0.313/-0.418$\\
H-score & $-0.636/-0.470$\\
TransRate & $0.217/0.284$\\
NCE-style cluster & $-0.645/-0.575$\\
\bottomrule
\end{tabular}
\end{table}

\begin{table}[!htbp]
\centering\scriptsize
\caption{PACTran-linear baseline on the strict CIFAR-100 backbone-selection evaluation. This is a PAC-Bayes-style regularized linear target-head score on diagnostic-heldout features, implemented as a practical PACTran proxy.}
\label{tab:pactran_linear_app}
\begin{tabular}{p{.35\linewidth}ccccc}
\toprule
Baseline set & base $R^2$\upgood & +$\Dker$ $R^2$\upgood & $\Delta R^2$\upgood & $p$\downgood & partial corr.\\
\midrule
PACTran-linear & $0.211$ & $0.281$ & $0.071$ & $0.110$ & $0.299/0.071$\\
feature transferability + PACTran-linear & $0.625$ & $0.648$ & $0.024$ & $0.059$ & $0.251/0.257$\\
all non-$\Dker$ + transferability + PACTran-linear & $0.745$ & $0.758$ & $0.014$ & $0.092$ & $0.232/0.207$\\
\bottomrule
\end{tabular}
\end{table}

\begin{table}[!htbp]
\centering\scriptsize
\caption{Source-logit transferability evaluation on the ResNet-50/ViT-B/16 subset. We downloaded/loaded the official torchvision ImageNet classifiers, forwarded CIFAR-100 training images, and computed LEEP, NCE, and source-logit LogME on diagnostic-heldout training examples. This source-head evaluation uses the two backbones with comparable 1000-way ImageNet classifiers.}
\label{tab:source_logit_app}
\setlength{\tabcolsep}{3pt}
\begin{tabular}{p{.31\linewidth}ccccc}
\toprule
Baseline set & base $R^2$\upgood & +$\Dker$ $R^2$\upgood & $\Delta R^2$\upgood & $p$\downgood & partial corr.\\
\midrule
LEEP + NCE & $0.199$ & $0.248$ & $0.049$ & $0.519$ & $0.247/0.234$\\
source-logit set & $0.260$ & $0.331$ & $0.071$ & $0.342$ & $0.310/0.284$\\
source + feature transferability & $0.653$ & $0.715$ & $0.062$ & $0.040$ & $0.423/0.435$\\
all non-$\Dker$ + source/feature & $0.863$ & $0.863$ & $0.000$ & $0.870$ & $0.036/0.047$\\
\bottomrule
\end{tabular}
\end{table}

\begin{table}[!htbp]
\centering\small
\caption{Single source-logit transferability correlations with pair-demeaned label budget on the ResNet-50/ViT-B/16 subset. Higher values are conventionally better, so negative correlations mean lower label demand.}
\label{tab:source_logit_single_app}
\begin{tabular}{lc}
\toprule
Metric & label-budget corr.\\
\midrule
LEEP & $-0.433/-0.686$\\
NCE & $-0.400/-0.571$\\
source-logit LogME & $-0.228/-0.528$\\
\bottomrule
\end{tabular}
\end{table}

\FloatBarrier
\endgroup
\subsection{External validation and acquisition-policy experiments}\label{app:external_audits}
\begin{table}[!htbp]
\centering\scriptsize
\caption{Independent Tiny-ImageNet label-budget evaluation over 40 binary class pairs and four frozen backbones. ViT-B/16, DINO, and SigLIP Tiny features were extracted for this evaluation; ResNet-50 uses the existing Tiny-ImageNet feature cache. Diagnostics are computed on diagnostic-heldout training features, and the incremental gains indicate that the binary external evaluation is close to saturation.}
\label{tab:tiny_independent_app}
\begin{tabular}{p{.30\linewidth}ccccc}
\toprule
Baseline set & base $R^2$\upgood & +$\Dker$ $R^2$\upgood & $\Delta R^2$\upgood & $p$\downgood & partial corr.\\
\midrule
NLL + ECE & $0.595$ & $0.657$ & $0.062$ & $0.072$ & $0.392/0.231$\\
margin + effective rank & $0.377$ & $0.649$ & $0.272$ & $<0.001$ & $0.661/0.758$\\
feature transferability & $0.636$ & $0.658$ & $0.023$ & $0.174$ & $0.249/0.336$\\
all non-$\Dker$ proxies & $0.688$ & $0.701$ & $0.013$ & $0.290$ & $0.203/0.357$\\
all non-$\Dker$ + transferability & $0.701$ & $0.709$ & $0.008$ & $0.388$ & $0.166/0.311$\\
\bottomrule
\end{tabular}
\end{table}

\begin{table}[!htbp]
\centering\scriptsize
\caption{Tiny-ImageNet single-metric backbone-selection summary. This independent evaluation provides replication evidence: $\Dker$ has low regret, while transferability metrics such as LogME are also strong.}
\label{tab:tiny_selection_app}
\begin{tabular}{lcccc}
\toprule
Metric & selection acc.\upgood & mean regret\downgood & rel. regret\downgood & mean AULC\upgood\\
\midrule
\psimethod{$\log(1+\Dker)$} & \psimethod{$77.5$} & \psimethod{$0.01$} & \psimethod{$0.5$} & \psimethod{$0.985$}\\
LogME & $87.5$ & $0.00$ & $0.0$ & $0.985$\\
H-score & $87.5$ & $0.61$ & $30.5$ & $0.982$\\
NLL & $37.5$ & $0.08$ & $3.8$ & $0.982$\\
ECE & $42.5$ & $0.33$ & $16.4$ & $0.980$\\
margin & $77.5$ & $0.01$ & $0.5$ & $0.985$\\
effective rank & $12.5$ & $18.80$ & $939.9$ & $0.878$\\
\bottomrule
\end{tabular}
\end{table}

\begin{table}[!htbp]
\centering\small
\caption{Tiny-ImageNet 5-way aggregation stress test. The evaluation compares the operational OvR score with an interface-counted argmax-geometry variant under the same cached-feature protocol. The 16-task evaluation reports backbone-selection regret on a label grid where most tasks reach the target at the smallest budget.}
\label{tab:multiclass_stress_app}
\resizebox{\linewidth}{!}{%
\begin{tabular}{lccccccc}
\toprule
Setting & tasks & backbones & base regret\downgood & $\psi$ regret\downgood & reduction\upgood & $p$\downgood & AULC\upgood\\
\midrule
Tiny-ImageNet 5-way OvR & 16 & 4 & $0.051$ & $0.100$ & $-0.049$ & $0.709$ & $0.961$\\
Tiny-ImageNet 5-way interface & 16 & 4 & $0.051$ & $0.450$ & $-0.398$ & $0.999$ & $0.947$\\
\bottomrule
\end{tabular}
}
\end{table}

\begin{table}[!htbp]
\centering\scriptsize
\caption{Tiny-ImageNet 5-way policy variants under saturation. Abstention and regression variants test whether $\psi$ should act only when non-boundary diagnostics are near-tied. The 16-task evaluation reports policy regret, budget reduction, AULC change, and win rate; the OvR abstention policy yields a positive budget reduction.}
\label{tab:multiclass_policy_supplement_app}
\begin{tabular}{lcccc}
\toprule
Policy & regret\downgood & reduction\upgood & $\Delta$AULC\upgood & win rate\upgood\\
\midrule
Non-$\psi$ base & $0.051$ & $0.000$ & $0.000$ & $0.00$\\
OvR top-2 & $0.100$ & $-0.049$ & $0.001$ & $0.12$\\
OvR abstain & $0.047$ & $0.004$ & $0.001$ & $0.12$\\
Interface top-2 & $0.450$ & $-0.398$ & $-0.013$ & $0.00$\\
Interface abstain & $0.145$ & $-0.093$ & $-0.004$ & $0.00$\\
Non-$\psi$ + OvR reg. & $0.051$ & $0.000$ & $0.000$ & $0.00$\\
Non-$\psi$ + interface reg. & $0.051$ & $0.000$ & $-0.000$ & $0.00$\\
\bottomrule
\end{tabular}
\end{table}

\FloatBarrier
\paragraph{Reading of the external experiments.}
The two Tiny-ImageNet binary tables are replication checks for the decision protocol. The binary evaluation is close to saturated: the best non-boundary selector already has near-zero regret, leaving little room for a second-stage boundary correction. The 5-way tables record the corresponding multi-class policy diagnostics under the same cached-feature protocol. In this evaluation, $87.5\%$ of tasks already attain the best budget at the smallest grid point, and each task has $1.69$ zero-regret backbones on average; the median-gap OvR variant gives a small positive reduction ($0.004$).
\paragraph{Acquisition-policy sensitivity.}\label{app:acquisition}
We evaluate 40 CIFAR-100 binary tasks over four frozen backbones. Each task starts with the same class-balanced 2-spc labeled set across acquisition policies. For each task--backbone pair, we fit the initial linear probe and compute Gaussian-kernel $\Dker$ on diagnostic-heldout features. Labels are then acquired through class-balanced Random sampling, Uncertainty sampling based on the current probe, or Hybrid sampling with an equal mixture of both. Across five seeds, the target is $95\%$ of each task's best full-probe accuracy. We record the realized budget required to reach this target, the Spearman correlation between initial kernel $\Dker$ and realized budget, and selection regret relative to the best backbone under the same acquisition policy. For the reported Spearman correlations, we average the five seeds within each task--backbone pair and correlate the resulting 160 diagnostic--budget pairs. This acquisition experiment evaluates label demand after a shared 2-spc initialization; the preceding tables evaluate the original backbone-selection policies.

\begin{table}[!htbp]
\centering\small
\caption{Label-acquisition policy sensitivity on 40 CIFAR-100 binary tasks and four frozen backbones. A shared class-balanced 2-spc labeled set initializes the probe for Smooth boundary-mass estimation, followed by Random, Uncertainty, or Hybrid acquisition. Budgets and selection regrets are in spc.}
\label{tab:acquisition_app}
\begin{tabular}{lccc}
\toprule
Label policy & Realized budget\downgood & \shortstack{Spearman\\(kernel $\Dker$, budget)\upgood} & Selection regret\downgood\\
\midrule
Random & $7.05$ & $0.779$ & $1.08$\\
Uncertainty & $5.57$ & $0.830$ & $0.58$\\
Hybrid & $5.47$ & $0.824$ & $0.42$\\
\bottomrule
\end{tabular}
\end{table}

The acquisition strategy changes the realized annotation budget while preserving a strong relationship between initial boundary mass and label demand. Mean budgets are $7.05$, $5.57$, and $5.47$ spc for Random, Uncertainty, and Hybrid acquisition, respectively. Spearman correlations remain high at $0.779$, $0.830$, and $0.824$, and selection regrets are $1.08$, $0.58$, and $0.42$ spc. These results support the stability of the predictive relationship captured by kernel $\Dker$ across the evaluated acquisition policies.

\noindent\textbf{Pilot-policy evaluation.} The following tables use only pilot labels and diagnostic-heldout features to select a backbone prospectively, before test labels are used.
\begin{table}[!htbp]
\centering\tiny
\renewcommand{\arraystretch}{0.82}
\caption{Prospective CIFAR-100 pilot-label policy simulation. Heads use 2, 5, or 10 labels per class; diagnostics are computed on held-out training features and policies are leave-one-class-pair-out budget regressions.}
\label{tab:pilot_policy_app}
\begin{tabular}{clcccc}
\toprule
pilot spc & policy & selection acc.\upgood & mean regret\downgood & rel. regret\downgood & mean AULC\upgood\\
\midrule
 \psimethod{2} & \psimethod{choose by $\Dker$} & \psimethod{$17.5$} & \psimethod{$0.05$} & \psimethod{$2.4$} & \psimethod{$0.979$}\\
2 & choose by margin & $15.0$ & $0.09$ & $4.7$ & $0.978$\\
2 & NLL/ECE & $22.5$ & $0.06$ & $3.2$ & $0.978$\\
2 & LogME/H-score & $52.5$ & $0.13$ & $6.3$ & $0.971$\\
2 & transferability set & $27.5$ & $0.05$ & $2.6$ & $0.977$\\
2 & combined non-$\Dker$ & $30.0$ & $0.03$ & $1.3$ & $0.975$\\
 \psimethod{2} & \psimethod{combined non-$\Dker$ + $\Dker$} & \psimethod{$27.5$} & \psimethod{$0.04$} & \psimethod{$1.8$} & \psimethod{$0.975$}\\
 \psimethod{5} & \psimethod{choose by $\Dker$} & \psimethod{$22.5$} & \psimethod{$0.05$} & \psimethod{$2.4$} & \psimethod{$0.979$}\\
5 & choose by margin & $22.5$ & $0.05$ & $2.4$ & $0.979$\\
5 & NLL/ECE & $17.5$ & $0.10$ & $5.2$ & $0.977$\\
5 & LogME/H-score & $52.5$ & $0.13$ & $6.3$ & $0.971$\\
5 & transferability set & $27.5$ & $0.05$ & $2.6$ & $0.977$\\
5 & combined non-$\Dker$ & $20.0$ & $0.06$ & $3.2$ & $0.979$\\
 \psimethod{5} & \psimethod{combined non-$\Dker$ + $\Dker$} & \psimethod{$20.0$} & \psimethod{$0.06$} & \psimethod{$3.2$} & \psimethod{$0.978$}\\
 \psimethod{10} & \psimethod{choose by $\Dker$} & \psimethod{$17.5$} & \psimethod{$0.10$} & \psimethod{$4.8$} & \psimethod{$0.979$}\\
10 & choose by margin & $22.5$ & $0.05$ & $2.4$ & $0.979$\\
10 & NLL/ECE & $27.5$ & $0.02$ & $0.9$ & $0.980$\\
10 & LogME/H-score & $52.5$ & $0.13$ & $6.3$ & $0.971$\\
10 & transferability set & $27.5$ & $0.05$ & $2.6$ & $0.977$\\
10 & combined non-$\Dker$ & $27.5$ & $0.00$ & $0.0$ & $0.978$\\
 \psimethod{10} & \psimethod{combined non-$\Dker$ + $\Dker$} & \psimethod{$22.5$} & \psimethod{$0.02$} & \psimethod{$0.9$} & \psimethod{$0.978$}\\
\bottomrule
\end{tabular}
\end{table}

\begin{table}[!htbp]
\centering\small
\caption{Paired effect of adding $\Dker$ to the combined non-$\Dker$ pilot policy. Positive reduction means the selected backbone has lower label budget.}
\label{tab:pilot_policy_savings_app}
\begin{tabular}{lccc}
\toprule
pilot spc & tasks with lower budget\upgood & mean budget reduction\upgood & mean relative reduction\upgood\\
\midrule
2 & $2.5$ & $-0.01$ & $-0.5$\\
5 & $0.0$ & $0.00$ & $0.0$\\
10 & $0.0$ & $-0.02$ & $-0.9$\\
\bottomrule
\end{tabular}
\end{table}

\paragraph{Pilot-policy reading.}
The pilot-label simulation is prospective: each policy sees the small pilot split and diagnostic-heldout features, then selects a backbone before test labels are used. Tables~\ref{tab:pilot_policy_app}--\ref{tab:pilot_policy_savings_app} show that LogME/H-score remain strong first-stage selectors and record when the boundary term changes the choice after pilot diagnostics are fixed.

\paragraph{Pilot-refitting stability.}\label{app:pilot_refitting}
We evaluate variation across independently sampled pilot-label sets on 40 CIFAR-100 binary tasks, four frozen backbones, and pilot sizes of 2, 5, and 10 labels per class. For each task--backbone--pilot-size configuration, we refit the linear head using five independent pilot draws while keeping the diagnostic pool fixed. The experiment comprises $40\times4\times3\times5=2{,}400$ linear-head refits and $40\times3\times5\times9=5{,}400$ seed-level policy choices. The nine policies are four single scores (boundary mass, margin, LogME, H-score) and five regression rules (NLL/ECE, LogME/H-score, the transferability set, the combined conventional set, and that combined set augmented with boundary mass).

\begin{table}[!htbp]
\centering\footnotesize
\caption{Pilot-refitting stability on 40 CIFAR-100 binary tasks and four backbones, with five pilot draws per configuration and a fixed diagnostic pool. Refit SD summarizes the across-draw standard deviation of $\log(1+\Dker)$. Exact agreement refers to the selected backbone under the lower-$\Dker$ and combined+$\Dker$ policies; budget agreement refers to the combined+$\Dker$ policy. Agreement columns are percentages.}
\label{tab:pilot_refitting_stability}
\setlength{\tabcolsep}{3pt}
\begin{tabular}{cccccc}
\toprule
\shortstack{Pilot\\spc} & \shortstack{Refit\\SD\downgood} & \shortstack{Backbone-order\\agreement\upgood} & \shortstack{Lower-$\Dker$\\exact\upgood} & \shortstack{Combined+$\Dker$\\exact\upgood} & \shortstack{Budget\\agreement\upgood}\\
\midrule
2 & $0.018$ & $77.2$ & $55.8$ & $57.2$ & $92.0$\\
5 & $0.008$ & $83.0$ & $66.2$ & $52.2$ & $91.5$\\
10 & $0.005$ & $83.5$ & $64.5$ & $54.3$ & $91.5$\\
\bottomrule
\end{tabular}
\end{table}

\FloatBarrier

Table~\ref{tab:pilot_refitting_stability} shows that the standard deviation of $\log(1+\Dker)$ decreases from $0.018$ at 2 spc to $0.005$ at 10 spc, while backbone-order agreement increases from $77.2\%$ to $83.5\%$. The combined boundary-mass policy achieves $91.5$--$92.0\%$ agreement in selected label budgets across pilot seeds. Exact backbone choices vary more than the resulting budgets, reflecting candidates with closely matched annotation costs. These results complement the fixed-probe split-half evaluation in Table~\ref{tab:estimatorreliability} by characterizing variation across independently refitted pilot heads.

\begin{table}[!htbp]
\centering\scriptsize
\caption{Prospective backbone-selection policy simulation. Policies are leave-one-class-pair-out regressions from the listed diagnostics to label budget; selection uses only diagnostic-heldout training features.}
\label{tab:transfer_policy_app}
\setlength{\tabcolsep}{2.2pt}
\begin{tabular}{p{.32\linewidth}cccc}
\toprule
Policy & selection acc.\upgood & mean regret\downgood & rel. regret\downgood & mean AULC\upgood\\
\midrule
\psimethod{$\log(1+\Dker)$} & \psimethod{$25.0$} & \psimethod{$0.10$} & \psimethod{$5.0$} & \psimethod{$0.977$}\\
LogME & $22.5$ & $0.02$ & $0.9$ & $0.980$\\
H-score & $25.0$ & $0.03$ & $1.4$ & $0.980$\\
TransRate & $42.5$ & $1.59$ & $79.5$ & $0.959$\\
transferability set & $27.5$ & $0.05$ & $2.6$ & $0.977$\\
\psimethod{transferability set + $\Dker$} & \psimethod{$27.5$} & \psimethod{$0.05$} & \psimethod{$2.6$} & \psimethod{$0.977$}\\
all non-$\Dker$ + transferability & $30.0$ & $0.06$ & $2.8$ & $0.976$\\
\psimethod{all non-$\Dker$ + transferability + $\Dker$} & \psimethod{$30.0$} & \psimethod{$0.06$} & \psimethod{$2.8$} & \psimethod{$0.976$}\\
\bottomrule
\end{tabular}
\end{table}

\begin{figure}[!htbp]
\centering
\includegraphics[width=.62\linewidth]{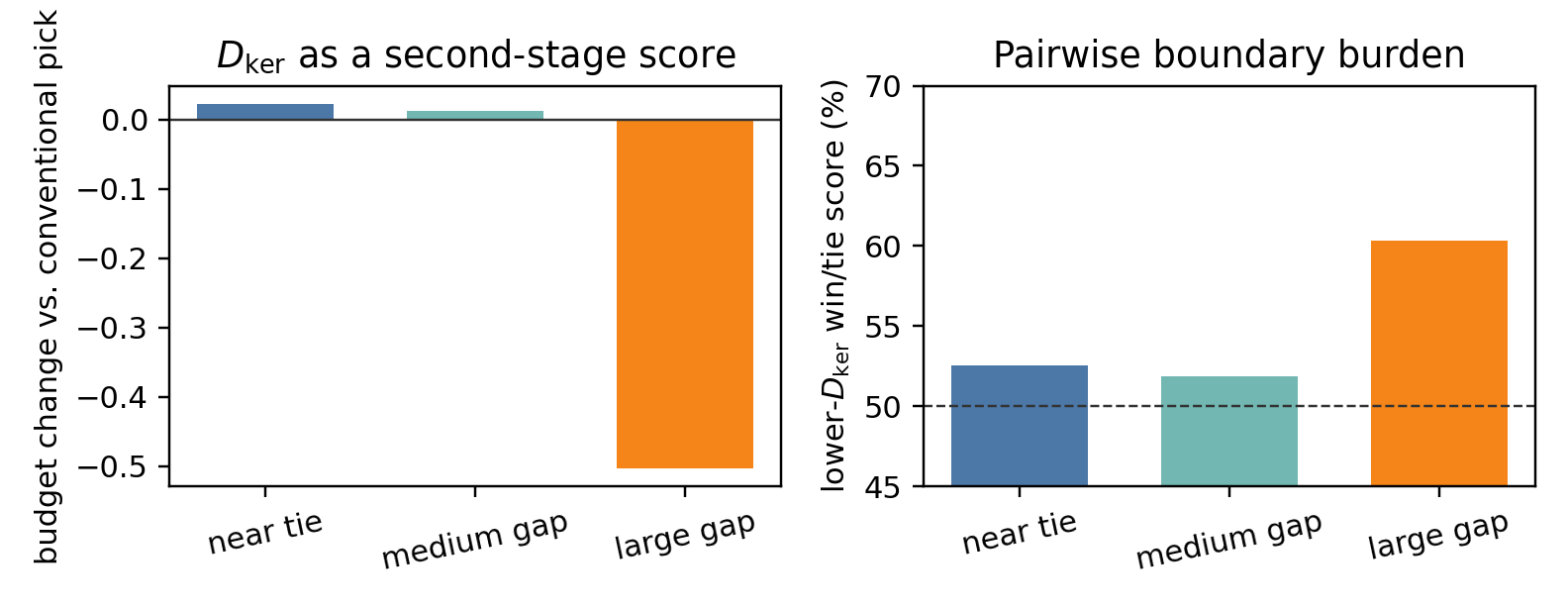}
\caption{Near-tie selection evaluation. Lower $\Dker$ is most informative when non-boundary metrics produce a near tie; large conventional-score gaps define the abstention regime.}
\label{fig:neartie_app}
\end{figure}
\begin{table}[!htbp]
\centering\small
\caption{Matched-utility pairwise evaluation under the strict diagnostic protocol. For the lower-$\Dker$ backbone, win/tie scores assign $1$, $0.5$, and $0$ to lower, equal, and higher label budgets, respectively. Comparisons group backbone pairs by full-shot accuracy gap.}
\label{tab:matchedutility_app}
\begin{tabular}{lccc}
\toprule
Full-shot acc. gap & comparisons & lower-$\Dker$ win/tie score\upgood & mean relative saving\upgood\\
\midrule
$\le 1\%$ & 176 & $54.3$ & $1.2$\\
$\le 2\%$ & 227 & $54.6$ & $2.4$\\
\bottomrule
\end{tabular}
\end{table}

\FloatBarrier
\paragraph{Interpretation of the policy tables.}
Tables~\ref{tab:transfer_policy_app} and~\ref{tab:matchedutility_app} evaluate practical decisions. The full baseline and boundary-augmented regression policies share the displayed aggregate point estimates: $30.0\%$ selection accuracy, $0.06$-spc mean regret, $2.8\%$ relative regret, and $0.976$ mean AULC. These results reflect the overlap among strong diagnostics. The separate top-2 rule in Section~\ref{sec:second_stage} uses boundary mass specifically within a shortlist and reduces regret from $0.058$ to $0.030$ spc. In the matched-utility evaluation, mean relative savings are $1.2\%$ for full-shot accuracy gaps up to $1\%$ and $2.4\%$ for gaps up to $2\%$.
\FloatBarrier
\subsection{Objective-separation and foundation-backbone experiments}\label{app:control_audits}
\begin{table}[!htbp]
\centering\small
\caption{Feature-noise control evaluation. Clean accuracy, NLL, and ECE dominate robust-accuracy selection under isotropic feature noise, while $\Dker$ measures the complementary boundary-mass axis.}
\label{tab:feature_noise_app}
\begin{tabular}{lcc}
\toprule
Metric & acc.-drop corr. & robust-selection acc.\upgood\\
\midrule
\psimethod{$\log(1+\Dker)$} & \psimethod{$0.222/0.036$} & \psimethod{$50.0\%$}\\
Gaussian-area proxy & $-0.202/0.054$ & $40.0\%$\\
noise flip proxy & $0.588/-0.029$ & $45.0\%$\\
median margin & $-0.199/-0.026$ & $52.5\%$\\
NLL & $0.065/0.175$ & $72.5\%$\\
ECE & $0.157/0.165$ & $75.0\%$\\
clean acc. & $0.012/0.013$ & $100.0\%$\\
\bottomrule
\end{tabular}
\end{table}

\begin{table}[!htbp]
\centering\small
\caption{Foundation-backbone evaluation status. Completed experiments include DINO and SigLIP alongside ResNet-50 and ViT-B/16.}
\label{tab:foundation_backbone_app}
\begin{tabular}{ll}
\toprule
Backbone & Evaluation status\\
\midrule
CLIP ViT-B/16 & Pending evaluation\\
DINO ViT-B/16 & Completed\\
DINOv2 Base & Pending evaluation\\
MAE ViT-Base & Pending evaluation\\
SigLIP-B/16 & Completed\\
\bottomrule
\end{tabular}
\end{table}

\paragraph{End-to-end diagnostic.} The ResNet-18/CIFAR runs examine the boundary readout when representation, density, and boundary evolve together. Boundary-pressure regularization lowered the held-out $\hat\psi$ estimate relative to standard, margin, entropy, and weight-decay controls while improving low-label probes. Because joint optimization with cross-entropy can still move margins and feature norms, the controlled frozen-feature experiments provide the mechanism evidence with the fewest moving parts.
\FloatBarrier
\section{Multiclass OvR--Interface Estimator Comparison}\label{app:multi}
For logits $g_j(z)=w_j^\top z+b_j$, $j\in[K]=\{1,\ldots,K\}$, define $f(z)=\arg\max_j g_j(z)$ with fixed tie breaking and cells $A_j=\{z:f(z)=j\}$. The interface-counted construction measures each active shared boundary once:
\[
\psieff^{\rm int}(f,P)=\sum_{j<k}\frac{2\int_{\partial A_j\cap\partial A_k}p(x)d\mathcal H^{d-1}(x)}{\min(P(A_j),P(A_k))}.
\]
The interface construction measures active pairwise boundaries of the argmax partition, while OvR aggregation matches the classwise surrogate used for measurement and intervention. Together, they provide geometric and operational views of multiclass boundary complexity. OvR uses $K$ classwise terms; the interface construction considers $O(K^2)$ candidate pairs.

\paragraph{Experimental protocol.}
We align class identities across feature caches using the Tiny-ImageNet class identifiers and evaluate Hard OvR, Smooth OvR, Hard Interface, and Smooth Interface on identical diagnostic sets covering 40 five-way Tiny-ImageNet tasks, four frozen backbones, and five random splits. Interface estimators gate pairwise boundaries by top-score competition, while OvR estimators aggregate one term per class. Hard estimators use tube indicators; Smooth estimators use sigmoid weights. Within-task Spearman correlation measures agreement between backbone rankings, pairwise concordance measures agreement on backbone pairs, selected-backbone agreement measures agreement on the lowest-score backbone, and mean rank displacement measures average rank changes. Table~\ref{tab:multiclass_agreement_app} reports the full comparison. The 16-task experiments in Tables~\ref{tab:multiclass_stress_app} and~\ref{tab:multiclass_policy_supplement_app} evaluate backbone-selection policy regret; this 40-task evaluation evaluates estimator-ranking agreement.

\begin{table}[!htbp]
\centering\small
\caption{Agreement among multiclass boundary-mass estimators on 40 five-way Tiny-ImageNet tasks, four frozen backbones, and five random splits with shared semantic class identities across backbones. The comparisons isolate multiclass aggregation, smoothing, and their combined effect on backbone ranking.}
\label{tab:multiclass_agreement_app}
\setlength{\tabcolsep}{3pt}
\begin{tabular}{p{.36\linewidth}cccc}
\toprule
Comparison & \shortstack{Within-task\\Spearman\upgood} & \shortstack{Pairwise\\concordance\upgood} & \shortstack{Selected-backbone\\agreement\upgood} & \shortstack{Mean rank\\displacement\downgood}\\
\midrule
Hard OvR vs.\ Hard Interface & $0.872$ & $90.0\%$ & $89.5\%$ & $0.2925$\\
Smooth OvR vs.\ Hard OvR & $0.988$ & $99.0\%$ & $98.5\%$ & $0.0300$\\
Smooth Interface vs.\ Hard Interface & $0.992$ & $99.3\%$ & $99.5\%$ & $0.0200$\\
Smooth OvR vs.\ Smooth Interface & $0.868$ & $89.7\%$ & $89.5\%$ & $0.3025$\\
\bottomrule
\end{tabular}
\end{table}

\paragraph{Competition-distance implementation.}
For logits at a diagnostic feature $z$, let $r_j=\arg\max_{k\ne j}g_k(z)$ and $d_j=(g_j(z)-g_{r_j}(z))/\|w_j-w_{r_j}\|$, with pairwise distances evaluated for distinct head weights. The ranking comparison sums the classwise tube scores based on $d_j$ and diagnostic class frequencies. Interface scores use $d_{jk}=(g_j-g_k)/\|w_j-w_k\|$, gated by whether $j,k$ are the sample's top two classes. For a competition distance $v\in\{d_j,d_{jk}\}$, Hard uses $\1\{|v|\le\varepsilon_0\}$ and Smooth uses $\sigma((\varepsilon_0-|v|)/\tau)$, with the same width and balance convention. These competition scores evaluate the multiclass partition; the frozen-feature intervention uses the independently parameterized classwise affine scores in Section~\ref{sec:frozen_estimator}.

\paragraph{Results.}
Hard OvR and Hard Interface achieve mean within-task Spearman $0.872$, pairwise concordance $90.0\%$, selected-backbone agreement $89.5\%$, and mean rank displacement $0.2925$. Their Smooth counterparts yield $0.868$, $89.7\%$, $89.5\%$, and $0.3025$, respectively. Smooth OvR versus Hard OvR yields Spearman $0.988$ and concordance $99.0\%$, while Smooth Interface versus Hard Interface yields $0.992$ and $99.3\%$. These results show strong empirical agreement between the multiclass constructions and stable backbone rankings under smoothing.

\FloatBarrier
\section{Supplementary Technical Appendix}\label{app:extended_tech}
The main claims rely on the smooth tube expansion, the smooth-diffeomorphism pushforward law, and the fixed-feature linear-head estimator. This supplementary technical appendix records longer proofs, supplementary derivations under additional regularity assumptions, non-asymptotic calculations, multi-class details, and extra experiments that provide context for the geometry--statistics--surrogate chain.

\subsection{Background}\label{app:ext_background}
\label{sec:prelim}
%=======================================================================

Throughout this appendix, $P$ is the input law with locally bounded density $p$ on $\R^d$; $Q=T_\#P$ or $Q=\phi_\#P$ is the represented law with density $q$. For maps $\R^d\to\R^m$, $d$ and $m$ are the input and feature dimensions. Random inputs and features are $X\sim P$ and $Z\sim Q$; their realizations are $x,z$. For a scalar random variable $U$, $p_U$ denotes its density. Training sample size is $n$, diagnostic sample size is $N$, and class count is $K$. Generic positive constants $c,C$ may vary across statements; subscripts indicate parameter dependence. The same conventions apply to indexed variants.
For $A\subset\R^d$ with $C^1$ boundary, $n(x)$ denotes the outward unit normal
at $x\in\partial A$.  $\mathcal H^r$ denotes $r$-dimensional Hausdorff measure, and $\Hd$ is its hypersurface specialization. We write $B(x,r)=\{u:\|u-x\|_2<r\}$, $\mathcal S^{d-1}=\{u\in\R^d:\|u\|_2=1\}$, and $e_j$ for the $j$th coordinate vector. $GL(d)$ and $O(d)$ are the invertible and orthogonal real matrix groups; $I$ denotes the identity in the dimension of the displayed matrix. Vector norms $|v|$ and $\|v\|$ are Euclidean, and $\|W\|_F=(\sum_{ij}W_{ij}^2)^{1/2}$. The measure $\lambda^r$ is $r$-dimensional Lebesgue measure. In learning bounds, $\eta$ is a label-flip probability, $\delta\in(0,1)$ a failure probability, and $\varepsilon>0$ a target disagreement error. In noise-sensitivity formulas, $\delta$ instead denotes the explicitly specified perturbation variance. Function spaces $C^j$ contain functions with $j$ continuous derivatives, $C_c^1$ additionally requires compact support, and $W^{2,1}$ denotes integrable weak derivatives through order two.

\paragraph{Area formula.}
For a $C^1$ diffeomorphism $T$, a hypersurface $\Sigma\subset\R^d$ with unit normal $n$, and a nonnegative measurable integrand $u$:
\begin{equation}
\label{eq:area}
  \int_{T(\Sigma)}u(z)\,d\Hd(z)
  =\int_\Sigma u(T(x))\,|\det DT(x)|\,|(DT(x)^{-\top})n(x)|\,d\Hd(x).
\end{equation}

\paragraph{Isoperimetry.}
$P$ is $c$-\textbf{isoperimetric} \citep{bubeck2021universal} if every $L$-Lipschitz
$g:\R^d\to\R$ satisfies $\Var_P(g)\le L^2/c^2$.
$\mathcal{N}(0,I_d)$ satisfies this with $c=1$.

We write $\gauss(t):=(2\pi)^{-1/2}e^{-t^2/2}$, $\Phi(t):=\int_{-\infty}^t\gauss(u)\,du$, and $\gauss_d(x):=(2\pi)^{-d/2}e^{-\|x\|^2/2}$,
and $J_0(s):=\sum_{k=0}^\infty(-1)^k(s/2)^{2k}/(k!)^2$ for the Bessel function.

%=======================================================================
\subsection{Effective Boundary Complexity}\label{app:ext_boundary}
\label{sec:psieff}
%=======================================================================

\subsubsection{Generalizes the halfspace margin}

\begin{proposition}[Margin connection]
\label{prop:margin}
For $P=\mathcal N(0,I_d)$ and threshold $a\in\R$, put $A_a=\{x:x_1\ge a\}$ and $f_a=\1_{A_a}$:
\begin{equation}
\label{eq:margin}
  \psieff(f_a,P)=\mathfrak g(a),\qquad \mathfrak g(a):=\frac{2\gauss(a)}{\Phi(-|a|)},\quad a\in\R.
\end{equation}
This function is continuous on $\R$, smooth on $(-\infty,0)$ and $(0,\infty)$, strictly increasing in $|a|$, attains its
global minimum $2\sqrt{2/\pi}$ at $a=0$, and satisfies
$\psieff(f_a,P)\sim 2|a|$ as $|a|\to\infty$.
\end{proposition}

\begin{proof}[Full proof in Appendix~\ref{app:margin}]
$\mup_P(f_a)=2\gauss(a)$ by L'H\^opital; $\min(P(f_a=1),P(f_a=0))=\Phi(-|a|)$.
Strict monotonicity follows from the classical inequality $\gauss(a)>a\Phi(-a)$
(proved via $\Phi(-a)<\gauss(a)/a$).
\end{proof}

\begin{remark}[Relation to Gaussian surface area and noise sensitivity]
\label{rem:gsa}
For $P=\mathcal{N}(0,I_d)$ and $P(A)=1/2$ (balanced):
$\psieff(f,P)=4\int_{\partial A}\gauss_d\,d\Hd=4\gamma^+(A)$, where $\gamma^+(A)$
is the \textbf{Gaussian surface area} (GSA) \citep{ledoux2001concentration}.  Thus
$\psieff$ equals four times the Gaussian surface area for balanced classifiers.

For \textbf{noise sensitivity} (NS) $\mathrm{NS}_\delta(f):=\Prob(f(X)\ne f(X_\delta))$ with
$X_\delta=\sqrt{1-\delta}X+\sqrt{\delta}U$, $0<\delta<1$, and independent $X,U\sim\mathcal N(0,I_d)$: for $C^2$ boundaries,
$\mathrm{NS}_\delta(f)=\gamma^+(A)\sqrt{2\delta/\pi}+O(\delta)$ as $\delta\to 0$
\citep{borell1975brunn}.  Hence
$\psieff(f,P)=2\sqrt{2\pi}\lim_{\delta\to 0}\mathrm{NS}_\delta(f)/\sqrt{\delta}$.

The key differences from these existing notions are: (a)~$\psieff$ is defined
for general distributions $P$ with the stated density regularity; (b)~the normalization by $\min(P(A),P(A^c))$
handles class imbalance; and (c)~$\psieff$ has an exact pushforward formula
(Theorem~\ref{thm:pushforward}) that makes it tractable under representations.
\end{remark}

\subsubsection{Boundary-tube coefficient and geometric bias}

\begin{theorem}[Supplementary tube-rate interpretation]
\label{thm:tsybakov}
Let $f=\1_A$ with compact $C^2$ boundary $\partial A$, and $P$ have density $p$
$C^1$ and bounded above and below on a neighborhood of $\partial A$.
\begin{enumerate}[label=(\roman*),leftmargin=*,itemsep=2pt]
\item \textbf{(Exact asymptotics)}
  $P(\operatorname{dist}(X,\partial A)\le t)=\mup_P(f)\cdot t+O(t^2)$ as $t\to 0^+$.

\item \textbf{(Two-sided tube bound)}
  There exist $t_0,c_1,c_2>0$ (depending on $p$ and the curvature of $\partial A$)
  such that for all $t\in(0,t_0)$:
  \[
    c_1\mup_P(f)\cdot t
    \;\le\;
    P(\operatorname{dist}(X,\partial A)\le t)
    \;\le\;
    c_2\mup_P(f)\cdot t.
  \]

\item \textbf{(Margin constant identification)}
  $C_{\mathrm{geo}}:=\lim_{t\to 0^+}P(\operatorname{dist}(X,\partial A)\le t)/t
  =\mup_P(f)=\psieff(f,P)\cdot\min(P(A),P(A^c))$.

\item \textbf{(Local-vote reading)}
  For voting radius $r>0$, the tube contributes $C_{\mathrm{geo}}r+O(r^2)$ to the constructive local-vote bound in Theorem~\ref{thm:upper_bound}.
\end{enumerate}
\end{theorem}

\begin{proof}[Full proof in Appendix~\ref{app:tsybakov}]
Part~(i): tubular neighborhood theorem plus co-area formula.
Part~(ii): $C^2$ boundary gives uniformly bounded principal curvatures, yielding the $O(t^2)$ error and the constants $c_1,c_2$.
Part~(iii): direct from~(i) and Definition~\eqref{eq:def}.
Part~(iv): this is the standard boundary-estimation interpretation of the tube
constant; the controlled experiments test the resulting normalization directly.
\end{proof}

\subsubsection[Supplementary axiomatic characterization of psieff]{Supplementary axiomatic characterization of $\psieff$}
\label{sec:uniqueness}

We show that $\psieff$ is the \textbf{unique} functional in a natural class
characterized by four axioms: boundary-integral locality, density linearity,
isotropy, and normalization.
This theorem provides a foundation-level justification for why $\psieff$
is the \textbf{canonical} representation-dependent geometric complexity measure within this axiomatic class.

\begin{definition}[Boundary complexity functional]
\label{def:bcf}
A \textbf{boundary complexity functional} is a map
$\mathcal C:\{(f,P)\}\to[0,\infty)$, where $f=\1_A$ has a $C^2$ boundary,
$P$ has a $C^1$ density near $\partial A$, $0<P(A)<1$, and
$\int_{\partial A}p\,d\Hd<\infty$. This domain includes compact boundaries and Gaussian halfspaces. The functional satisfies:
\begin{enumerate}[label=(B\arabic*),leftmargin=*,itemsep=2pt]
\item $\mathcal C(f,P)\ge 0$ with $\mathcal C(f,P)=0$ exactly when $\int_{\partial A}p\,d\Hd=0$.
\item $\mathcal C(f,P)<\infty$ for all admissible $(f,P)$.
\end{enumerate}
\end{definition}

\begin{theorem}[Supplementary axiomatic characterization of $\psieff$]
\label{thm:unique}
Let $\mathcal C$ be a boundary complexity functional satisfying the following axioms:
\begin{enumerate}[label=(U\arabic*),leftmargin=*,itemsep=3pt]
\item \textbf{(Boundary integral locality)}
  There exists a measurable kernel $\kappa:\R^d\times\mathcal{S}^{d-1}\times\R_+\to\R_+$ such that for all admissible $(f,P)$:
  \begin{equation}
  \label{eq:bcf_form}
    \mathcal C(f,P)=\frac{\displaystyle\int_{\partial A}\kappa(x,n(x),p(x))\,d\Hd(x)}
    {\min(P(A),P(A^c))}.
  \end{equation}

\item \textbf{(Density linearity)}
  For all $\alpha>0$: $\kappa(x,n,\alpha p)=\alpha\kappa(x,n,p)$,
  i.e., $\kappa$ is linear in its third argument.

\item \textbf{(Spatial and directional isotropy)}
  For all orthogonal $R\in O(d)$ and all translations $v\in\R^d$:
  $\kappa(Rx+v,Rn,p)=\kappa(x,n,p)$.
  That is, $\kappa$ depends on $(x,n,p)$ only through $p$.

\item \textbf{(Normalization)}
  $\mathcal C(f_0,\mathcal{N}(0,I_d))=2\sqrt{2/\pi}$, where
  $f_0:=\1\{x_1\ge 0\}$ is the balanced halfspace.
\end{enumerate}
Then $\mathcal C=\psieff$, i.e., $\kappa(x,n,p)=2p$ and $\mathcal C(f,P)=\psieff(f,P)$.
\end{theorem}

\begin{proof}[Full proof in Appendix~\ref{app:uniqueness}]
(U1)--(U3) give $\kappa(x,n,p)=cp$ for some $c>0$;
(U4) fixes $c=2$ via Proposition~\ref{prop:margin}.
\end{proof}

\begin{remark}[Interpretation of axioms]
Axiom~(U1) expresses $\mathcal C$ as an integral of local boundary contributions.
Axiom~(U2) says $\mathcal C$ scales linearly with the data density (if we scale up the
density by $\alpha$, the boundary is $\alpha$ times as ``dangerous'').
Axiom~(U3) expresses invariance of the local kernel under translations and rotations.
These three axioms together force $\kappa(x,n,p)=cp$---the only local, density-linear,
isotropic integrand is a constant multiple of $p$.
Axiom~(U4) then fixes $c=2$ via the known halfspace formula.

Within this restricted class of local, density-linear, isotropic functionals, this
characterizes $\psieff$ up to the stated Gaussian halfspace normalization.  This
axiom system explains the normalization used by the boundary-mass diagnostic.
\end{remark}

\begin{remark}[Relation to the transformation law]
Theorem~\ref{thm:unique} is complementary to Theorem~\ref{thm:pushforward}:
\begin{itemize}[leftmargin=*,itemsep=2pt]
\item Theorem~\ref{thm:unique} characterizes \textbf{which} functional $\psieff$ is
  (it is the unique isotropic density-linear boundary functional).
\item Theorem~\ref{thm:pushforward} characterizes \textbf{how} $\psieff$ transforms
  under representations (it obeys the exact Jacobian formula).
\end{itemize}
Together they give two complementary checks on the diagnostic object: an
axiomatic normalization in a restricted functional class and a concrete
pushforward law for smooth representation maps.
\end{remark}

\subsubsection{Separation from Gaussian surface area and noise sensitivity}
\label{sec:separation}

The supplementary characterization (Theorem~\ref{thm:unique}) explains the
normalization used by $\psieff$.  One may still ask whether $\psieff$ essentially
\textbf{behave} like GSA or noise sensitivity, just in a different normalization?
We answer this with explicit separation examples: concrete $(f,P,T)$ triples where
$\psieff$ and GSA (or NS) respond differently to class imbalance and representation geometry.

\begin{proposition}[Separation 1: class-balance sensitivity of $\psieff$]
\label{prop:sep_imbalance}
For a halfspace $f_a=\1\{x_1\ge a\}$ under $P=\mathcal{N}(0,I_d)$, $a\ne 0$:
\begin{enumerate}[label=\normalfont(\roman*),leftmargin=*,itemsep=2pt]
\item $\psieff(f_a,P)=2\gauss(a)/\Phi(-|a|)\to\infty$ as $|a|\to\infty$.
\item $\gamma^+(A_a)=\gauss(a)\to 0$ as $|a|\to\infty$.
\end{enumerate}
So $\psieff$ diverges as the classifier becomes more imbalanced,
correctly reflecting that learning from a heavily imbalanced boundary is harder.
GSA \textbf{decreases}, failing to capture this.
\end{proposition}

\begin{proof}
Both formulas are from Proposition~\ref{prop:margin} and the definition of GSA.
For $a\to\infty$: $\gauss(a)\sim\gauss(a)\to 0$ (Gaussian tail) while
$\Phi(-|a|)\sim\gauss(a)/a\to 0$ faster, so $\psieff=2\gauss(a)/\Phi(-a)\sim 2a\to\infty$.
\end{proof}

\begin{proposition}[Separation 2: representation sensitivity of $\psieff$]
\label{prop:sep_transform}
Let $P=\mathcal{N}(0,I_d)$, $W_s=\mathrm{diag}(s,1,\ldots,1)$ for $s>0$.

\textbf{(Sanity checks for the pushforward formula.)}
We verify four special cases of Theorem~\ref{thm:pushforward}:

\begin{enumerate}[label=\normalfont(\alph*),leftmargin=*,itemsep=3pt]
\item \textbf{Orthogonal transform ($W\in O(d)$):}
  $|W^{-\top}n|=|W^{-1}n|=|n|=1$, so $\psieff$ is invariant.
  This matches the isometry interpretation: orthogonal maps leave the boundary coefficient unchanged.

\item \textbf{Isotropic scaling ($W=cI$, $c>0$):}
  $|(cI)^{-\top}n|=c^{-1}$. The pushforward density also scales:
  $q(z)=c^{-d}p(z/c)$.  The net effect:
  $\mup_Q(g)=2c^{-1}\int_{\partial A}p(x)\,d\mathcal{H}^{d-1}=c^{-1}\mup_P(f)$.
  Class balance is unchanged, so $\psieff(g,Q)=c^{-1}\psieff(f,P)$.
  This matches the normal-stretch interpretation: scaling by $c$ reduces the coefficient by $c^{-1}$.

\item \textbf{Axis-aligned halfspace $f=\1\{x_1\ge 0\}$ under $W_s$:}
  Outward normal $n=-e_1$, $|W_s^{-\top}n|=s^{-1}$, $p(0,x_2,\ldots)=(2\pi)^{-d/2}e^{-\|(x_2,\ldots)\|^2/2}$.
  The boundary integral: $\mup_Q(g)=2s^{-1}\varphi(0)=2s^{-1}/\sqrt{2\pi}$.
  Class balance $Q(g=1)=1/2$ (symmetric image halfspace).
  Hence $\psieff(g,Q)=2s^{-1}\sqrt{2/\pi}$ \textbf{decreases as $s$ grows}.

\item \textbf{Diagonal halfspace $f=\1\{x_1+x_2\ge 0\}$ under $W_s$:}
  Outward normal $n=-(1,1,0,\ldots)/\sqrt{2}$, $|W_s^{-\top}n|=\sqrt{s^{-2}/2+1/2}=\sqrt{(1+s^{-2})/2}$.
  The $\psieff$ in image space: $\psieff(g,Q)\propto\sqrt{(1+s^{-2})/2}$,
  which \textbf{decreases differently} from the axis-aligned case.
  By contrast, $\gamma^+(\{x:x_1+x_2\ge0\})=\gauss(0)$ is independent of $s$.
\end{enumerate}
\vspace{-0.35em}
\noindent\textbf{The key comparison:}
Cases (c) and (d) show that $\psieff$ transforms differently for different boundary orientations under the same anisotropic scaling $W_s$, whereas GSA in case~(d) is insensitive to $s$.
This establishes the orientation-sensitive response of $\psieff$ to anisotropy, alongside the invariance of the reference GSA.
\end{proposition}

\begin{proposition}[First-order Gaussian noise-sensitivity coefficient]
\label{prop:sep_ns}
Let $P=\mathcal N(0,I_d)$ and $f=\1_A$ have compact $C^2$ boundary with $P(A)=1/2$. For $X,U\sim\mathcal N(0,I_d)$ independent and $X_\delta=\sqrt{1-\delta}X+\sqrt\delta U$,
\begin{equation}\label{eq:ns_expansion}
\mathrm{NS}_\delta(f)=\frac{\psieff(f,P)}{4}\sqrt{\frac{2\delta}{\pi}}+O(\delta).
\end{equation}
The same leading coefficient holds for Gaussian halfspaces by direct calculation.
\end{proposition}
\begin{proof}
Integrating the Gaussian crossing probability over both sides of a boundary gives $2\sqrt\delta\int_0^\infty\Phi(-u)\,du=\sqrt{2\delta/\pi}$ per unit Gaussian surface area. Appendix~\ref{app:ns_expansion} supplies the normal-coordinate argument.
\end{proof}
\begin{remark}[Geometric interpretation]
The normalized surface integral determines the leading response to small Gaussian perturbations. The remainder records the additional local geometry at positive noise levels.
\end{remark}

\subsubsection{Supplementary sample-complexity lower-bound discussion}

\begin{proposition}[Two-point testing under symmetric label noise]\label{prop:lb}
Let $f_0,f_1$ be two classifiers with disagreement probability $\Delta=P(f_0\ne f_1)>0$. For $n$ training observations with independent symmetric label noise $\eta\in(0,1/2)$, write $\Prob_j,\E_j$ for their law and expectation under target $f_j$, $\hat J\in\{0,1\}$ for a decoder, and $\hat f$ for a fitted classifier. Set $D_\eta=(1-2\eta)\log((1-\eta)/\eta)$. If $nD_\eta\Delta\le1/2$, then
\[
\inf_{\hat J}\max_{j=0,1}\Prob_j(\hat J\ne j)\ge1/4,
\qquad
\inf_{\hat f}\max_{j=0,1}\E_j[P(\hat f\ne f_j)]\ge\Delta/8.
\]
\end{proposition}
\begin{proof}
The one-observation KL divergence is $D_\eta\Delta$. Pinsker's inequality bounds the total variation between the $n$-sample laws by $1/2$. Le Cam's inequality gives the testing bound. Decoding a fitted classifier by its nearest hypothesis in disagreement distance gives the risk bound. Appendix~\ref{app:lb} supplies the calculation.
\end{proof}

\subsubsection{Multi-class extension}
\label{sec:multiclass}

\begin{definition}[Multi-class $\psieff$]
\label{def:multiclass}
For $K$-class classification $f:\R^d\to[K]:=\{1,\ldots,K\}$ with disjoint
decision regions $A_k:=\{x:f(x)=k\}$ ($k=1,\ldots,K$) and $P$ having bounded density $p$ with $0<P(A_k)<1$, define the \textbf{per-class} effective boundary complexity:
\[
  \psieff^{(k)}(f,P):=\frac{\mup_P(\1_{A_k})}{\min(P(A_k),1-P(A_k))},
  \qquad\mup_P(\1_{A_k})=2\int_{\partial A_k}p\,d\Hd,
\]
and the \textbf{total} multi-class effective boundary complexity:
\[
  \psieff(f,P):=\sum_{k=1}^K\psieff^{(k)}(f,P).
\]
\end{definition}

\begin{proposition}[Multi-class pushforward]
\label{prop:mc_pushforward}
Under the conditions of Theorem~\ref{thm:pushforward}: for each $k\in[K]$,
\[
  \mup_{T_\#P}(\1_{T(A_k)})=2\int_{\partial A_k}p(x)\,|(DT^{-\top})n_k(x)|\,d\Hd(x),
\]
where $n_k(x)$ is the outward unit normal to $A_k$.  Hence the total complexity
transforms as $\psieff(f\circ T^{-1},T_\#P)\le C_T\,\psieff(f,P)$ where
$C_T:=\sup_{k,x\in\partial A_k}|(DT^{-\top})n_k|$.
\end{proposition}

\begin{proof}
Apply Theorem~\ref{thm:pushforward} to each binary classifier $\1_{A_k}$;
the bound follows from the definition of $C_T$.
\end{proof}

\begin{remark}[Application to multi-class linear heads]
For 100-class one-vs-rest heads $(w_k,b_k)$ in feature dimension $m$, put $h_k(z)=\1\{w_k^\top z+b_k\ge0\}$ and $Z\sim Q$. The operational class-$k$ coefficient is $\psieff(h_k,Q)$; its surrogate is $\hat\psi_{\tau,\rho}^{(k)}$.
The differentiable surrogate (Section~\ref{sec:surrogate}) applies per-class and
is summed to give $\hat\psi_{\tau,\rho}:=\sum_k\hat\psi_{\tau,\rho}^{(k)}$.
For partition indicators, the per-class sum counts a shared interface from both sides. The operational linear-head sum applies the same binary coefficient to independently parameterized classwise halfspaces, matching the measured surrogate and its gradients. When the goal is a purely
geometric evaluation of a multi-class partition, one can count each
class--class interface once, with the interface-weighted variant stated in
Appendix~\ref{app:mc_interface}. We use the OvR version in the main experiments
because it is the quantity directly optimized and measured by the smooth
linear-head surrogate.
\end{remark}

\begin{proposition}[OvR local-vote upper bound]\label{prop:mc_sample}
For a $K$-class partition, put $f_k=\1_{A_k}$, $\alpha_k=\min(P(A_k),1-P(A_k))>0$, and $\psi_k=\psieff^{(k)}(f,P)$. Let $\hat f$ denote the classifier formed from the $K$ binary local-vote rules, and let $c_\eta=1-e^{-(1-2\eta)^2/2}$. Assume classwise tube bounds $P(\operatorname{dist}(X,\partial A_k)\le r)\le\alpha_k\psi_k r+L_kr^2$ for $0<r\le r_0$, with remainder constants $L_k\ge0$, and $n$ independent training observations whose classwise binary labels have conditional error at most $\eta<1/2$. The OvR local-majority classifier satisfies
\[
\E[P(\hat f\ne f)]\le r\sum_k\alpha_k\psi_k+r^2\sum_k L_k
+K\int e^{-c_\eta nP(B(x,r))}\,dP(x).
\]
With uniform small-ball coverage and logarithmic bandwidth, the leading geometric term is $\widetilde O(\psieff(f,P)n^{-1/d})$.
\end{proposition}

%=======================================================================
\subsection{Pushforward Geometry Under Smooth Representations}\label{app:ext_pushforward}
\label{sec:push}
%=======================================================================

\begin{theorem}[Global pushforward]
\label{thm:pushforward}
Let $P$ have density $p$ continuous near the $C^1$ boundary of $A$, and let $T:\R^d\to\R^d$ be a $C^1$ diffeomorphism with finite displayed weighted surface integral.  With $Q:=T_\#P$, $g:=\1_{T(A)}$:
\begin{equation}
\label{eq:pf}
  \mup_Q(g)=2\int_{\partial A}p(x)\,|(DT(x)^{-\top})n(x)|\,d\Hd(x).
\end{equation}
\end{theorem}

\begin{proof}[Full proof in Appendix~\ref{app:push}]
$\partial T(A)=T(\partial A)$; area formula~\eqref{eq:area} with the change-of-density
$q(T(x))=p(x)/|\det DT(x)|$ causes the Jacobian determinant to cancel.
\end{proof}

\paragraph{Supplementary extensions.}
The diffeomorphic formula above is the core pushforward result used in the
main text.  The remaining statements in this section record broader variants
for piecewise-smooth, nonsmooth, or dimensionality-reducing maps and specify
their additional regularity assumptions.

\begin{theorem}[Piecewise-diffeomorphic pushforward]\label{thm:pw}
Let $T:\R^d\to\R^d$ be a bi-Lipschitz homeomorphism, $S$ a Lebesgue-null singular set, and $\{C_k\}$ disjoint open cells covering $\R^d\setminus S$. Assume each restriction $T_k=T|_{C_k}$ is a $C^1$ diffeomorphism, $p$ is continuous near $\partial A$, $\Hd(\partial A\cap S)=0$, and
$\sum_k\int_{\partial A\cap C_k}p\|DT_k^{-\top}n\|\,d\Hd<\infty$.
Then
\[
\mup_{T_\#P}(\1_{T(A)})=2\sum_k\int_{\partial A\cap C_k}p(x)\|DT_k(x)^{-\top}n(x)\|\,d\Hd(x).
\]
The class probabilities are preserved by $T$.
\end{theorem}

\begin{corollary}[Family-level contraction]
\label{cor:contraction}
Let $\mathcal F$ be a classifier family, $A_f=\{x:f(x)=1\}$, and $n_f$ its boundary unit normal. For a normal-expansion factor $\beta_{\rm n}>0$, suppose $|(DT^{-\top})n_f|\le\beta_{\rm n}^{-1}$ on $\partial A_f$ for every $f\in\mathcal F$. Then
$\sup_{f\in\mathcal{F}}\psieff(f\circ T^{-1},T_\#P)\le\beta_{\rm n}^{-1}\sup_f\psieff(f,P)$.
\end{corollary}

\begin{corollary}[Boundary contribution under representation]\label{cor:sc_improve}
Under Corollary~\ref{cor:contraction}, let $\psi_0=\sup_{f\in\mathcal F}\psieff(f,P)$ and suppose the original and represented local-vote protocols satisfy the coverage and uniform tube assumptions of Theorem~\ref{thm:tight_bilateral}. If the normal factor is at most $\beta_{\rm n}^{-1}$, the represented tube contribution satisfies
\[
\sup_f Q\{\operatorname{dist}(Z,\partial T(A_f))\le r\}
\le\tfrac12\beta_{\rm n}^{-1}\psi_0 r+L_T r^2,\qquad Z\sim Q,
\]
Here $L_T$ is the uniform represented-space tube-remainder constant. At bandwidth $r_n=(\gamma_n/n)^{1/d}$ with $\gamma_n=C\log n$, the leading geometric term scales by $\beta_{\rm n}^{-1}$. Holding the coverage, bandwidth, and remainder constants fixed, its sufficient sample-size contribution at error $\varepsilon$ scales as
\[
\widetilde O\!\left((\beta_{\rm n}^{-1}\psi_0/\varepsilon)^d\right).
\]
\end{corollary}
\begin{proof}
The balance $\alpha_f=\min(P(A_f),P(A_f^c))$ is preserved, and the leading tube coefficient is $\alpha_f\psieff\le\psi_0/2$. Apply the normal contraction bound and Theorem~\ref{thm:upper_bound}. Solving $\beta_{\rm n}^{-1}\psi_0(\gamma_n/n)^{1/d}\le2\varepsilon$ gives the displayed geometric contribution.
\end{proof}
\begin{remark}[Learner and coverage]
The complete local-vote bound also includes represented-space ball probabilities and tube remainders. For a known invertible $T$, unrestricted statistical experiments in the two coordinate systems are equivalent: an estimator can compose its input with $T$ or $T^{-1}$. The displayed improvement concerns the geometric contribution for the specified local-vote protocols.
\end{remark}

\begin{corollary}[Supplementary full-rank piecewise-affine cellwise formula]
\label{cor:relu}
Under Theorem~\ref{thm:pw}, for $T(x)=M_kx+b_k$ on cell $C_k$ with $M_k\in GL(d)$:
$\mup_{T_\#P}(\1_{T(A)})=2\sum_k\int_{\partial A\cap C_k}p|M_k^{-\top}n|\,d\Hd$.
\end{corollary}

\begin{theorem}[Pushforward under patchwise $C^1$ regularity]\label{thm:lipschitz_push}
Let $T:\R^d\to\R^d$ be a bi-Lipschitz homeomorphism and $A$ have compact $C^1$ boundary. Suppose, up to an $\Hd$-null set, $\partial A$ has a countable cover of patches on neighborhoods of which $T$ is $C^1$ with invertible derivative. Let $p$ be continuous near $\partial A$ and assume the weighted integral below is finite. Then
\begin{equation}\label{eq:lipschitz_push}
\mup_{T_\#P}(\1_{T(A)})=2\int_{\partial A}p(x)\|DT(x)^{-\top}n(x)\|\,d\Hd(x).
\end{equation}
\end{theorem}
\begin{proof}
The homeomorphism identifies $\partial T(A)=T(\partial A)$ and supplies a unique preimage for each image point. The smooth change-of-variables formula applies in each patch neighborhood. A disjoint measurable refinement of the cover and the area formula give Eq.~\eqref{eq:lipschitz_push}; Lipschitz continuity preserves the exceptional surface-null set.
\end{proof}
\begin{remark}[Patchwise representations]
The assumptions cover globally invertible piecewise-smooth representations whose activation interfaces meet the decision boundary in surface measure zero. They also provide the image and density correspondence used in the cellwise specialization.
\end{remark}

\begin{theorem}[Fiber-marginal density under dimensionality reduction]
\label{thm:submersion_push}
Let $T:\R^d\to\R^m$ with $m\le d$ be a $C^1$ \textbf{submersion}
(i.e., $DT(x):\R^d\to\R^m$ has full row rank $m$ for all $x$).
Let $P$ have bounded $C^1$ density $p$ on $\R^d$, and $f=\1_A$ with compact $C^1$ boundary.
Define $Q:=T_\#P$ (a measure on $\R^m$) and $g:=\1_{\overline{T(A)}}$.

For $m<d$, the coarea formula gives the pushforward density as a fiber marginal:
\begin{equation}
\label{eq:submersion_density}
q(y)=\int_{T^{-1}(y)}\frac{p(x)}{J_d^mT(x)}\,d\mathcal H^{d-m}(x),
\end{equation}
where $J_d^mT(x)=\det(DT(x)DT(x)^\top)^{1/2}$. The image $T(\partial A)$ can have full dimension $m$ in $\R^m$. Boundary-mass calculations therefore use the actual boundary $\partial T(A)$ together with the fiber marginal; transversality and fiber structure determine how that boundary relates to $\partial A$.
\end{theorem}

\begin{proof}[Full proof in Appendix~\ref{app:submersion_push}]
This is the coarea formula applied to the pushforward density and identifies the fiber marginal that replaces the diffeomorphic density identity.
\end{proof}

\begin{remark}[Relation to the diffeomorphism formula]
For a smooth map $T:\R^d\to\R^m$, the geometry determines the appropriate change-of-variables formula:
\begin{itemize}[leftmargin=*,itemsep=0pt]
\item $m=d$, $T$ diffeomorphism: the area formula gives Eq.~\eqref{eq:pf}.
\item $m<d$, submersion: the correct density is the fiber integral~\eqref{eq:submersion_density}; a separate analysis is needed to identify $\partial T(A)$.
\item $m=d$, rank-deficient $T$: collapsed boundary pieces require a singular-map analysis.
\end{itemize}
The experiments and main claims use the dimension-preserving law or the finite-feature linear-head estimator.
\end{remark}

%=======================================================================
\subsection{Extensions to Finite-Perimeter Sets and Manifolds}
\label{sec:bv}
%=======================================================================

This section extends the geometric interpretation to bounded-variation (BV) boundaries and manifold-supported data.

\subsubsection{Bounded-variation (BV) classifiers and piecewise-linear boundaries}
\label{sec:bv_main}

\begin{definition}[BV classifier]
\label{def:bv_classifier}
A classifier $f=\1_A$ is a \textbf{BV classifier} if the perimeter of $A$ is
finite: $\mathrm{Per}(A,\R^d):=\sup\{\int_A\operatorname{div}v\,dx:v\in C^1_c(\R^d,\R^d),\|v\|_\infty\le 1\}<\infty$.
The BV perimeter equals $\mathcal{H}^{d-1}(\partial^*A)$ where $\partial^*A$ is
the reduced boundary (De Giorgi's structure theorem \citep{ambrosio2000functions}).
\end{definition}

\begin{remark}[ReLU networks are BV classifiers]
On any bounded domain $\Omega\subset\R^d$, a ReLU network $f_\theta:\R^d\to\{0,1\}$ with parameters $\theta$
has a decision boundary that is a finite union of polyhedral pieces inside $\Omega$.
These sets have finite $(d-1)$-dimensional Hausdorff measure on $\Omega$, hence
finite perimeter there.  Therefore ReLU classifiers are locally BV, and whenever the
density-weighted surface integral is finite the BV extension below is well defined.
\end{remark}

For BV classifiers, $\mup_P(f)$ extends naturally via the reduced boundary:
\begin{equation}
\label{eq:mup_bv}
  \mup_P(f) := 2\int_{\partial^* A} p(x)\,d\mathcal{H}^{d-1}(x),
\end{equation}
and $\psieff(f,P):=\mup_P(f)/\min(P(A),P(A^c))$ as before.

\begin{theorem}[Supplementary finite-perimeter tube-content extension]
\label{thm:bv_bilateral}
Let $f=\1_A$ be a BV classifier with $\psieff(f,P)=\psi<\infty$ and $P(A)=1/2$.
Assume:
\begin{enumerate}[label=\normalfont(\roman*),leftmargin=*,itemsep=2pt]
\item $P$ has bounded density $p$ with $0<p_{\min}\le p\le p_{\max}<\infty$ on
  a neighborhood of $\partial^*A$.
\item The reduced boundary $\partial^*A$ satisfies a \textbf{$\delta$-regular isoperimetric}
  condition: for all $r\in(0,\delta)$ and $x\in\partial^*A$,
  $P(B(x,r)\cap A)\ge c_0 r^d$ and $P(B(x,r)\cap A^c)\ge c_0 r^d$ for some $c_0>0$.
\item The weighted Minkowski content exists at first order:
  $P(\operatorname{dist}(X,\partial^*A)\le t)=2t\int_{\partial^*A}p\,d\mathcal H^{d-1}+o(t)$.
  This holds for smoother finite-perimeter sets such as positive-reach or suitable $W^{2,1}$ boundaries.
\end{enumerate}
Then:
\begin{enumerate}[label=\normalfont(\alph*),leftmargin=*,itemsep=2pt]
\item \textbf{(Tube content)}
  $P(\operatorname{dist}(X,\partial^*A)\le t)=\mup_P(f)\cdot t + o(t)$
  as $t\to 0^+$.

\item \textbf{(Geometric bias bound)} Any plug-in rule whose errors outside estimation noise are confined to an $r$-tube around $\partial^*A$ has geometric bias $\mup_P(f)r+o(r)$.

\item \textbf{(Statistical interpretation)} The weighted tube coefficient supplies the geometric bias constant for finite-perimeter sets satisfying the stated content assumption.
\end{enumerate}
Full proof in Appendix~\ref{app:bv_bilateral}.
\end{theorem}

\begin{remark}[What changes and what stays the same]
The first-order weighted Minkowski content in part~(iii) gives the tube-content extension to finite-perimeter sets. Under this assumption, $\psieff$ controls the geometric bias constant of tube-based arguments.
\end{remark}

\begin{corollary}[Tube-bias coefficient for ReLU decision regions]
\label{cor:relu_rate}
Let $f_\theta$ be any ReLU network classifier.  Under the conditions of
Theorem~\ref{thm:bv_bilateral}, $\psieff(f_\theta,P)$ controls the first-order tube-bias constant of plug-in arguments. Minimax excess-risk analysis additionally uses regularity conditions on the decision-set class.
\end{corollary}

\subsubsection{Manifold distributions}
\label{sec:manifold}

When data lies near a $k$-dimensional submanifold $\mathcal{M}\subset\R^d$
($k<d$), the ``dimension'' of the classification problem is the intrinsic dimension $k$.  We formalize this and show $\psieff$ adapts automatically.

\begin{assumption}[Manifold distribution]
\label{assump:manifold}
$P$ is supported on a compact $C^2$ submanifold $\mathcal{M}\subset\R^d$
of intrinsic dimension $k$ ($1\le k\le d-1$), with $P$ having a bounded
$C^1$ density $p_\mathcal{M}$ with respect to the $k$-dimensional volume
form $\mathrm{vol}_\mathcal{M}$.
\end{assumption}

Assume additionally that the decision surface intersects $\mathcal M$ transversely. The intrinsic decision boundary $\partial_{\mathcal M}(A\cap\mathcal M)=\partial A\cap\mathcal{M}$
is a $(k-1)$-dimensional submanifold of $\mathcal{M}$.  We redefine $\psieff$
intrinsically:
\begin{equation}
\label{eq:psieff_manifold}
  \psieff^\mathcal{M}(f,P)
  :=\frac{2\int_{\partial A\cap\mathcal{M}}p_\mathcal{M}(x)\,d\mathcal{H}^{k-1}(x)}
  {\min(P(A),P(A^c))},
\end{equation}
using the $(k-1)$-dimensional Hausdorff measure within $\mathcal{M}$.

\begin{theorem}[Intrinsic-dimensional local-vote bound]\label{thm:manifold_rate}
Under Assumption~\ref{assump:manifold}, let the intrinsic decision boundary be compact $C^2$, with class balance $\alpha=\min(P(A),P(A^c))>0$ and $\psieff^{\mathcal M}=\psi$. Let $L$ bound the quadratic tube remainder. Assume geodesic balls $B_{\mathcal M}(x,r)$ of radius $r$ satisfy $P(B_{\mathcal M}(x,r))\ge a r^k$ for $P$-almost every $x$ and $0<r\le r_0$. For $n$ independent labeled observations with symmetric noise $\eta<1/2$, the intrinsic radius-$r$ majority-vote classifier $\hat f_{n,r}$ satisfies
\[
\E[P(\hat f_{n,r}\ne f)]\le\alpha\psi r+Lr^2+e^{-c_\eta a n r^k},
\quad c_\eta=1-e^{-(1-2\eta)^2/2}.
\]
With $r_n=(\gamma_n/n)^{1/k}$ and $\gamma_n=C\log n$, this gives the constructive rate $\widetilde O(\psi n^{-1/k}+L n^{-2/k})$ for sufficiently large $C$.
\end{theorem}
\begin{remark}[Intrinsic geometry]
The manifold tube coefficient supplies the geometric term, and the intrinsic ball volume determines neighborhood coverage. Both quantities enter the bound explicitly; Appendix~\ref{app:manifold_rate} gives the proof.
\end{remark}

\subsection{Supplementary Note: Scale-Normalized Linear Representation Shaping}
\label{sec:optimal_rep}
%=======================================================================

The pushforward law (Theorem~\ref{thm:pushforward}) tells us \textbf{how} $\psieff$
changes under a representation $T$.  A natural next question is: what is the
\textbf{optimal} $T$, the one minimizing $\psieff$ in the represented space?
This section answers that question for linear representations, giving the
first characterization theorem connecting representation geometry to learning difficulty.

\paragraph{Positioning.}
The pushforward formula motivates a constrained representation-design problem that characterizes boundary-normal shaping.

\subsubsection{Supplementary scale-normalized linear representation shaping}
\label{sec:opt_rep_main}

Let $T_W(x)=Wx$ with $W\in GL(d)$.  By Theorem~\ref{thm:pushforward}:
\[
  \psieff(f\circ T_W^{-1},(T_W)_\#P)
  =\frac{2\int_{\partial A}p(x)|W^{-\top}n(x)|\,d\mathcal{H}^{d-1}(x)}{\min(P(A),P(A^c))},
\]
For $P=\mathcal{N}(0,I_d)$ and $P(A)=1/2$, this equals $4\int_{\partial A}p(x)\|W^{-\top}n(x)\|\,d\Hd(x)$.

\begin{remark}[Why a scale constraint is necessary]
Minimizing $\psieff(f\circ T_W^{-1},(T_W)_\#P)$ over the unconstrained set $W\in GL(d)$ is ill-posed: taking $W=cI$ with $c\to 0$ sends $|W^{-\top}n|=c^{-1}\to\infty$
and hence $\psieff\to\infty$, while $c\to\infty$ sends $|W^{-\top}n|\to 0$
and $\psieff\to 0$ trivially.
A meaningful optimization requires a normalization.
We impose $\det W=1$ (volume-preserving representations), which is a natural
constraint for dimension-preserving feature maps.
\end{remark}

\begin{theorem}[Supplementary derivation: scale-normalized linear representation shaping]
\label{thm:opt_rep}
Let $f=\1_A$ have $C^1$ boundary with $0<\int_{\partial A}p\,d\Hd<\infty$, $P=\mathcal{N}(0,I_d)$, and $P(A)=1/2$.
Define the \textbf{boundary normal second-moment matrix}:
\begin{equation}
\label{eq:bnm}
  \Sigma_f := \int_{\partial A} p(x)\, n(x)n(x)^\top \, d\mathcal{H}^{d-1}(x)
  \;\;\in\R^{d\times d}.
\end{equation}
Consider the constrained optimization
$\min_{W\in GL(d),\,\det W=1}\psieff(f\circ T_W^{-1},(T_W)_\#P)$.
\begin{enumerate}[label=\normalfont(\roman*),leftmargin=*,itemsep=3pt]
\item \textbf{(Stationarity condition)}
  $W$ is a stationary point of the constrained problem if and only if:
  \begin{equation}
  \label{eq:stationarity}
    W^{-\top}\!\left[\int_{\partial A}p(x)\frac{n(x)n(x)^\top}{\|W^{-\top}n(x)\|}\,d\mathcal{H}^{d-1}(x)\right]W^{-1} = \lambda I
    \quad\text{for some }\lambda>0,
  \end{equation}
  with $\det W=1$. In words: the representation equalizes a reweighted boundary-normal second moment; the weights depend on $W$ through $\|W^{-\top}n(x)\|$.

\item \textbf{(Orthogonal transforms are flat)}
  For $W\in O(d)$ (orthogonal): $\psieff(f\circ W^{-1},W_\#P)=\psieff(f,P)$
  is constant.  Hence orthogonal transforms preserve $\psieff$; beneficial volume-preserving representations use anisotropic shaping.

\item \textbf{(Second-moment whitening is a heuristic preconditioner)}
  Let $\Sigma_f=V\Lambda V^\top$ (eigendecomposition, $\Lambda=\mathrm{diag}(\lambda_1\ge\cdots\ge\lambda_d>0)$).
  The volume-normalized map $W_{\rm pre}:=(\det\Sigma_f)^{-1/(2d)}V\Lambda^{1/2}V^\top$ whitens the unweighted second moment $\Sigma_f$ and provides a heuristic preconditioner. Solving~\eqref{eq:stationarity} additionally incorporates the weights $\|W^{-\top}n(x)\|^{-1}$.
\end{enumerate}
Full proof in Appendix~\ref{app:opt_rep}.
\end{theorem}

\begin{remark}[Interpretation]
The stationarity condition~\eqref{eq:stationarity} is nonlinear in $W$ because the normal weights depend on the transformed norm. This gives a nonlinear boundary-sensitive whitening problem, with ordinary eigendecomposition applicable in special symmetric cases.
\end{remark}

\begin{corollary}[Local descent from an anisotropic normal moment]
\label{cor:opt_rep_reduction}
Under Theorem~\ref{thm:opt_rep}, set
$H=\Sigma_f-d^{-1}\operatorname{tr}(\Sigma_f)I$.
The curve $W_t=\exp(tH)$ has determinant one and satisfies
\[
\left.\frac{d}{dt}\psieff(f\circ W_t^{-1},(W_t)_\#P)\right|_{t=0}
=-4\|H\|_F^2.
\]
Thus every anisotropic $\Sigma_f$ supplies a strict local descent direction within volume-preserving linear maps.
\end{corollary}

\subsection{Constructive Boundary-Rate Calculations}\label{sec:upper}
\subsubsection{Local majority-vote bound}
\begin{theorem}[Boundary bias and local coverage]\label{thm:upper_bound}
Let $f=\1_A$, $\alpha=\min(P(A),P(A^c))>0$, and $\psi=\psieff(f,P)$. Let $L\ge0$ be a tube-remainder constant and $r_0>0$ an admissible radius. Assume
\[
P\{\operatorname{dist}(X,\partial A)\le r\}\le\alpha\psi r+Lr^2,
\qquad 0<r\le r_0.
\]
For $n$ independent observations with symmetric label noise $\eta\in(0,1/2)$, let $\hat f_{n,r}$ predict the majority label in $B(x,r)$, using a fixed tie and empty-ball rule. Then
\begin{equation}\label{eq:local_vote_coverage}
\E[P(\hat f_{n,r}\ne f)]\le\alpha\psi r+Lr^2
+\int e^{-c_\eta n P(B(x,r))}\,dP(x),
\quad c_\eta=1-e^{-(1-2\eta)^2/2}.
\end{equation}
For a measurable coverage set $G$, tail bound $\xi\ge0$, and mass constant $a>0$, if $P(G^c)\le\xi$ and $P(B(x,r))\ge a r^d$ for $x\in G$, the last term is at most $\xi+e^{-c_\eta a n r^d}$. In particular, a uniform small-ball bound gives
\[
\E[P(\hat f_{n,r_n}\ne f)]
\le\alpha\psi(\gamma_n/n)^{1/d}+L(\gamma_n/n)^{2/d}+e^{-c_\eta a\gamma_n}
\]
for any bandwidth factor $\gamma_n>0$ with $r_n=(\gamma_n/n)^{1/d}\le r_0$.
\end{theorem}
\begin{proof}
Outside the boundary tube, every training point in $B(x,r)$ has the same clean label as $x$. Conditional on its sample count $N_r(x)$, Hoeffding's inequality gives error at most $e^{-(1-2\eta)^2N_r(x)/2}$. Averaging its binomial distribution yields the integrand in Eq.~\eqref{eq:local_vote_coverage}. Appendix~\ref{app:ub} gives the details.
\end{proof}
\begin{remark}[Gaussian coverage]
For $P=\mathcal N(0,I_d)$, choose $G=B(0,R)$. For $r\le r_0$,
$P(B(x,r))\ge a_R r^d$ on $G$, where
$a_R=v_d(2\pi)^{-d/2}e^{-(R+r_0)^2/2}$ and $v_d$ is unit-ball volume.
The explicit tail term is $\xi_R=P(\|X\|>R)$. Thus the Gaussian bound retains both coverage and tail probability.
\end{remark}

\subsubsection{Uniform constructive bound}
\begin{theorem}[A family with uniform tube and coverage constants]\label{thm:tight_bilateral}
Let $\mathcal F$ have tube widths $r\le r_0$, remainder constant $L$, and complexities at most $\Psi$, all under one distribution $P$ with $P(B(x,r))\ge a r^d$ almost everywhere. Under symmetric noise $\eta$, local majority vote obeys
\[
\sup_{f\in\mathcal F}\E[P(\hat f_{n,r}\ne f)]
\le\tfrac12\Psi r+Lr^2+e^{-c_\eta a n r^d}.
\]
Choosing $\gamma_n=C\log n$ with $c_\eta a C\ge2$ gives a uniform constructive upper bound
$\widetilde O(\Psi n^{-1/d}+L n^{-2/d})$.
\end{theorem}
\begin{proof}
Take the supremum in Theorem~\ref{thm:upper_bound}, using $\alpha_f\le1/2$ and the shared constants. Appendix~\ref{app:tight_bilateral} records the substitution.
\end{proof}
\begin{remark}[Boundary-class theory]
Classical boundary-estimation rates additionally depend on decision-set smoothness, metric entropy, density, and label-noise conditions~\citep{mammen1999smooth,tsybakov2009introduction}. The constructive result here identifies the linear dependence of its geometric bias on $\psieff$. The finite-family lower bounds below state their packing and information quantities explicitly.
\end{remark}

\subsubsection{Halfspaces under fixed symmetric noise}
\begin{theorem}[Halfspace learning with a fixed noise gap]\label{thm:halfspace_tight}
Let $\mathcal H_d$ be the class of affine halfspaces in $\R^d$, $f^*\in\mathcal H_d$, and $Y=f^*(X)\oplus\xi$ with independent $\xi\sim\Ber(\eta)$, $\eta<1/2$, where $\Ber(\eta)$ is the Bernoulli law and $\oplus$ adds binary labels modulo two. For $n\ge d+1$, a measurable empirical-risk minimizer satisfies, with probability at least $1-\delta$,
\[
P(\hat f\ne f^*)\le
C\frac{(d+1)\log(en/(d+1))+\log(2/\delta)}{(1-2\eta)^2n}.
\]
Consequently $\widetilde O((d+\log(1/\delta))/((1-2\eta)^2\varepsilon))$ samples suffice for clean disagreement at most $\varepsilon$. The bound applies in particular to Gaussian inputs.
\end{theorem}
\begin{proof}
The excess noisy risk is $(1-2\eta)P(h\ne f^*)$ and the loss-difference variance is at most $P(h\ne f^*)$. A relative VC bound therefore gives the displayed fast rate for VC dimension $d+1$; Appendix~\ref{app:halfspace_tight} details this calculation.
\end{proof}
\begin{remark}[Agnostic comparison]
For arbitrary conditional label distributions, standard VC theory supplies an excess-risk upper bound of order $\sqrt{(d+\log(1/\delta))/n}$. The fixed symmetric-noise model above uses its positive noise gap to obtain the faster clean-disagreement guarantee.
\end{remark}

\subsection{Finite-Sample Bounds}\label{sec:finite_sample}
\begin{theorem}[A high-probability local-vote bound]\label{thm:nonasymptotic_ub}
Under Theorem~\ref{thm:upper_bound}, suppose $P(G^c)\le\xi$ and $P(B(x,r))\ge a r^d$ on $G$. With probability at least $1-\delta$ over training observations,
\begin{equation}\label{eq:nonasymptotic_ub}
P(\hat f_{n,r}\ne f)\le\alpha\psi r+Lr^2+\xi+\delta^{-1}e^{-c_\eta a n r^d}.
\end{equation}
For $r_n=(\gamma_n/n)^{1/d}\le r_0$ and $\gamma_n=(c_\eta a)^{-1}\log(n^2/\delta)$,
\begin{equation}\label{eq:nonasymptotic_explicit}
P(\hat f_{n,r_n}\ne f)\le\alpha\psi(\gamma_n/n)^{1/d}
+L(\gamma_n/n)^{2/d}+\xi+n^{-2}.
\end{equation}
\end{theorem}
\begin{theorem}[Finite-family information lower bound]\label{thm:nonasymptotic_lb}
Suppose a classifier family $\mathcal F$ contains a packing of $M\ge2$ members with pairwise disagreement in $[\Delta,D]$, where $0<\Delta\le D\le1$ are its separation and diameter bounds. For independent symmetric label noise $\eta\in(0,1/2)$, let $D_\eta=(1-2\eta)\log((1-\eta)/\eta)$. Then
\[
\inf_{\hat f}\sup_{f\in\mathcal F}\E[P(\hat f\ne f)]
\ge\frac{\Delta}{2}\left(1-\frac{nD_\eta D+\log2}{\log M}\right)_+.
\]
\end{theorem}
\begin{remark}[Complementary quantities]
The upper bound uses tube mass and local sample coverage. The lower bound uses hypothesis separation, packing size, and information per observation. These quantities specify the learning family as well as its geometry.
\end{remark}
\begin{corollary}[Sufficient sample size for local voting]\label{cor:sample_size}
Assume $\psi>0$ and choose $G$ with $P(G^c)\le\varepsilon/4$. Let
\[
r_\varepsilon=\min\left\{r_0,\frac{\varepsilon}{4\alpha\psi},
\sqrt{\frac{\varepsilon}{4L}}\right\},
\]
with the last term interpreted as $+\infty$ when $L=0$. If $P(B(x,r_\varepsilon))\ge a r_\varepsilon^d$ on $G$, then
\[
n\ge\frac{1}{c_\eta a r_\varepsilon^d}\log\frac{4}{\delta\varepsilon}
\]
ensures $P(\hat f_{n,r_\varepsilon}\ne f)\le\varepsilon$ with probability at least $1-\delta$. In particular, the geometric sample-size contribution is proportional to $(\alpha\psi/\varepsilon)^d$, with the coverage constant and remaining terms shown explicitly.
\end{corollary}

For halfspace classifiers under Gaussian inputs and $\eta\in(0,1/2)$, we can
identify the geometric consequence of the population logistic minimizer in
$\psieff$ terms.
We use the affine-intercept convention $h_{w,b}(x)=\1\{w^\top x+b\ge0\}$ throughout. For $Y=f^*(X)\oplus\xi$, with independent $\xi\sim\Ber(\eta)$ and signed label $\tilde Y=2Y-1$, define $\mathcal L(w,b)=\E\log(1+e^{-\tilde Y(w^\top X+b)})$. The main result is:

\begin{theorem}[Supplementary observation: population logistic minimizer and $\psieff$]
\label{thm:gf}
Let $P=\mathcal{N}(0,I_d)$, $w_*\in\mathcal S^{d-1}$, $b_*\in\R$, and $f^*=\1\{w_*^\top x+b_*\ge0\}$, and $\eta\in(0,1/2)$.
The population logistic loss has a unique finite minimizer $(\hat w,\hat b)$.
\begin{enumerate}[label=(\roman*),leftmargin=*,itemsep=1pt]
\item $\psieff(h_{w,b},P)=\mathfrak g(b/\|w\|)$ with the Gaussian profile $\mathfrak g$ from Proposition~\ref{prop:margin}.
\item $\hat w=\hat a w_*$ for a finite $\hat a>0$.
\item For $b_*=0$: $\hat b=0$ and $\hat h=h_{\hat w,0}$, giving $\psieff(\hat h,P)=\mathfrak g(0)=2\sqrt{2/\pi}$
  (the global isoperimetric minimum).
\end{enumerate}
This result follows from strict convexity and Gaussian symmetry for the halfspace model.
Full proof in Appendix~\ref{app:gf_sec}.
\end{theorem}

\begin{remark}[Interpretation]
Theorem~\ref{thm:gf} characterizes Gaussian halfspaces.
The explicit gradient (Proposition~\ref{prop:gradient}) provides a direct
optimization route for any architecture, bypassing the implicit-bias question entirely.
\end{remark}

%=======================================================================
\subsection{Differentiable Surrogate for Deep Models}\label{app:ext_surrogate}
\label{sec:surrogate}
%=======================================================================

Proposition~\ref{prop:gradient} gives the exact gradient $\nabla_W\psieff$
for linear representations $T_W(x)=Wx$.  In deep learning, the feature map
is nonlinear, motivating a sample-based measurement of the decision boundary.
This section defines a smooth surrogate $\hat\psi_{\tau,\rho}$ for the
fixed-backbone linear-head setting.  The extra parameter $\rho>0$ smooths the
class-balance term and removes the non-differentiability caused by the raw
minimum operation.

\subsubsection{Definition of the smooth surrogate}

For temperature $\tau>0$, write $\sigma_\tau(u):=\sigma(u/\tau)$ for the scaled sigmoid.
Given diagnostic feature vectors $Z_1,\ldots,Z_N\in\R^m$ with law $Q$,
weight vector $w\in\R^m\setminus\{0\}$, bias $b\in\R$, tube width $\varepsilon_0>0$,
temperature $\tau>0$, and balance-smoothing parameter $\rho>0$.

\begin{definition}[Differentiable $\psieff$ surrogate]
\label{def:surrogate}
Define the signed distance $s_i:=(w^\top Z_i+b)/\|w\|$ and the smooth minimum
\begin{equation}
\label{eq:smin}
  m_\rho(u):=\frac{1-\sqrt{(2u-1)^2+\rho^2}+\rho}{2},
  \qquad u\in[0,1].
\end{equation}
Then the smooth surrogate is
\begin{equation}
\label{eq:surrogate}
  \hat\psi_{\tau,\rho}(w,b;\{Z_i\}):=
  \frac{\widehat B_{\varepsilon_0,\tau}(w,b)}
  {\varepsilon_0\cdot m_\rho\!\left(\hat u_\tau(w,b)\right)},
\end{equation}
where:
\begin{align}
  \widehat B_{\varepsilon_0,\tau}(w,b)&:=\frac{1}{N}\sum_{i=1}^N\sigma_\tau(\varepsilon_0-|s_i|),
  \label{eq:tube_smooth}\\
  \hat u_\tau(w,b)&:=\frac{1}{N}\sum_{i=1}^N\sigma_\tau(s_i).
  \label{eq:class_smooth}
\end{align}
\end{definition}

The identity
\[
  \min(u,1-u)=\frac{1-|2u-1|}{2}
\]
shows that $m_\rho(u)\to \min(u,1-u)$ uniformly on $[0,1]$ as $\rho\to 0$.
Thus $\hat\psi_{\tau,\rho}$ is a smooth approximation to the hard estimator.
The quantity $\widehat B_{\varepsilon_0,\tau}/\varepsilon_0$ approximates the tube probability divided by width, $Q(|s(Z)|\le\varepsilon_0)/\varepsilon_0$, where $Z\sim Q$, $s(z)=(w^\top z+b)/\|w\|$, and $\partial h=\{z:s(z)=0\}$.

\subsubsection{Consistency and smoothness}

\begin{proposition}[Surrogate consistency and smoothness]
\label{prop:surrogate}
Let $Z_1,\ldots,Z_N$ be independent diagnostic samples from $Q$ with bounded density $q$, and let $h=h_{w,b}=\1\{w^\top z+b\ge0\}$ with $w\ne0$. Put $s(z)=(w^\top z+b)/\|w\|$ and $\partial h=\{z:s(z)=0\}$; fix $\varepsilon_0>0$.
\begin{enumerate}[label=(\roman*),leftmargin=*,itemsep=2pt]
\item \textbf{(Smoothness)} For any $\tau,\rho>0$ and any sample $\{Z_i\}$,
  $\hat\psi_{\tau,\rho}$ is $C^\infty$ in $(w,b)$ on every open region of
  $\R^m\setminus\{0\}\times\R$ where each $s_i(w,b)\ne0$. Globally it is locally Lipschitz and differentiable almost everywhere; a fully smooth variant replaces $|s_i|$ by $(s_i^2+\nu^2)^{1/2}$.

\item \textbf{(Hard-limit consistency)} Let
  \[
    \widehat{\psi}_{\mathrm{eff},\varepsilon_0}^{\mathrm{hard}}(w,b;\{Z_i\})
    :=\frac{\frac{1}{N}\sum_i\1\{|s_i|\le\varepsilon_0\}}
    {\varepsilon_0\cdot\min(\hat u,1-\hat u)},
    \qquad
    \hat u:=\frac{1}{N}\sum_i\1\{s_i\ge0\}.
  \]
  If $0<\hat u<1$ and every sample satisfies $|s_i|\ne\varepsilon_0$ and $s_i\ne0$, then $\hat\psi_{\tau,\rho}(w,b;\{Z_i\})\to\widehat{\psi}_{\mathrm{eff},\varepsilon_0}^{\mathrm{hard}}(w,b;\{Z_i\})$
  as $\tau,\rho\to 0$.

\item \textbf{(Population consistency)} Under mild regularity (bounded density $q$
  near the boundary and $\min(Q(h_{w,b}=1),Q(h_{w,b}=0))>0$): as
  $N\to\infty$ and then $\tau,\rho\to 0$,
  \begin{equation}
  \label{eq:consistency}
    \hat\psi_{\tau,\rho}(w,b;\{Z_i\})\xrightarrow{\,P\,}
    \frac{Q(\operatorname{dist}(Z,\partial h)\le\varepsilon_0)}
    {\varepsilon_0\cdot\min(Q(h=1),Q(h=0))}.
  \end{equation}
  If the slice integral $r_s(t)=\int_{\{z:s(z)=t\}}q(z)\,d\mathcal H^{m-1}(z)$ is finite and continuous at zero, shrinking $\varepsilon_0$ afterward gives $2r_s(0)/\min(Q(h=1),Q(h=0))=\psieff(h_{w,b},Q)$ by one-dimensional integration.
\end{enumerate}
\end{proposition}

\begin{proposition}[Finite-width and smoothing bias]\label{prop:surrogate_bias}
Suppose the signed-distance slice integral $r_s$ is a bounded density and is Lipschitz near zero, and has class balance $\beta=\min(Q(h=1),Q(h=0))>0$. For sufficiently small positive $\varepsilon_0,\tau,\rho$ with $\tau\le\varepsilon_0$ and $\tau+\rho\le c\beta$, define the population surrogate $\psi_{\tau,\rho,\varepsilon_0}=\E_Q[\sigma((\varepsilon_0-|s(Z)|)/\tau)]/(\varepsilon_0m_\rho(\E_Q[\sigma(s(Z)/\tau)]))$. It obeys
\[
|\psi_{\tau,\rho,\varepsilon_0}-\psieff(h,Q)|
\le C\left(\frac{\varepsilon_0}{\beta}
+\frac{\tau}{\varepsilon_0\beta}+\frac{\tau+\rho}{\beta^2}\right),
\]
where $C$ depends on the marginal density bounds. For fixed positive balance this has order $O(\varepsilon_0)+O(\tau/\varepsilon_0)+O(\rho)$.
\end{proposition}
\begin{proof}
Integrating the marginal near zero gives $Q(|s|\le\varepsilon_0)/\varepsilon_0=2r_s(0)+O(\varepsilon_0)$. The integral of the difference between a sigmoid and a step is $O(\tau)$, so tube smoothing changes its numerator by $O(\tau)$ and class-probability smoothing changes its argument by $O(\tau)$. The smooth minimum differs uniformly from the exact minimum by at most $\rho/2$. Consequently the denominator balance differs by $O(\tau+\rho)$ and remains at least $\beta/2$. The quotient inequality gives the three displayed contributions.
\end{proof}

\begin{proof}[Full proof of Proposition~\ref{prop:surrogate} in Appendix~\ref{app:surrogate}]
Part~(i): $m_\rho$ is analytic for every $\rho>0$ and composition preserves
smoothness.
Part~(ii): both sigmoid terms converge pointwise to hard indicators as
$\tau\to 0$, while $m_\rho$ converges uniformly to the exact minimum as
$\rho\to 0$.
Part~(iii): LLN controls $\widehat B_{\varepsilon_0,\tau}$ and $\hat u_\tau$, and
Dominated convergence gives the hard population limit, and continuity of the signed-distance marginal identifies the shrinking-width limit.
\end{proof}

\subsubsection{Fixed-backbone interpretation}

When the backbone $\phi:\R^d\to\R^m$ is fixed and only the linear head is
optimized, the relevant geometric object lives in feature space.

\begin{proposition}[Feature-space target for fixed-backbone linear heads]
\label{prop:bridge}
Let $\phi:\R^d\to\R^m$ be a fixed backbone and let $Q:=\phi_\#P$ be the induced
feature distribution.  For the feature-space halfspace
$h_{w,b}(z):=\1\{w^\top z+b\ge 0\}$:
\begin{enumerate}[label=(\roman*),leftmargin=*,itemsep=2pt]
\item The population quantity optimized by the surrogate is
  \[
    G(w,b;Q)
    :=\psieff(h_{w,b},Q)
    =\frac{\mup_Q(h_{w,b})}{\min(Q(h_{w,b}=1),Q(h_{w,b}=0))}.
  \]

\item If $Z_i=\phi(X_i)$, then $\hat\psi_{\tau,\rho}(w,b;\{Z_i\})$ is a sample
  estimator of $G(w,b;Q)$ and, under the conditions of
  Proposition~\ref{prop:surrogate}, converges in probability to $G(w,b;Q)$ as
  $N\to\infty$, then $\tau,\rho\to 0$, and then $\varepsilon_0\to 0$.

\item In the balanced zero-bias case, if $q$ is bounded and continuous near
  $\{z:w^\top z=0\}$, then
  \[
    G(w,0;Q)=4\int_{\{w^\top z=0\}} q(z)\,d\mathcal{H}^{m-1}(z).
  \]
\end{enumerate}
\end{proposition}

\begin{proof}[Full proof in Appendix~\ref{app:bridge}]
Part~(i) is the definition of the feature-space target.  Part~(ii) is
Proposition~\ref{prop:surrogate} applied to the feature samples
$Z_i=\phi(X_i)$.  Part~(iii) is the balanced specialization of
Definition~\eqref{eq:def}.
\end{proof}

\begin{remark}[Relation to Proposition~\ref{prop:gradient}]
When $\phi(x)=x$ and $Q=P$, Proposition~\ref{prop:bridge} says that the smooth
surrogate targets the same feature-space quantity that appears in the exact
linear-representation analysis of Proposition~\ref{prop:gradient}.  The former
is useful for fitting a linear head on fixed features; the latter gives an
exact gradient when the representation itself is linear and explicitly modeled.
\end{remark}

\begin{remark}[Multi-class linear head]
For $K$-class one-vs-rest with linear heads $(w_k,b_k)_{k=1}^K$, write $\hat\psi_{\tau,\rho}^{(k)}$ for the binary surrogate of head $k$:
apply Proposition~\ref{prop:bridge} per class.  The total surrogate
$\hat\psi_{\tau,\rho}:=\sum_k\hat\psi_{\tau,\rho}^{(k)}$ is differentiable in
each $w_k$ and targets the summed per-class feature-space complexity used in
Experiment~D1.
This is the implementation used in Experiment~D1.
\end{remark}

\paragraph{Intermediate-feature transfer and parameter-efficient fine-tuning.}
Selected or concatenated intermediate features, as in Head2Toe~\citep{evci2022head2toe}, form a fixed composite representation. Fitting a linear head on these features gives the same signed-distance and boundary-tube estimator. The pushforward law describes transformations between representations when the map is a smooth diffeomorphism on the relevant data support. Dimension-changing feature selection and projection admit a feature-space diagnostic directly, with fiber geometry described by the coarea formulation in Theorem~\ref{thm:submersion_push}.

For adaptive methods such as LoRA~\citep{hu2022lora}, the diagnostic can be evaluated throughout training. At iteration $t$, temporarily fix the adapted model, cache its features with distribution $Q_t$, retain or refit the downstream linear head $h_t$, and compute
\[
\hat\psi_t=\hat\psieff(h_t,Q_t).
\]
The resulting trajectory tracks boundary complexity during adaptation. For representations at nearby training iterations related by a smooth diffeomorphic feature transformation, the pushforward law describes representation transport; changes in the classification head contribute boundary motion. The end-to-end analysis in Appendix~\ref{app:ext_endtoend} provides an application of this measurement protocol across training stages.

\subsubsection{Pseudocode for the differentiable surrogate}

\begin{center}
\begin{minipage}[t]{0.56\linewidth}
\small
\refstepcounter{algorithm}\label{alg:smooth-surrogate}
\textbf{Algorithm~\thealgorithm: Smooth boundary-mass surrogate for one linear head}
\begin{algorithmic}[1]
\Require feature matrix $Z\in\R^{N\times m}$ with rows $z_i^\top$, head $(w,b)$, tube width $\varepsilon_0$, smoothing $\tau$, balance smoothing $\rho$
\Ensure smooth estimate $\hat\psi_{\tau,\rho}$
\State $r \gets \lVert w\rVert_2$
\For{$i=1,\ldots,N$}
  \State $s_i \gets (w^\top z_i+b)/r$
  \State $a_i \gets \sigma((\varepsilon_0-|s_i|)/\tau)$
  \State $q_i \gets \sigma(s_i/\tau)$
\EndFor
\State $\widehat B_{\varepsilon_0,\tau} \gets N^{-1}\sum_i a_i$
\State $\hat u_\tau \gets N^{-1}\sum_i q_i$
\State $\hat\beta_{\tau,\rho} \gets \{1-\sqrt{(2\hat u_\tau-1)^2+\rho^2}+\rho\}/2$
\State \Return $\widehat B_{\varepsilon_0,\tau} /(\varepsilon_0\hat\beta_{\tau,\rho})$
\end{algorithmic}
\end{minipage}\hfill
\begin{minipage}[t]{0.39\linewidth}
\small
Algorithm~\ref{alg:smooth-surrogate} is the differentiable version of the finite boundary-tube count used in the experiments.
The signed distance $s_i$ measures distance to the linear-head boundary after normalizing by $\lVert w\rVert_2$.

\medskip
The tube weight $a_i$ softly counts samples inside the boundary slab, and $q_i$ estimates the positive-class mass used in the balance denominator.
\end{minipage}
\end{center}

All operations are differentiable for $w\ne0$ away from sample-boundary collisions $s_i=0$, and gradients flow through the sigmoid functions. For multi-class one-vs-rest heads, the implementation applies the same routine to each class head and sums the returned estimates.

The surrogate $\hat\psi_{\tau,\rho}$ is smooth on these open regions and is differentiable almost everywhere globally. Ordinary minibatch training uses direct differentiation of the sigmoid terms; a fully smooth implementation can replace $|s_i|$ by $(s_i^2+\nu^2)^{1/2}$. In practice we use a small fixed $\rho$ and suppress it in plots and tables when the notation is clear.

%=======================================================================
\subsection{Gradient Formula (Linear Representation)}
\label{sec:grad}
%=======================================================================

\begin{proposition}[Gradient of $\psieff$ w.r.t.\ linear representation]
\label{prop:gradient}
Let $T_W(x)=Wx$, $W\in GL(d)$, $f=\1_A$ with $C^1$ boundary, $P=\mathcal{N}(0,I_d)$,
$P(A)=1/2$ and $\int_{\partial A}p\,d\Hd<\infty$. Then $\psi_W:=\psieff(f\circ T_W^{-1},(T_W)_\#P)$ satisfies:
\begin{equation}
\label{eq:grad}
  \nabla_W\psi_W
  =-4\int_{\partial A}p(x)\,
  \frac{(W^{-\top}n(x))\,[(W^\top W)^{-1}n(x)]^\top}
       {|W^{-\top}n(x)|}
  \,d\Hd(x),
\end{equation}
a $d\times d$ matrix (a $d$-dimensional outer product at each boundary point).
\end{proposition}

\begin{proof}[Full proof in Appendix~\ref{app:grad_proof}]
Class balance preserves: $\psi_W=4\int_{\partial A}p|W^{-\top}n|d\Hd$.
Using $\partial(W^{-\top})_{kl}/\partial W_{ij}=-(W^{-\top})_{kj}(W^{-\top})_{il}$
(implicit differentiation) gives the matrix gradient of $|W^{-\top}n|$, then
interchange of integration and differentiation completes the proof.
\end{proof}

%=======================================================================
\subsection{Supplementary Notes on Optimization Dynamics}
\label{sec:implicit_bias}
%=======================================================================

The gradient formula (Proposition~\ref{prop:gradient}) enables explicit
$\psieff$ optimization.  A separate question is whether common optimization
dynamics reduce $\psieff$ implicitly, under the unregularized objective.
The results in this section examine population logistic flow and the neural tangent kernel regime.

\subsubsection{Supplementary population gradient flow on linear classifiers}

\begin{theorem}[Boundary mass along population logistic gradient flow]\label{thm:gf_psieff}
Let $X\sim\mathcal N(0,I_d)$, $f^*(x)=\1\{w_*^\top x\ge0\}$ with $\|w_*\|=1$, and independent symmetric label noise $\eta\in(0,1/2)$. For parameters $\theta_t=(w_t,b_t)$ and population logistic gradient flow $\dot\theta_t=-\nabla\mathcal L(\theta_t)$ from any finite initialization, the unique minimizer is $(\hat w,0)$ with $\|\hat w\|>0$. The following hold:
\begin{enumerate}[label=(\roman*),leftmargin=*,itemsep=2pt]
\item The parameters converge to $(\hat w,0)$, and for all sufficiently large $t$ the boundary complexity is defined and converges to $\mathfrak g(0)=2\sqrt{2/\pi}$.
\item There is $T_0<\infty$ such that $\psieff(h_{w_t,b_t},P)$ is nonincreasing for $t\ge T_0$.
\item For $\delta_t=\psieff(h_{w_t,b_t},P)-\mathfrak g(0)$, constants $c>0$ and $T_0$ can be chosen so that
$\dot\delta_t\le-c\delta_t$ and $\delta_t\le\delta_{T_0}e^{-c(t-T_0)}$ for $t\ge T_0$.
\end{enumerate}
\end{theorem}
\begin{proof}[Proof sketch]
Coercivity and the positive-definite logistic Hessian give convergence on the compact initial sublevel set. Symmetry yields $\partial_b\mathcal L(w,b)=b\int_0^1\partial_{bb}\mathcal L(w,sb)\,ds$. Near $(\hat w,0)$ this integral has a positive lower bound, while $d\log\|w_t\|/dt\to0$. Thus $|b_t|/\|w_t\|$ eventually decreases exponentially. The one-sided derivative $\mathfrak g'(0^+)=4/\pi$ transfers the differential inequality to $\delta_t$. Appendix~\ref{app:gf_psieff} gives the full argument.
\end{proof}
\begin{remark}[Population dynamics]
The result connects the positive noise gap, Gaussian symmetry, and a locally Lipschitz boundary diagnostic. The normal-offset ratio supplies the link between logistic optimization and boundary mass.
\end{remark}

\subsubsection[Supplementary lazy-training discussion: neural tangent kernel-induced psieff control]{Supplementary lazy-training discussion: neural tangent kernel (NTK)-induced $\psieff$ control}

\begin{proposition}[Boundary stability for a regular finite-dimensional kernel flow]\label{prop:lazy_psieff}
Let $Q$ have $C^1$ density $q$ on a compact feature domain $\Omega\subset\R^m$. Let $V\subset C^2(\overline\Omega)$ be a fixed finite-dimensional function space on which $\|\cdot\|_{L^2(Q)}$ is a norm. Consider fixed-kernel squared-loss flow
\[
\partial_tF_t=-\mathcal K(F_t-F^*),\quad F_0,F^*\in V,
\]
where $\mathcal K:V\to V$ is self-adjoint and has eigenvalues at least $\lambda_0>0$ in $L^2(Q)$. Assume $\Sigma_*=(F^*)^{-1}(0)$ is a compact regular $C^2$ hypersurface inside $\Omega$, $\|\nabla F^*\|\ge a_0>0$ on a tubular neighborhood, and $|F^*|$ is bounded below by a positive constant outside that neighborhood. Assume both limiting classes have probability at least $\beta_0>0$. Then:
\begin{enumerate}[label=(\roman*),leftmargin=*,itemsep=2pt]
\item $\|F_t-F^*\|_{L^2(Q)}\le e^{-\lambda_0t}\|F_0-F^*\|_{L^2(Q)}$.
\item For all sufficiently large $t$, $\partial\{F_t\ge0\}$ is a normal graph over $\Sigma_*$ and
\[
|\psieff(\1\{F_t\ge0\},Q)-\psieff(\1\{F^*\ge0\},Q)|
\le C_{\rm geom,V}\|F_t-F^*\|_{L^2(Q)},
\]
where $C_{\rm geom,V}$ depends on the displayed geometry, density, balance, and function space.
\item The boundary complexity therefore converges exponentially to its limiting value, with the same iterate-to-limit distance as in part~(i).
\end{enumerate}
\end{proposition}
\begin{proof}
The spectral bound gives part~(i). Finite dimensionality gives
$\|G\|_{C^1(\overline\Omega)}\le C_V\|G\|_{L^2(Q)}$ for $G\in V$.
Regularity of the limiting level set, the normal-graph area formula, and the positive class balance yield part~(ii). Appendix~\ref{app:lazy_psieff} provides the geometric calculation.
\end{proof}
\begin{remark}[Fixed-feature kernel interpretation]
This result applies to a finite-rank smooth feature model with a positive population kernel spectrum. It specifies the function-norm control and boundary regularity that connect kernel convergence to the boundary-mass diagnostic.
\end{remark}

\subsection{Oscillatory Feature Geometry}\label{sec:case}
Let $P=\mathcal N(0,I_d)$, $X\sim P$, and $\omega>0$ be the frequency. Define $T_\omega(x)=(\sin(\omega x_1),\ldots,\sin(\omega x_d))$, and $Q_\omega=(T_\omega)_\#P$. Put $Z=T_\omega(X)$, and let $r_\omega$ be the scalar density of $\sin(\omega X_1)$.
For $w\in\mathcal S^{d-1}$, write $f_{w,\omega}(x)=\1\{w^\top T_\omega(x)\ge0\}$ and $h_w(z)=\1\{w^\top z\ge0\}$.
Surface integrals in this periodic construction use the regular boundary, whose singular remainder has zero $\Hd$ measure (Lemma~\ref{lem:patch}). For $d\ge3$ and $0<\kappa\le1/\sqrt3$, define
\[
\mathcal W_{d,\kappa}=\{w\in\mathcal S^{d-1}:\#\{i:|w_i|\ge\kappa\}\ge3\}.
\]
\begin{theorem}[Geometric reduction for controlled normal directions]\label{thm:family_reduction}
For $\omega\ge\omega_0(d)=\sqrt{2\log(16d)}$, there are positive constants depending on $d,\kappa$ such that, uniformly for $w\in\mathcal W_{d,\kappa}$,
\begin{enumerate}[label=(\roman*),leftmargin=*,itemsep=2pt]
\item $c_d\omega\le\psieff(f_{w,\omega},P)\le C_d\omega$;
\item $c_d'\le\psieff(h_w,Q_\omega)\le C_\kappa'$;
\item the feature map $T_\omega$ is shared by every such $w$;
\item for the fixed three-coordinate direction $\bar w=(1,1,1,0,\ldots,0)/\sqrt3$, the ratio is $\Theta(\omega)$ with constants independent of the ambient dimension. In particular, $\omega=e^d$ gives ratio $\Theta(e^d)$ for this embedded family.
\end{enumerate}
\end{theorem}
\begin{proof}
The periodic zero set has upper and lower surface-area bounds (Lemma~\ref{lem:patch}); periodized Gaussian density contributes the factor $\omega$ (Lemma~\ref{lem:poisson}). Three quantitatively nonzero coefficients give a bounded convolution density at zero (Lemma~\ref{lem:density_bound}). For $\bar w$, integration over the remaining coordinates is exactly one. Appendix~\ref{app:family} gives the calculations.
\end{proof}
\begin{remark}[Two-coordinate convolution]
For $w=(1,1,0,\ldots,0)/\sqrt2$, the endpoint behavior of $r_\omega$ gives
$p_{w^\top Z}(0)=\sqrt2\int_{-1}^1r_\omega(z)^2\,dz=+\infty$.
The three-coordinate condition in $\mathcal W_{d,\kappa}$ provides integrable convolution control and an explicit uniform upper constant.
\end{remark}
\begin{corollary}[Geometry and the specified learning family]\label{cor:sc_gap}
Let $\mathcal F_\omega=\{f_{w,\omega}:w\in\mathcal S^{d-1}\}$ with known $\omega$. Its VC dimension is at most $d+1$, and empirical-risk minimization over this feature family has the fixed-noise sample bound in Theorem~\ref{thm:halfspace_tight}. The same family is a linear-head family in represented coordinates. Theorem~\ref{thm:family_reduction} quantifies its geometric boundary reduction. Theorem~\ref{thm:minimax_family} gives a dimension-dependent packing lower bound; Theorem~\ref{thm:minimax_local} separately constructs an interval-modulated family with a frequency-dependent lower bound.
\end{corollary}
\begin{proof}
Any dichotomy on inputs $x_i$ is a halfspace dichotomy on the fixed points $T_\omega(x_i)$. Thus the VC dimension is at most $d+1$ in either coordinate description. Apply the relative VC argument of Theorem~\ref{thm:halfspace_tight}.
\end{proof}

\subsection{Experiments: Measuring Representation-Induced Statistical Difficulty}\label{app:ext_experiments}
\label{sec:exp}
%=======================================================================

\paragraph{Experimental thesis.}
The experiments test $\psieff$ as a measurable, representation-sensitive
statistical constant. They address four questions: whether the quantity measures
the pushforward geometry predicted by the theorem, predicts sample efficiency,
experiments frozen representations through cached features, and supports controlled
interventions on the geometry it measures.

\subsubsection{Controlled verification of the pushforward law}
Theorem~\ref{thm:pushforward} says that the representation changes boundary
complexity through $\lVert DT^{-\top}n\rVert$. This central geometric claim is
tested in a setting where the density, boundary, and map are known. We construct a
two-dimensional distribution with a known linear boundary, apply controlled
representations, and compare the analytic pushforward prediction with a
finite-sample hard-tube estimate of $\psieff$.

The sweep includes isotropic scaling, anisotropic normal and tangent scaling,
shears, boundary-orientation changes under a fixed anisotropic map, and
tube-width convergence.  Across 74 transformation/boundary cases
with 2,000,000 samples per case, the measured value matches the
prediction with median relative error 0.38\%,
mean relative error 0.50\%, and maximum relative
error 2.58\%.  The orientation curve is the key
stress test: the same anisotropic map produces different complexity depending
on the boundary normal, exactly as $\lVert DT^{-\top}n\rVert$ predicts.

\begin{figure}[t]
\centering
\begin{minipage}{0.49\textwidth}
\includegraphics[width=\textwidth]{fig03a_fig08a_predicted_vs_measured.png}
\end{minipage}\hfill
\begin{minipage}{0.49\textwidth}
\includegraphics[width=\textwidth]{fig03b_fig08b_normal_orientation.png}
\end{minipage}\\[4pt]
\begin{minipage}{0.49\textwidth}
\includegraphics[width=\textwidth]{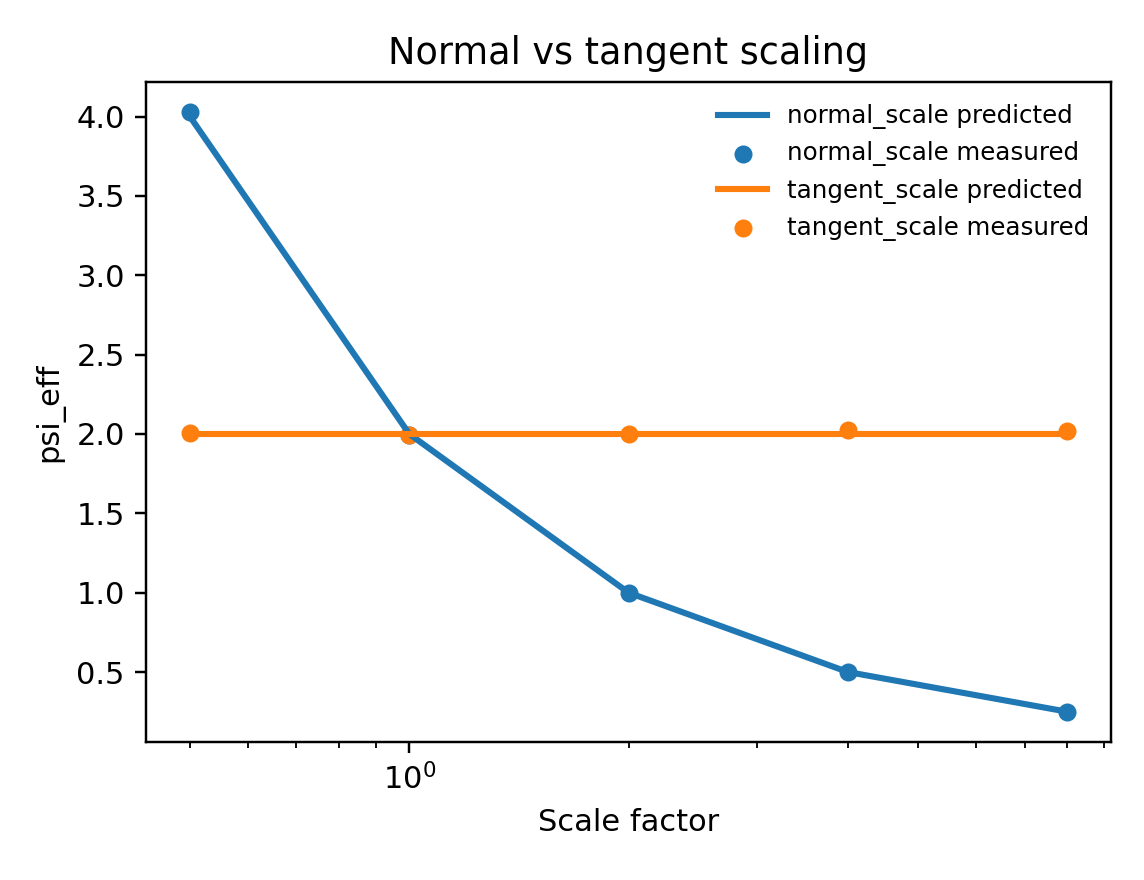}
\end{minipage}\hfill
\begin{minipage}{0.49\textwidth}
\includegraphics[width=\textwidth]{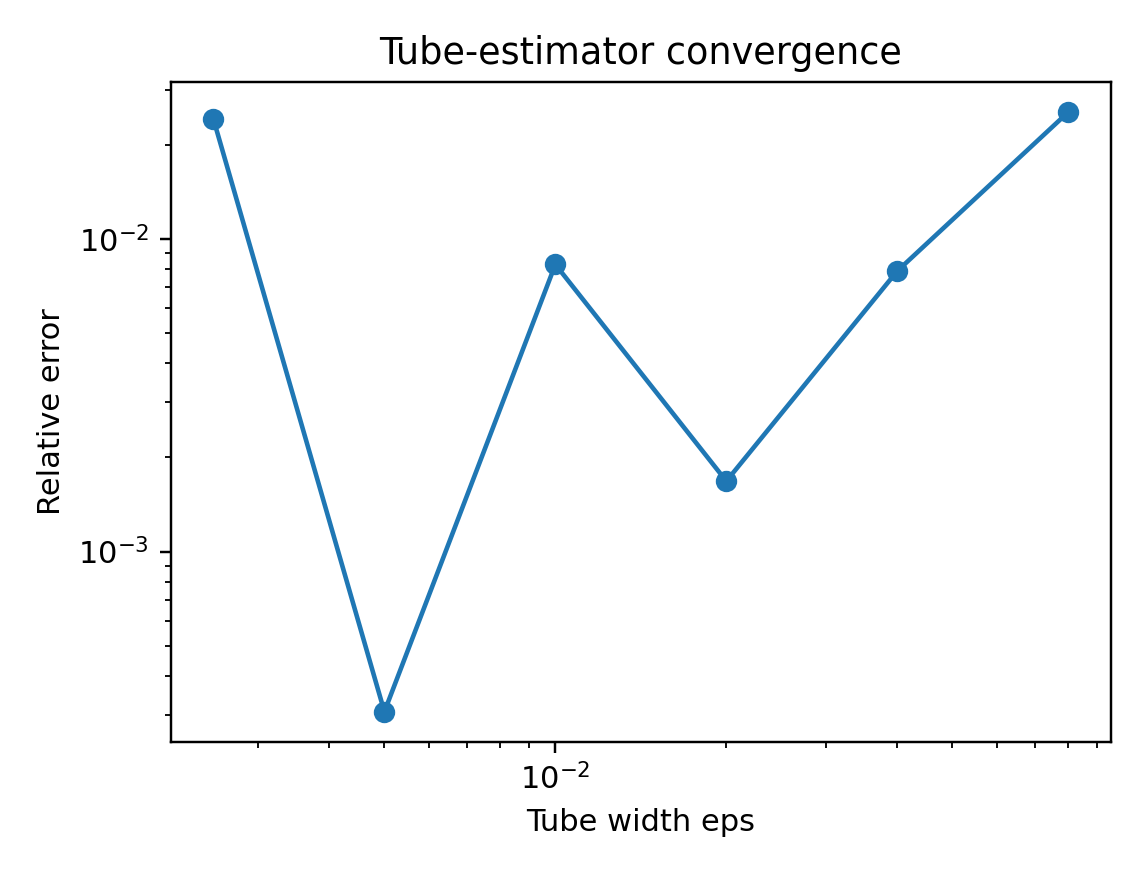}
\end{minipage}
\caption{Controlled pushforward validation.  Predicted $\psieff$ and measured
tube estimates agree across scalings, shears, and boundary orientations.  The
orientation plot validates the boundary-normal mechanism
$\lVert DT^{-\top}n\rVert$ directly.}
\label{fig:pushforward_validation}
\end{figure}

\subsubsection{Sample-complexity collapse}
The statistical claim is that $\psieff$ behaves like a difficulty constant.  To
test this, we build representations of the same binary problem that differ in
the boundary complexity they present to the learner.  A local learner sees
different raw learning curves because the representations induce different
effective boundary mass.  If $\psieff$ is the right constant, those curves should
collapse after normalizing sample size by $\psieff$.

That is what happens.  In the full run, fitting log-error using raw sample size
gives $R^2=0.698$.  Replotting the same
points against $n/\psieff$ raises the fit to
$R^2=0.895$, with a log-log slope of $-0.503$ (magnitude $0.503$). In two dimensions the theory predicts a decay exponent of $1/2$; the measured slope magnitude agrees with that exponent.  This supports the interpretation of $\psieff$ as a sample-efficiency
constant in this controlled regime.

\begin{figure}[t]
\centering
\begin{minipage}{0.49\textwidth}
\includegraphics[width=\textwidth]{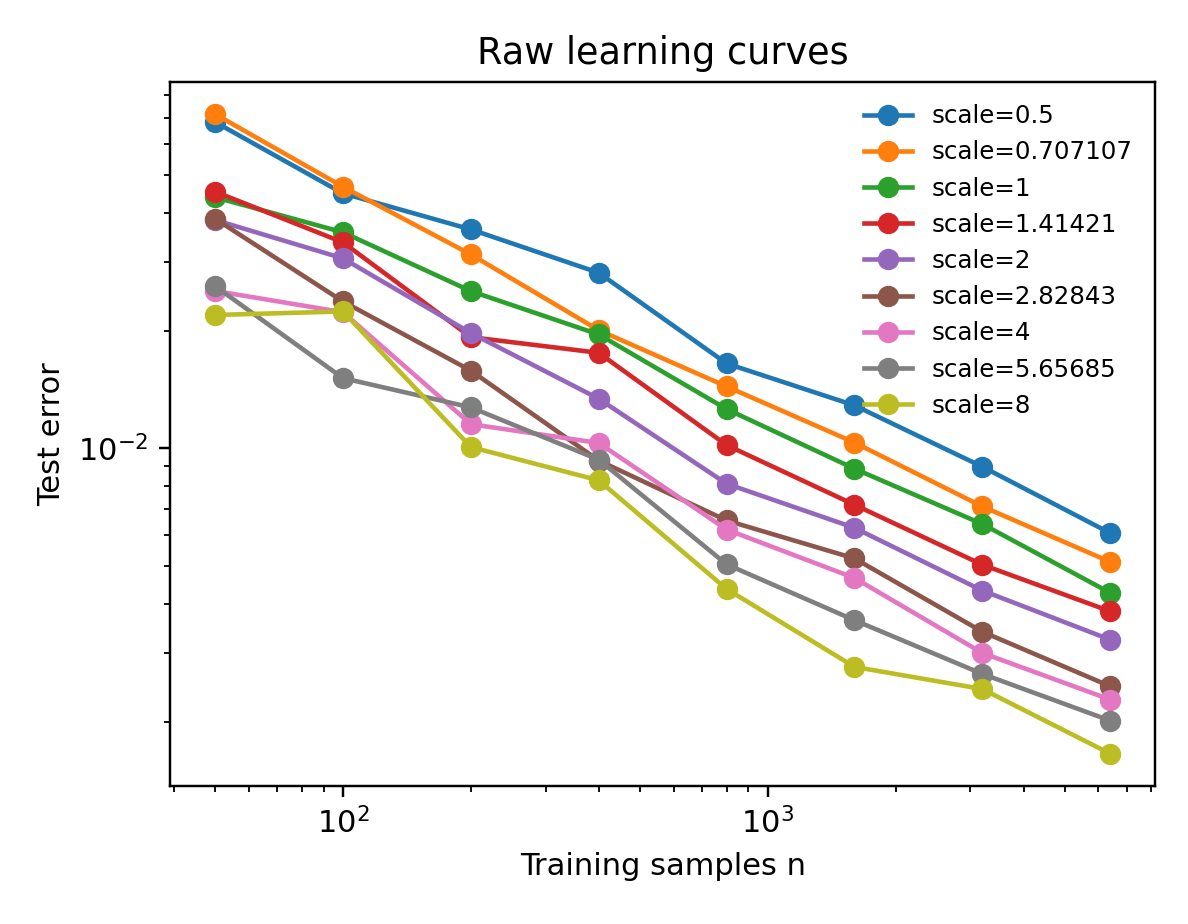}
\end{minipage}\hfill
\begin{minipage}{0.49\textwidth}
\includegraphics[width=\textwidth]{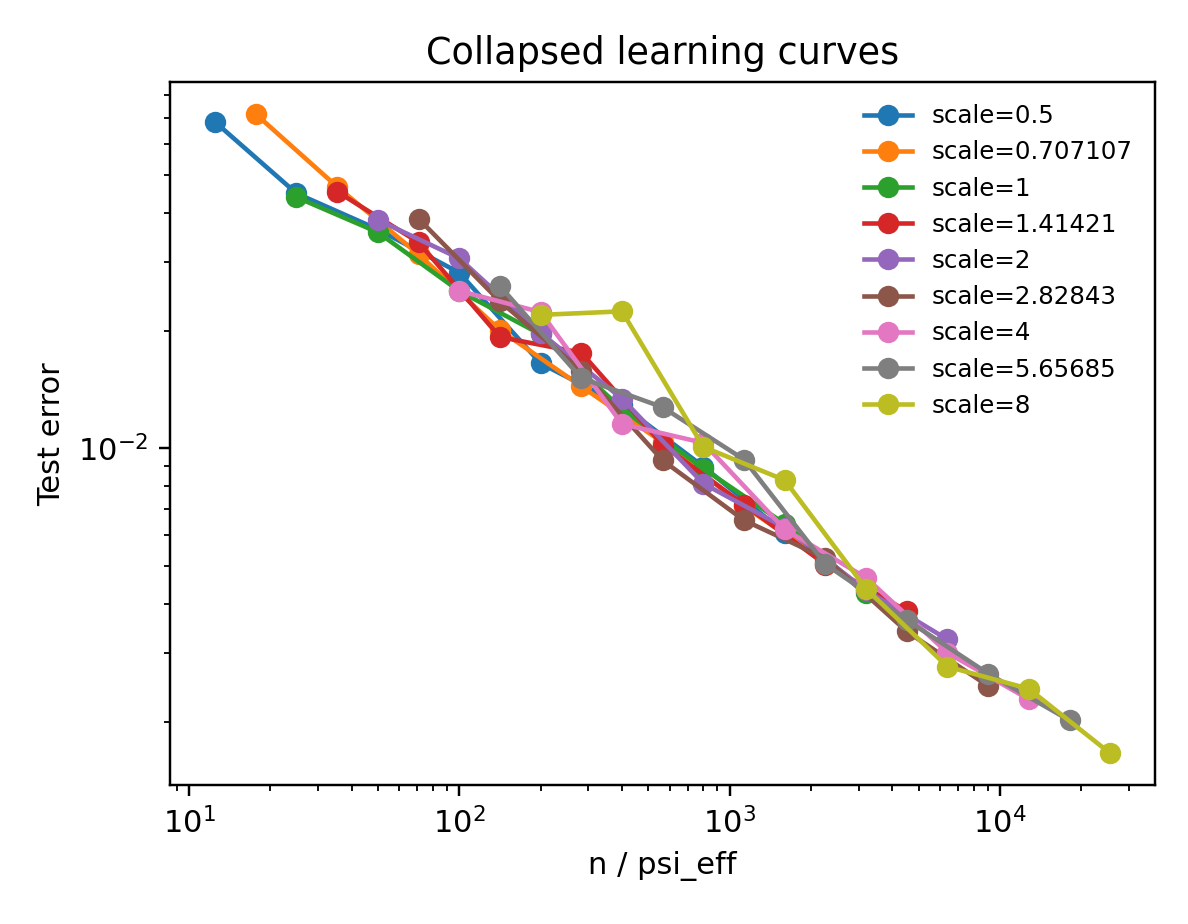}
\end{minipage}
\caption{Sample-complexity collapse.  Different representations produce
different raw learning curves.  Normalizing by $n/\psieff$ collapses the curves
and recovers the expected two-dimensional exponent.}
\label{fig:sample_collapse}
\end{figure}

\subsubsection{Boundary--Utility Plane for frozen-feature learning}
In realistic frozen-feature experiments, $\psieff$ should be treated as a state
variable.  We therefore plot training trajectories in a Boundary--Utility Plane:
$\log(1+\hat\psieff)$ on one axis and downstream accuracy on the other.  This is
a diagnostic plot for boundary geometry and utility.

We mined existing frozen-backbone runs and obtained 30,004
epoch records from 254 runs. A single mixed final-model correlation
between $\log(1+\hat\psieff)$ and utility is 0.044, so the plot is interpreted
as a grouped state-plane diagnostic. The useful analysis is grouped: within fixed
sample-per-class regimes the correlations are often much stronger (median
within-regime correlation $-0.521$ when computable), and the trajectory slopes
separate CE, weight decay, and $\psieff$ intervention as different geometric
pressures. The recovered BoundaryEfficiency statistic,
$\mathrm{utility}/\log(1+\hat\psieff)$, is highest on average for the
$\psieff$-intervention runs among the available modes, consistent with a
state-plane diagnostic for representation comparison.

\begin{figure}[t]
\centering
\includegraphics[width=0.92\textwidth]{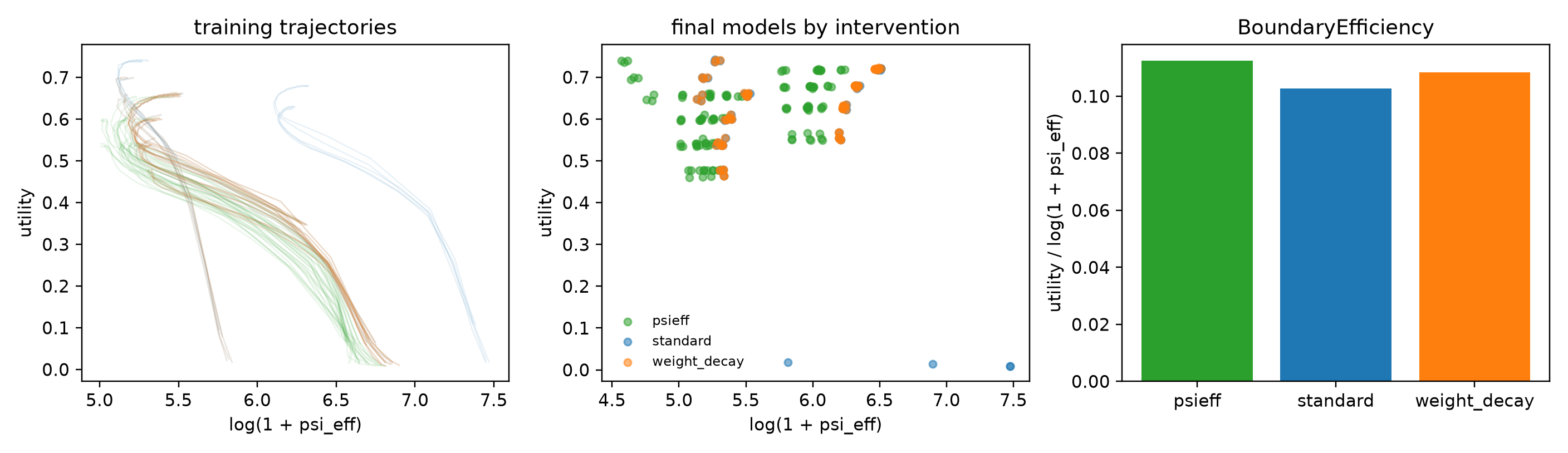}
\caption{Grouped Boundary--Utility Plane from the frozen-feature runs. The
grouping by dataset and sample regime
reveals interpretable trajectories: CE, weight decay, and $\psieff$ intervention
move through utility--boundary space with different slopes and efficiency
profiles. The plane is used as a coordinate system for training dynamics, complementing a theorem-level optimality claim.}
\label{fig:boundary_utility}
\end{figure}

\subsubsection{Evaluating pretrained representations}
The real downstream use case is representation evaluating.  Given a frozen
backbone and a downstream task, $\hat\psieff$ estimates how much boundary mass
the trained head places near the feature distribution.  The evaluation question is
which representation buys the most utility per unit boundary complexity, and
which metric predicts low-shot sample efficiency before expensive adaptation.

The proxy evaluation uses the current saved frozen-feature
runs. For 106 selected linear heads on frozen ResNet-50 features for CIFAR-100, we use each fitted head and its feature data to compute
weight norm, normalized multiclass margin, labeled-feature effective rank, and a
feature-noise flip-rate proxy. Because raw correlations are heavily confounded
by the sample-per-class budget, Table~\ref{tab:proxy_recovered} reports
correlations after demeaning both predictor and Top-1 within each spc group; the
mixed Pearson column reports the raw confounded association.

This evaluation targets the classifier-quality axis. Margin and AURC are strong final Top-1 predictors in this same-backbone
evaluation, while weight norm mostly tracks the sample budget and becomes weak after
spc adjustment. $\hat\psieff$ contributes a different readout: it measures the
boundary-geometry state that the controlled transformations and low-label experiments
connect to representation-induced label demand.

\begin{table}[t]
\centering
\small
\caption{Completed proxy evaluation on 106 selected linear heads on frozen ResNet-50 features for CIFAR-100.
All rows are computed from the fitted heads and their feature data. The main
columns use spc-residual correlations, which compare runs at a matched label
budget; the mixed Pearson column is included to show how strongly raw
correlations can be driven by sample budget.}
\label{tab:proxy_recovered}
\begin{tabular}{lcccc}
\toprule
Predictor & spc-resid. Pearson & spc-resid. Spearman & mixed Pearson & Runs\\
\midrule
$\hat\psieff$ & $0.200$ & $0.228$ & $0.369$ & 106\\
$\log(1+\hat\psieff)$ & $0.204$ & $0.236$ & $0.339$ & 106\\
Weight norm & $0.070$ & $0.127$ & $0.965$ & 106\\
Normalized margin & $0.576$ & $0.567$ & $0.996$ & 106\\
Feature effective rank & $0.442$ & $0.409$ & $0.960$ & 106\\
Noise flip rate & $-0.183$ & $-0.088$ & $-0.884$ & 106\\
NLL & $-0.529$ & $-0.508$ & $-0.996$ & 106\\
ECE & $0.315$ & $0.323$ & $-0.964$ & 106\\
AURC & $-0.749$ & $-0.743$ & $-0.995$ & 106\\
\bottomrule
\end{tabular}
\end{table}

\subsubsection{Smooth surrogate as controlled intervention}
The regularization experiments are interpreted as controlled interventions on
feature-space geometry. Weight decay is isotropic parameter pressure; $\psieff$
intervention is boundary-tube pressure. As the surrogate penalty weight $\lambda$ increases, $\hat\psieff$
decreases and accuracy stays stable until a threshold.

\begin{table}[t]
\centering
\caption{Smooth surrogate as geometry intervention on frozen ImageNet-pretrained
ResNet-50 features.  The intervention moves feature-space boundary complexity
strongly while preserving nearly the same utility, illustrating controllability
of the measured geometry.}
\label{tab:intervention_summary}
\small
\begin{tabular}{llccc}
\toprule
Dataset & spc & Standard Top-1 / $\hat\psieff$ & $\psieff$-int. Top-1 / $\hat\psieff$ & Complexity drop\\
\midrule
CIFAR-100 & 10 & $53.97\% / 201.68$ & $53.84\% / 169.22$ & $-16.1\%$\\
CIFAR-100 & 20 & $60.03\% / 215.13$ & $59.90\% / 173.80$ & $-19.2\%$\\
CIFAR-100 & 50 & $65.82\% / 246.23$ & $65.74\% / 185.57$ & $-24.6\%$\\
Tiny-ImageNet & 10 & $62.85\% / 507.04$ & $62.73\% / 387.09$ & $-23.7\%$\\
Tiny-ImageNet & 20 & $68.01\% / 558.31$ & $67.84\% / 397.38$ & $-28.8\%$\\
Tiny-ImageNet & 50 & $72.15\% / 659.58$ & $71.98\% / 420.08$ & $-36.3\%$\\
\bottomrule
\end{tabular}
\end{table}

Table~\ref{tab:intervention_summary} shows the essential effect:
$\psieff$ intervention reduces final $\hat\psieff$ by large margins while
leaving Top-1 nearly unchanged.  That is valuable because it demonstrates
controllability of the measured geometry; regularization is one way to act on
that measurement.

\paragraph{Experimental conclusion.}
The supplementary experiments provide additional evidence that $\psieff$ is
measurable, representation-sensitive, and controllable under fixed-feature
interventions. Pushforward validation shows that the geometric law is visible in
finite samples. Sample-complexity collapse shows that the quantity predicts
statistical difficulty in a controlled regime. Boundary--Utility trajectories and
smooth interventions show that the measured geometry can be monitored and moved
with the backbone fixed.

%=======================================================================
\subsection{Supplementary Discussion}
\label{sec:disc}
%=======================================================================

\paragraph{Geometric and empirical connection.}
We studied $\psieff(f,P)$ as a measurable coefficient for
representation-induced statistical difficulty.  Boundary mass in feature space
connects representation geometry, constructive tube-bias coefficients, and
practical frozen-feature diagnostics.

\paragraph{Supplementary evidence.}
The pushforward experiment validates the geometric part: $\lVert DT^{-\top}n\rVert$ is visible in finite-sample boundary measurements.  The
sample-complexity collapse supports the statistical part in a controlled regime:
normalizing by $\psieff$ explains learning curves far better than raw sample
size and recovers the expected two-dimensional exponent.  The frozen-feature
intervention results add a practical point: the smooth
surrogate can move feature-space boundary complexity with the backbone fixed.

\paragraph{Practical role.}
In low-label transfer, $\hat\psieff$ experiments whether a backbone--task pair is
likely to be sample-inefficient before expensive adaptation. During probing or
fine-tuning, the Boundary--Utility Plane shows whether extra utility is
accompanied by a more complex boundary or by a simpler representation. The same
diagnostic can be paired with calibration and perturbation sensitivity to
identify models whose decision geometry is brittle even when Top-1 looks
acceptable.

\paragraph{Practical value.}
Boundary-mass diagnostics support label-budget triage by comparing target tasks and representations before annotation-intensive adaptation. Combined with accuracy, calibration, and perturbation measurements, they provide a geometric view of the candidates' annotation demand.

\paragraph{Broader impacts.}
Boundary-mass diagnostics support allocation of annotation effort and early comparison of representation--task pairs. Overweighting an aggregate score can shift attention away from classwise errors and calibration. Reporting annotation cost together with classwise accuracy, calibration, and robustness supports decisions that account for both labeling effort and prediction quality.

\paragraph{Extensions and practical use.}
The estimator applies to fixed composite features and to measurements during adaptation, including the Head2Toe and LoRA settings described in Appendix~\ref{app:ext_surrogate}. Multiclass comparisons support OvR ranking fidelity, and acquisition-policy experiments show a stable relationship with realized label demand. Proposition~\ref{prop:surrogate_bias} and Appendix~\ref{app:compute} quantify the roles of tube width, smoothing, class balance, and diagnostic sample size in the measurement.

\paragraph{Summary.}
$\psieff$ is a measurement of representation-induced difficulty. The smooth
surrogate makes this measurement usable in modern feature spaces, and
regularization is one possible intervention on the measured geometry.

\newpage
\subsection{Proof of Proposition~\ref{prop:margin} (Margin Connection)}
\label{app:margin}
%=======================================================================

\begin{proof}[Complete proof]

\textbf{Step 1: Minkowski content of $f_a$.}
For $f_a(x)=\1\{x_1\ge a\}$ under $P=\mathcal N(0,I_d)$, write $x'=(x_2,\ldots,x_d)$ and let $\varepsilon>0$ be the tube half-width. Its probability is:
\begin{align}
  P(\{x:|x_1-a|\le\varepsilon\})
  &=\int_{\R^{d-1}}\!\int_{a-\varepsilon}^{a+\varepsilon}\gauss(t)\,dt\;\gauss_{d-1}(x')\,dx'
  =\int_{a-\varepsilon}^{a+\varepsilon}\gauss(t)\,dt,
  \label{eq:tube_fb}
\end{align}
where we used the factorization $\gauss_d(x)=\gauss(x_1)\gauss_{d-1}(x')$
and $\int_{\R^{d-1}}\gauss_{d-1}=1$.
By the fundamental theorem of calculus:
$\lim_{\varepsilon\to 0}(2\varepsilon)^{-1}\int_{a-\varepsilon}^{a+\varepsilon}\gauss(t)dt=\gauss(a)$.
Hence $\mup_P(f_a):=\lim_{\varepsilon\to 0}P(\{|x_1-a|\le\varepsilon\})/\varepsilon=2\gauss(a)$.

\textbf{Step 2: Class probabilities.}
$P(f_a=1)=P(X_1\ge a)=\Phi(-a)$ and $P(f_a=0)=\Phi(a)=1-\Phi(-a)$.
For the normalization:
\begin{equation}
\label{eq:class_probs}
  \min(P(f_a=1),P(f_a=0))=\min(\Phi(-a),\Phi(a))=\Phi(-|a|),
\end{equation}
since $\Phi(-a)\le\Phi(a)$ iff $a\ge 0$.

\textbf{Step 3: Formula.}
$\psieff(f_a,P)=\mup_P(f_a)/\Phi(-|a|)=2\gauss(a)/\Phi(-|a|)=2\gauss(|a|)/\Phi(-|a|)$
(using $\gauss(a)=\gauss(|a|)$ by symmetry).

\textbf{Step 4: Strict monotonicity for $a>0$.}
Differentiate:
\begin{align}
  \frac{d}{da}\!\left[\frac{2\gauss(a)}{\Phi(-a)}\right]
  &=\frac{2\gauss'(a)\Phi(-a)-2\gauss(a)\Phi'(-a)(-1)}{\Phi(-a)^2}
  =\frac{2\Phi(-a)(-a\gauss(a))+2\gauss(a)^2}{\Phi(-a)^2}
  \notag\\
  &=\frac{2\gauss(a)[\gauss(a)-a\Phi(-a)]}{\Phi(-a)^2}.
  \label{eq:derivative_margin}
\end{align}
We claim $\gauss(a)>a\Phi(-a)$ for all $a>0$.
Since $\Phi(-a)=\int_a^\infty\gauss(t)\,dt$ and $t>a$ throughout the domain:
\begin{equation}
\label{eq:gaussian_ineq}
  \Phi(-a)=\int_a^\infty\gauss(t)\,dt
  <\int_a^\infty\frac{t}{a}\gauss(t)\,dt
  =\frac{1}{a}\int_a^\infty t\gauss(t)\,dt
  =\frac{\gauss(a)}{a},
\end{equation}
where the final equality uses $\int_a^\infty t\gauss(t)dt=[-\gauss(t)]_a^\infty=\gauss(a)$.
Hence $a\Phi(-a)<\gauss(a)$, so~\eqref{eq:derivative_margin} is positive for $a>0$.

\textbf{Step 5: Global minimum at $a=0$.}
At $a=0$: $\gauss(0)=(2\pi)^{-1/2}$ and $\Phi(0)=1/2$.
\[
  \mathfrak g(0):=\frac{2\gauss(0)}{\Phi(0)}=\frac{2(2\pi)^{-1/2}}{1/2}=\frac{4}{\sqrt{2\pi}}=2\sqrt{\frac{2}{\pi}}.
\]

\textbf{Step 6: Asymptotic $\mathfrak g(a)\sim 2a$ as $a\to+\infty$.}
By the Mills' ratio asymptotic \citep[Eq.~7.1.23]{abramowitz1965handbook}, for $a>0$,
$\Phi(-a)=\int_a^\infty\gauss(t)dt\sim\gauss(a)/a$ as $a\to\infty$.
Hence $\mathfrak g(a)=2\gauss(a)/\Phi(-a)\sim 2\gauss(a)/(\gauss(a)/a)=2a$.

\textbf{Step 7: Regularity at the balanced cut.}
The quotient is smooth on each open half-line and continuous at zero. Its one-sided derivatives are
\[
\mathfrak g'(0^+)=4/\pi,\qquad \mathfrak g'(0^-)=-4/\pi.
\]
Thus $\mathfrak g(a)=\mathfrak g(0)+(4/\pi)|a|+O(a^2)$ near zero, giving a locally Lipschitz minimum with a cusp.
\end{proof}

%=======================================================================
\subsection{Proof of Theorem~\ref{thm:tsybakov} (Supplementary Tube-Rate Interpretation)}
\label{app:tsybakov}
%=======================================================================

\begin{proof}[Complete proof]

\textbf{Preliminary: Tubular Neighborhood Theorem.}
For a compact $C^2$ hypersurface $\partial A\subset\R^d$, the \textbf{reach}
$\varrho(\partial A):=\sup\{r>0:\text{the signed distance }d_A\text{ (positive outside }A\text{) is }C^2
\text{ on }\{|d_A|<r\}\}$ is strictly positive \citep[Thm.~1.12]{federer1969geometric}.
Throughout, let $t_0:=\varrho(\partial A)/2>0$.  The map
$\mathcal T_A:\partial A\times(-t_0,t_0)\to\R^d$, $\mathcal T_A(x,s)=x+sn(x)$,
is a $C^1$ diffeomorphism onto $\mathcal U_{t_0}(\partial A):=\{|d_A|<t_0\}$.

\textbf{Part~(i): Exact asymptotics.}
For $t\in(0,t_0)$, using the co-area formula for the function $d_A$
\citep[Theorem~3.2.22]{federer1969geometric}:
\begin{equation}
\label{eq:tube_coarea}
  P(\operatorname{dist}(X,\partial A)\le t)
  =\int_{-t}^{t}\!\int_{\{d_A(x)=s\}}p(x)\,d\mathcal{H}^{d-1}(x)\,ds.
\end{equation}
Denote $I(s):=\int_{\{d_A=s\}}p\,d\Hd$ the level-set integral.
At $s=0$: $I(0)=\int_{\partial A}p\,d\Hd=\mup_P(f)/2$.

Differentiability of $I(s)$ near $s=0$: by the $C^2$ assumption on $\partial A$,
the level sets $\{d_A=s\}$ move smoothly.  The principal curvatures
$\kappa_1(x),\ldots,\kappa_{d-1}(x)$ of $\partial A$ are uniformly bounded by
$\kappa_{\max}:=\max_{x\in\partial A,i}|\kappa_i(x)|<\infty$.
The Hausdorff measure of the level set at height $s$ satisfies:
\begin{equation}
\label{eq:level_set_area}
  \mathcal{H}^{d-1}(\{d_A=s\}\cap B)
  =\int_{\{x\in\partial A:x+sn(x)\in B\}}\prod_{i=1}^{d-1}(1-s\kappa_i(x))\,d\mathcal{H}^{d-1}(x)
  \quad\text{for any Borel }B.
\end{equation}
Hence:
\begin{align}
  I(s)
  &=\int_{\partial A}p(x+sn(x))\prod_{i=1}^{d-1}(1-s\kappa_i(x))\,d\mathcal{H}^{d-1}(x).
  \label{eq:I_s}
\end{align}
At $s=0$: $I(0)=\int_{\partial A}p(x)\,d\Hd(x)$.
The derivative: $I'(0)=\int_{\partial A}\bigl[\nabla p(x)\cdot n(x)-p(x)\sum_i\kappa_i(x)\bigr]d\Hd(x)$,
which is finite (as $p$ is $C^1$ near $\partial A$ and curvatures are bounded).
Taylor expansion: $I(s)=I(0)+O(s)$, so
\begin{align}
  P(\operatorname{dist}\le t)
  &=\int_{-t}^t I(s)\,ds
  =2t\cdot I(0)+O(t^2)
  =2t\int_{\partial A}p\,d\Hd+O(t^2)
  =\mup_P(f)\cdot t+O(t^2).
\end{align}

\textbf{Part~(ii): Two-sided tube bound.}
The positive density and nonempty regular boundary give $\mu=\mup_P(f)>0$. Part~(i) supplies $|P(\operatorname{dist}\le t)-\mu t|\le Lt^2$. Choose $t_1>0$ so that $Lt_1\le\mu/2$. Then
\[
\tfrac12\mu t\le P(\operatorname{dist}\le t)\le\tfrac32\mu t,
\qquad0<t\le t_1,
\]
which proves the two-sided statement with explicit relative constants.

\textbf{Part~(iii).} Direct from Part~(i): $C_{\mathrm{geo}}:=\lim_{t\to 0}t^{-1}P(\operatorname{dist}\le t)=\mup_P(f)$.
By Definition~\eqref{eq:def}: $\mup_P(f)=\psieff(f,P)\cdot\min(P(A),P(A^c))$.

\textbf{Part~(iv).} See Theorem~\ref{thm:tight_bilateral}.
\end{proof}

%=======================================================================
\subsection{Proof of Proposition~\ref{prop:lb}}\label{app:lb}
\begin{proof}
For the noisy observation law $\mathbb P_f$, the conditional labels agree off $\{f_0\ne f_1\}$. On this disagreement set their Bernoulli parameters are $\eta$ and $1-\eta$, giving
\[
\KL(\mathbb P_{f_0}\Vert\mathbb P_{f_1})
=P(f_0\ne f_1)(1-2\eta)\log\frac{1-\eta}{\eta}=D_\eta\Delta.
\]
Independence multiplies this quantity by $n$. Pinsker's inequality gives
$\operatorname{TV}(\mathbb P_{f_0}^{\otimes n},\mathbb P_{f_1}^{\otimes n})\le\sqrt{nD_\eta\Delta/2}\le1/2$.
Le Cam's average testing error is at least $(1-\operatorname{TV})/2\ge1/4$. Given any fitted classifier, decode by the nearest $f_j$ in $L^1(P)$. A decoding error implies disagreement at least $\Delta/2$ from the true hypothesis. Taking expectations proves the risk bound $\Delta/8$.
\end{proof}

\subsection{Proofs of Pushforward Theorems}
\label{app:push}
%=======================================================================

\begin{proof}[Complete proof of Theorem~\ref{thm:pushforward}]

\textbf{Step 1: Boundary map.}
$T:\R^d\to\R^d$ is a $C^1$ diffeomorphism, so it is a homeomorphism.
A homeomorphism maps open sets to open sets and closed sets to closed sets.
Hence $T(\overline{A})=\overline{T(A)}$ and $T(\mathrm{int}\,A)=\mathrm{int}\,T(A)$.
Therefore $\partial T(A)=\overline{T(A)}\setminus\mathrm{int}\,T(A)
=T(\overline{A})\setminus T(\mathrm{int}\,A)=T(\partial A)$.

\textbf{Step 2: Density of $Q=T_\#P$.}
By the change-of-variables formula: $Q$ has density $q$ given by
$q(z)=p(T^{-1}(z))\,|\det DT(T^{-1}(z))|^{-1}$.
Equivalently, $q(T(x))=p(x)/|\det DT(x)|$.

\textbf{Step 3: Surface-integral definition.}
The continuous density trace and the boundary identity give:
\begin{equation}
\label{eq:push_co_area}
  \mup_Q(g)=2\int_{T(\partial A)}q(z)\,d\mathcal{H}^{d-1}(z).
\end{equation}

\textbf{Step 4: Area formula.}
Apply~\eqref{eq:area} with $\Sigma=\partial A$, $u(z):=q(z)$, and diffeomorphism $T$:
\begin{align}
  \int_{T(\partial A)}q(z)\,d\mathcal{H}^{d-1}(z)
  &=\int_{\partial A}q(T(x))\,|\det DT(x)|\,
  |(DT(x)^{-\top})n(x)|\,d\mathcal{H}^{d-1}(x).
  \label{eq:area_applied}
\end{align}

\textbf{Step 5: Jacobian cancellation.}
Substitute $q(T(x))=p(x)/|\det DT(x)|$ into~\eqref{eq:area_applied}:
\begin{align}
  \int_{T(\partial A)}q\,d\mathcal{H}^{d-1}
  &=\int_{\partial A}
  \frac{p(x)}{|\det DT(x)|}\,|\det DT(x)|\,|(DT^{-\top})n|\,d\mathcal{H}^{d-1}
  \notag\\
  &=\int_{\partial A}p(x)|(DT^{-\top})n|\,d\mathcal{H}^{d-1}.
\end{align}
Hence $\mup_Q(g)=2\int_{\partial A}p(x)|(DT^{-\top})n|\,d\mathcal{H}^{d-1}$.
\end{proof}

\begin{proof}[Proof of Theorem~\ref{thm:pw}]
The global homeomorphism gives $\partial T(A)=T(\partial A)$ and disjoint open image cells $U_k=T(C_k)$. On $U_k$, the density is
$q(Tx)=p(x)/|\det DT_k(x)|$. The exceptional set $T(\partial A\cap S)$ has zero $\Hd$ measure by Lipschitz continuity. Applying the area formula on each cell gives
\[
\int_{T(\partial A\cap C_k)}q\,d\Hd
=\int_{\partial A\cap C_k}p\|DT_k^{-\top}n\|\,d\Hd.
\]
Summation proves the formula. Injectivity also gives $T^{-1}(T(A))=A$, hence preservation of class probabilities.
\end{proof}

\subsection{Supplementary Piecewise-Affine/ReLU Derivations}
\label{app:relu}
%=======================================================================

We record a full-rank cellwise specialization for piecewise-affine
representations, the architecture class underlying ReLU networks away from kink
sets.  This appendix develops cellwise pushforward and gradient calculations for full-rank piecewise-affine maps and connects them to the smooth law and cached-feature estimator.

\subsubsection{Setup and notation}

\begin{definition}[Piecewise-affine network representation]
\label{def:relu_network}
A \textbf{piecewise-affine representation} $T:\R^d\to\R^m$ is defined by:
\begin{itemize}[leftmargin=*,itemsep=2pt]
\item A \textbf{partition} $\{C_k\}_{k=1}^{J_{\rm cell}}$ of $\R^d$ into convex polyhedral cells
  (determined by the activation pattern of ReLU units), where $J_{\rm cell}$ is the number of cells.

\item On each cell: $T(x)=M_k x+b_k$ with $M_k\in\R^{m\times d}$, $b_k\in\R^m$.

\item The \textbf{singular set} $S:=\bigcup_k\partial C_k$ (hyperplane boundaries
  between cells) has $\lambda^d(S)=0$.
\end{itemize}
\noindent
Let $L$ be the layer count and $m_\ell$ the width of layer $\ell$, with $m_0=d$ and $m_L=m$. Write
$T=T^{(L)}\circ\cdots\circ T^{(1)}$, where $T^{(\ell)}(z)=\operatorname{ReLU}(W_\ell z+c_\ell)$, $W_\ell\in\R^{m_\ell\times m_{\ell-1}}$, and $c_\ell\in\R^{m_\ell}$.
Here $\operatorname{ReLU}(u)_i=\max(u_i,0)$. On cell $C_k$, let $D_k^{(\ell)}$ be the diagonal matrix of active units and put $R_k^{(\ell)}=D_k^{(\ell)}W_\ell$. Then
\[
M_k=R_k^{(L)}\cdots R_k^{(1)},\qquad
b_k=\sum_{\ell=1}^L R_k^{(L)}\cdots R_k^{(\ell+1)}D_k^{(\ell)}c_\ell,
\]
with empty products equal to the identity.
\end{definition}

\subsubsection[Cellwise psieff formula for full-rank piecewise-affine T]{Cellwise $\psieff$ formula for full-rank piecewise-affine $T$}

\begin{theorem}[Invertible piecewise-affine pushforward]\label{thm:relu_exact}
Let the finite piecewise-affine map $T:\R^d\to\R^d$ be a bi-Lipschitz homeomorphism, with $T(x)=M_kx+b_k$ and $M_k\in GL(d)$ on every open cell. Let $A$ have compact $C^1$ boundary, $p$ be continuous nearby, and $\Hd(\partial A\cap S)=0$ for the cell-boundary set $S$. Then
\begin{equation}\label{eq:relu_pushforward}
\mup_{T_\#P}(\1_{T(A)})
=2\sum_k\int_{\partial A\cap C_k}p(x)\|M_k^{-\top}n(x)\|\,d\Hd(x).
\end{equation}
\end{theorem}
\begin{proof}
Each cell restriction is a smooth diffeomorphism. Global injectivity supplies disjoint image interiors and the boundary correspondence. The assumed surface-null intersection with $S$ also has surface-null image. Theorem~\ref{thm:pw} therefore applies with $DT_k=M_k$.
\end{proof}
\begin{remark}[Density under multiple preimages]
For a general finite-to-one full-rank piecewise-affine map, the pushforward density on regular image points is
\[
q(y)=\sum_{k:y\in T(C_k)}\frac{p(M_k^{-1}(y-b_k))}{|\det M_k|}.
\]
Its boundary coefficient is $2\int_{\partial T(A)}q\,d\Hd$ whenever this surface trace is well defined. This formula explicitly includes every inverse branch and the actual image boundary.
\end{remark}

\subsubsection{Per-cell gradient formula}

We now compute $\nabla_{M_k}\psieff$ and $\nabla_{b_k}\psieff$ for each cell.

\begin{proposition}[Supplementary derivation: cell-level gradient for piecewise-affine $T$]
\label{prop:relu_grad}
Under the conditions of Theorem~\ref{thm:relu_exact}, with $P(A)=1/2$ and
$P=\mathcal N(0,I_d)$ (so class balance is preserved under $T$):
\begin{equation}
\label{eq:relu_psieff}
  \psieff(f\circ T^{-1},T_\#P)
  =4\sum_{k=1}^{J_{\rm cell}}\int_{\partial A\cap C_k}p(x)\,|M_k^{-\top}n(x)|\,d\mathcal{H}^{d-1}(x).
\end{equation}
Holding the original integration patches $\partial A\cap C_k$ fixed, the partial derivatives for each cell $k\in[J_{\rm cell}]$ are:
\begin{equation}
\label{eq:relu_grad_Mk}
  \nabla_{M_k}\psieff
  =-4\int_{\partial A\cap C_k}p(x)\,
  \frac{(M_k^{-\top}n(x))\,[(M_k^\top M_k)^{-1}n(x)]^\top}{|M_k^{-\top}n(x)|}
  \,d\mathcal{H}^{d-1}(x).
\end{equation}
The gradient with respect to the bias $b_k$ is zero: $\nabla_{b_k}\psieff=0$.
\end{proposition}

\begin{proof}[Complete proof]

\textbf{Bias gradient.}
From~\eqref{eq:relu_psieff}: $\psieff$ depends on $T_k(x)=M_kx+b_k$ only through the
Jacobian $DT_k=M_k$ (constant) and the boundary $\partial A\cap C_k$ (which lies
in the original space and is independent of $b_k$).
Since $M_k^{-\top}n(x)$ is independent of $b_k$:
$\partial\psieff/\partial(b_k)_i=0$ for all $i$.

\textbf{Matrix gradient.}
For each cell $k$, define the scalar:
$I_k(M_k):=\int_{\partial A\cap C_k}p(x)\,|M_k^{-\top}n(x)|\,d\mathcal{H}^{d-1}(x)$.
The gradient $\nabla_{M_k}I_k$ is computed exactly as in Proposition~\ref{prop:gradient}
(Appendix~\ref{app:grad_proof}), with the integration domain restricted to
$\partial A\cap C_k$:

\textbf{Step~A.}
$\partial(M_k^{-\top})_{ab}/\partial(M_k)_{ij}=-(M_k^{-\top})_{aj}(M_k^{-\top})_{ib}$
(implicit differentiation of $(M_k^{-\top})M_k^\top=I$; same calculation as in
Appendix~\ref{app:grad_proof}, Step~1).

\textbf{Step~B.}
For $a_k(x):=M_k^{-\top}n(x)$:
$\partial(a_k)_a/\partial(M_k)_{ij}=-(M_k^{-\top})_{aj}(a_k)_i$
(same calculation as Appendix~\ref{app:grad_proof}, Step~2--3).

\textbf{Step~C.}
$\partial|a_k(x)|/\partial(M_k)_{ij}=-(a_k)_i\cdot[(M_k^\top M_k)^{-1}n(x)]_j/|a_k(x)|$
(same as Appendix~\ref{app:grad_proof}, Step~4).

\textbf{Step~D.}
Integrate over $\partial A\cap C_k$ (using dominated convergence, justified by
compactness of $\partial A\cap\overline{C_k}$ and boundedness of $p$):
\begin{align}
  \nabla_{M_k}I_k
  &=\int_{\partial A\cap C_k}p(x)\nabla_{M_k}|M_k^{-\top}n(x)|\,d\mathcal{H}^{d-1}(x)
  \notag\\
  &=-\int_{\partial A\cap C_k}p(x)
    \frac{(M_k^{-\top}n)\,[(M_k^\top M_k)^{-1}n]^\top}{|M_k^{-\top}n|}
    \,d\mathcal{H}^{d-1}(x).
\end{align}
Hence $\nabla_{M_k}\psieff=4\nabla_{M_k}I_k$, giving~\eqref{eq:relu_grad_Mk}.
\end{proof}

\begin{remark}[Independence of cells]
The gradient $\nabla_{M_k}\psieff$ depends only on the boundary $\partial A\cap C_k$
within cell $k$.  Different cells contribute independently.
This enables efficient per-cell gradient computation (analogous to backpropagation).
\end{remark}

\subsubsection{Gradient for the full network: chain rule}

For a full ReLU network $T=T^{(L)}\circ\cdots\circ T^{(1)}$ with $T^{(\ell)}(z)=\operatorname{ReLU}(W_\ell z+c_\ell)$:

\begin{proposition}[Chain rule for fixed activation patterns]\label{prop:relu_network_grad}
Under Proposition~\ref{prop:relu_grad}, hold the integration patches and activation masks fixed. Let
$A_k^{>\ell}=R_k^{(L)}\cdots R_k^{(\ell+1)}$ and $A_k^{<\ell}=R_k^{(\ell-1)}\cdots R_k^{(1)}$. Then $M_k=A_k^{>\ell}D_k^{(\ell)}W_\ell A_k^{<\ell}$; let $G_k$ denote the cell contribution to $\nabla_{M_k}\psieff$ in Eq.~\eqref{eq:relu_grad_Mk}. Then the cell-interior derivative is
\begin{equation}\label{eq:network_grad}
\left.\nabla_{W_\ell}\psieff\right|_{\rm interior}
=\sum_k (A_k^{>\ell}D_k^{(\ell)})^\top G_k(A_k^{<\ell})^\top.
\end{equation}
Here products are ordered by layer, and the component derivative is
\begin{equation}\label{eq:chain_rule_M}
\frac{\partial(M_k)_{ab}}{\partial(W_\ell)_{ij}}
=(A_k^{>\ell}D_k^{(\ell)})_{ai}(A_k^{<\ell})_{jb}.
\end{equation}
\end{proposition}
\begin{proof}
For a perturbation $H$ of $W_\ell$, $dM_k[H]=A_k^{>\ell}D_k^{(\ell)}H A_k^{<\ell}$. The Frobenius identity
$\langle G_k,dM_k[H]\rangle
=\langle(A_k^{>\ell}D_k^{(\ell)})^\top G_k(A_k^{<\ell})^\top,H\rangle$
proves Eq.~\eqref{eq:network_grad}. When the integration patches also evolve, Reynolds differentiation adds their moving-facet contributions to this fixed-pattern term.
\end{proof}

\begin{remark}[Practical computation]
The fixed-pattern interior term in Eq.~\eqref{eq:network_grad} admits an efficient algorithm:
(a)~forward pass to determine activation cells $\{C_k^{\rm net}\}$ and matrices $\{M_k^{\rm net}\}$;
(b)~boundary estimation via the differentiable surrogate (Section~\ref{sec:surrogate});
(c)~backpropagation through the chain-rule term~\eqref{eq:chain_rule_M}.
The smooth surrogate used in experiments avoids explicitly differentiating moving kink facets.
\end{remark}

\subsubsection{Comparison with weight decay}

\begin{proposition}[Supplementary note: ReLU $\psieff$ vs.\ weight decay]
\label{prop:relu_vs_wd}
For a single-cell affine map $T(x)=Mx$ with $M\in GL(d)$ and $P=\mathcal{N}(0,I_d)$:
\begin{enumerate}[label=(\roman*),leftmargin=*,itemsep=2pt]
\item \textbf{Weight decay}: minimizes $\|M\|_F^2$ (Frobenius norm of weights).
  Gradient: $\nabla_M\|M\|_F^2=2M$.
\item \textbf{$\psieff$-regularization}: minimizes $\psieff(f\circ T^{-1},T_\#P)$.
  Gradient: equation~\eqref{eq:relu_grad_Mk} (anisotropic, depends on boundary geometry).
\end{enumerate}
The two regularizers act through distinct geometric quantities.
Weight decay penalizes all singular values of $M$ isotropically.
$\psieff$-regularization penalizes the singular values of $M^{-\top}$ as projected
onto the normal directions $n(x)$ of the boundary, which is anisotropic and
depends on the geometry of $\partial A$.

For an isotropic boundary-normal distribution and $M=cI$, the two objectives have explicit scale dependence: $\|M\|_F^2=dc^2$ and $\psieff(f\circ M^{-1},M_\#P)=\psieff(f,P)/c$.
\end{proposition}
\begin{proof}
The weight-decay gradient is $2M$. The boundary gradient follows from Eq.~\eqref{eq:relu_grad_Mk}; at $M=cI$, the objective follows directly from $\|(cI)^{-\top}n\|=c^{-1}$. These expressions give the stated parameter and geometric responses.
\end{proof}

\subsection[Proof of the axiomatic uniqueness theorem for psieff]{Proof of Theorem~\ref{thm:unique} (Axiomatic Uniqueness of $\psieff$)}
\label{app:uniqueness}
%=======================================================================

\begin{proof}[Complete proof]

\textbf{Step~1: (U1)--(U3) force $\kappa(x,n,p)=cp$.}

From axiom~(U1): $\mathcal C(f,P)=\int_{\partial A}\kappa(x,n(x),p(x))d\Hd/\min(P(A),P(A^c))$.

From axiom~(U2) (density linearity): for every $\alpha>0$,
$\kappa(x,n,\alpha p)=\alpha\kappa(x,n,p)$.
This means $p\mapsto\kappa(x,n,p)$ is a positive homogeneous function of degree~1.
A degree-1 positive homogeneous function $\R_+\to\R_+$ must be of the form
$\kappa(x,n,p)=\kappa(x,n,1)\cdot p=:\tilde\kappa(x,n)\cdot p$,
where $\tilde\kappa(x,n):=\kappa(x,n,1)$ is the kernel evaluated at unit density.

From axiom~(U3) (spatial and directional isotropy): for all $R\in O(d)$ and $v\in\R^d$,
$\kappa(Rx+v,Rn,p)=\kappa(x,n,p)$,
which translates to $\tilde\kappa(Rx+v,Rn)=\tilde\kappa(x,n)$.

We claim this forces $\tilde\kappa$ to be a constant.  To see this:
\begin{itemize}[leftmargin=*,itemsep=2pt]
\item \textbf{Translational invariance ($v$ arbitrary, $R=I$):}
  $\tilde\kappa(x+v,n)=\tilde\kappa(x,n)$ for all $v\in\R^d$ and all $x$.
  Hence $\tilde\kappa(x,n)=\tilde\kappa(0,n)$ for all $x$; the $x$-dependence drops.

\item \textbf{Rotational invariance of the remaining $n$-dependence:}
  $\tilde\kappa(0,Rn)=\tilde\kappa(0,n)$ for all $R\in O(d)$.
  Since $O(d)$ acts transitively on $\mathcal{S}^{d-1}$ (for $d\ge 2$):
  $\tilde\kappa(0,n)$ is constant on $\mathcal{S}^{d-1}$.
  Write $\tilde\kappa(0,n)=c$ for all $n\in\mathcal{S}^{d-1}$.
\end{itemize}
Hence $\kappa(x,n,p)=cp$ for some constant $c>0$.

Substituting back into~\eqref{eq:bcf_form}:
\begin{equation}
\label{eq:phi_c}
  \mathcal C(f,P)
  =\frac{c\int_{\partial A}p\,d\Hd}{\min(P(A),P(A^c))}
  =\frac{c\cdot\mup_P(f)/2}{\min(P(A),P(A^c))}
  =\frac{c}{2}\cdot\psieff(f,P).
\end{equation}

\textbf{Step~2: Axiom~(U4) fixes $c=2$.}

Axiom~(U4): $\mathcal C(f_0,\mathcal{N}(0,I_d))=2\sqrt{2/\pi}$ where $f_0=\1\{x_1\ge 0\}$.

From~\eqref{eq:phi_c}: $\mathcal C(f_0,\mathcal{N})=(c/2)\cdot\psieff(f_0,\mathcal{N})$.
By Proposition~\ref{prop:margin}: $\psieff(f_0,\mathcal{N})=2\gauss(0)/\Phi(0)=2\cdot(2\pi)^{-1/2}/(1/2)=2\sqrt{2/\pi}$.
Hence $(c/2)\cdot 2\sqrt{2/\pi}=2\sqrt{2/\pi}$, giving $c=2$.

\textbf{Conclusion.}
$\mathcal C(f,P)=\psieff(f,P)$ for all admissible $(f,P)$.
\end{proof}

\begin{remark}[Role of the axioms]
Locality specifies the integral representation, density homogeneity gives the factor $p$, Euclidean invariance fixes the spatial and directional dependence, and the Gaussian reference fixes the multiplier $2$. The resulting kernel satisfies the density-weighted zero condition in Definition~\ref{def:bcf}, including densities that vanish on parts of the boundary.
\end{remark}

\subsection{Proofs of Theorems~\ref{thm:lipschitz_push} and~\ref{thm:submersion_push}}
\label{app:lipschitz_push}
\label{app:submersion_push}
%=======================================================================

\subsubsection{Proof of Theorem~\ref{thm:lipschitz_push}}
\begin{proof}
Refine the countable boundary cover into disjoint measurable patches $S_k$. On a neighborhood of each patch, the inverse function theorem and global injectivity give the density identity $q(Tx)=p(x)/|\det DT(x)|$. The image boundary is $T(\partial A)$ because $T$ is a homeomorphism. Apply the surface area formula to each $S_k$ and sum the integrable contributions. The exceptional set and its image have zero surface measure. This proves Eq.~\eqref{eq:lipschitz_push}.
\end{proof}

\subsubsection{Proof of Theorem~\ref{thm:submersion_push} (Dimensionality-reducing pushforward)}

\begin{proof}[Complete proof]
For $m<d$, the coarea formula for Lipschitz maps \citep[Theorem~3.2.22]{federer1969geometric} states that, for nonnegative $u$,
\begin{equation}
\label{eq:coarea_submersion}
  \int_{\R^d}u(x)\,J_d^m T(x)\,d\lambda^d(x)
  =\int_{\R^m}\!\left[\int_{T^{-1}(y)}u(x)\,d\mathcal{H}^{d-m}(x)\right]d\lambda^m(y),
\end{equation}
where $J_d^m T(x):=\sqrt{\det(DT\,DT^\top)}$ is the $m$-dimensional Jacobian.
Applying this identity to $Q=T_\#P$ yields the density formula
\[
q(y)=\int_{T^{-1}(y)}\frac{p(x)}{J_d^mT(x)}\,d\mathcal H^{d-m}(x).
\]
This fiber marginal integrates contributions from all preimages. A generic submersion can map $\partial A$ to an $m$-dimensional subset of $\R^m$ when $m<d$. The boundary coefficient therefore uses the actual codimension-one boundary $\partial T(A)$ with the marginal density $q$, whereas the diffeomorphic case preserves the original boundary correspondence.
\end{proof}

\begin{example}[Linear projection (PCA)]
\label{ex:pca}
Let $T=V^\top:\R^d\to\R^m$ where $V\in\R^{d\times m}$ has orthonormal columns
(the top-$m$ PCA directions). The coarea formula gives the projected density
\[
q(y)=\int_{\{x:V^\top x=y\}}p(x)\,d\mathcal H^{d-m}(x),
\]
because $J_d^mT=1$. This density is a marginal over the discarded coordinates. A projected-boundary formula additionally requires $V^\top(\partial A)$ to coincide with the actual decision boundary at the correct codimension. Table~\ref{tab:normalization_robustness} evaluates PCA-whitened protocols directly through the finite-feature estimator.
\end{example}

%=======================================================================
\subsection{Proof of Theorem~\ref{thm:upper_bound}}\label{app:ub}
\begin{proof}
For a query $x$, let $N_r(x)$ count training samples in $B(x,r)$ and predict their majority label, with a fixed rule for ties and empty balls. If $\operatorname{dist}(x,\partial A)>r$, the ball lies on one side of the boundary. Conditional on $N_r(x)=m$, the labels are independent with correctness probability $1-\eta$. Hoeffding's inequality gives error at most $e^{-h_\eta m}$, where $h_\eta=(1-2\eta)^2/2$; the same upper bound equals $1$ for an empty ball.

Writing $p_r(x)=P(B(x,r))$, the binomial generating function gives
\[
\E[e^{-h_\eta N_r(x)}]=(1-p_r(x)+p_r(x)e^{-h_\eta})^n
\le e^{-c_\eta n p_r(x)},\quad c_\eta=1-e^{-h_\eta}.
\]
The boundary tube contributes at most $\alpha\psi r+Lr^2$. Integrate the remaining pointwise bound against an independent query $X$ to obtain Eq.~\eqref{eq:local_vote_coverage}. Splitting the integral over $G$ and $G^c$ gives $e^{-c_\eta a n r^d}+\xi$. For Gaussian inputs, the density on $B(x,r)$ with $\|x\|\le R$ and $r\le r_0$ is at least $(2\pi)^{-d/2}e^{-(R+r_0)^2/2}$, giving the stated $a_R$ and explicit tail term.
\end{proof}

\subsection{Proof of Theorem~\ref{thm:tight_bilateral}}\label{app:tight_bilateral}
\begin{proof}
For every member of the family, $\alpha_f\psi_f\le\Psi/2$. The tube remainder, coverage constant, and admissible radius are uniform, so taking a supremum in Theorem~\ref{thm:upper_bound} preserves the same right-hand side. Substituting $r_n=(C\log n/n)^{1/d}$ bounds the local-vote term by $n^{-2}$ when $c_\eta a C\ge2$. The two geometric terms remain linear in $\Psi$ and $L$, respectively.
\end{proof}

\subsection{Proof of Theorem~\ref{thm:halfspace_tight}}\label{app:halfspace_tight}
\begin{proof}
Write $d_P(h)=P(h\ne f^*)$, $R(h)=\Prob(h(X)\ne Y)$, and
$Z_h=\1\{h(X)\ne Y\}-\1\{f^*(X)\ne Y\}$. Under symmetric noise,
\[
R(f^*)=\eta,\quad \E Z_h=(1-2\eta)d_P(h),\quad
\E Z_h^2=d_P(h).
\]
Let $\mathbb P$ be the joint observation law of $(X,Y)$ and $\mathbb P_n$ its empirical measure. For a VC class of dimension $v$, the relative VC/Bernstein inequality for these bounded loss differences gives, simultaneously in $h$, with probability at least $1-\delta$,
\[
|(\mathbb P_n-\mathbb P)Z_h|\le C\left(\sqrt{d_P(h)A_n/n}+A_n/n\right),
\quad A_n=v\log(en/v)+\log(2/\delta).
\]
This follows from VC growth-function bounds and variance-sensitive concentration, with localization by $d_P(h)$~\citep{vapnik1998statistical,shalev2014understanding}. For empirical-risk minimization, $\mathbb P_nZ_{\hat f}\le0$, hence
$(1-2\eta)d_P(\hat f)\le C\sqrt{d_P(\hat f)A_n/n}+CA_n/n$.
Applying $uv\le u^2/2+v^2/2$ to the square-root term and solving for $d_P(\hat f)$ gives $d_P(\hat f)\le C'A_n/((1-2\eta)^2n)$. Affine halfspaces have $v=d+1$.
\end{proof}

\subsection{Proofs for Section~\ref{sec:implicit_bias} and Population Logistic Minimizer (Gradient Flow)}
\label{app:gf_dir}
\label{app:gf_sec}
\label{app:gf_sym}
\label{sec:gf}
\label{sec:gf_abbrev}
%=======================================================================

We provide complete proofs for the components of Theorem~\ref{thm:gf}.
We restate the key lemmas with their labels for cross-reference.

\begin{lemma}[Minimizer direction via Jensen]
\label{lem:dir}
Let $P=\mathcal{N}(0,I_d)$, $w_*\in\mathcal S^{d-1}$, $b_*\in\R$, and $f^*=\1\{w_*^\top x+b_*\ge0\}$, and $\eta\in(0,1/2)$.
The population logistic loss $\mathcal{L}(w,b)=\E[\log(1+e^{-\tilde Y(w^\top X+b)})]$
has a unique finite minimizer $(\hat w,\hat b)$.
Moreover, $\hat w\parallel w_*$.
\end{lemma}

\begin{lemma}[Balanced target symmetry]
\label{lem:sym}
For $b_*=0$ and $\eta\in[0,1/2)$:
\begin{enumerate}[label=(\roman*),leftmargin=*,itemsep=1pt]
\item $\mathcal{L}(w,b)=\mathcal{L}(w,-b)$ for all $(w,b)$ \textbf{(symmetry)}.
\item The unique $b$-minimizer given $w$ is $b=0$ for $\eta\in[0,1/2)$.
\item For $\eta\in(0,1/2)$: the global minimizer has $\hat b=0$.
\end{enumerate}
\end{lemma}

\begin{corollary}[$\psieff$ minimizer for balanced halfspaces]
\label{cor:psieff_min}
Under $P=\mathcal{N}(0,I_d)$ and $\eta\in(0,1/2)$ with $b_*=0$:
the population logistic minimizer $\hat h=h_{\hat w,0}$ satisfies
$\psieff(\hat h,P)=\mathfrak g(0)=2\sqrt{2/\pi}$, the global isoperimetric minimum over
all halfspace classifiers.
\end{corollary}

\begin{proof}[Complete proof of Lemma~\ref{lem:dir}]

\textbf{Part A: Strict convexity.}
The population logistic loss is:
\[
  \mathcal{L}(w,b)=\E\!\bigl[\log(1+e^{-\tilde Y(w^\top X+b)})\bigr],
  \quad\tilde Y:=2Y-1\in\{-1,+1\}.
\]
Let $\theta:=(w,b)\in\R^{d+1}$ and $\ell_\theta(X):=w^\top X+b$.
The Hessian is:
\begin{equation}
\label{eq:hessian}
  \nabla^2_\theta\mathcal{L}(\theta)
  =\E\!\left[\sigma\!\left(\ell_\theta(X)\right)\!\left(1-\sigma(\ell_\theta(X))\right)
  \begin{pmatrix}X\\1\end{pmatrix}\begin{pmatrix}X\\1\end{pmatrix}^{\!\top}\right],
\end{equation}
where $\sigma(u):=1/(1+e^{-u})$ and we used
$\partial_\theta^2\log(1+e^{-\tilde Y\ell_\theta})=\sigma(\tilde Y\ell_\theta)(1-\sigma(\tilde Y\ell_\theta))(X,1)(X,1)^\top$
(with the augmented feature vector $(X,1)$).

For any nonzero $u=(v,c)\in\R^{d+1}$:
\begin{equation}
\label{eq:hessian_pos}
  u^\top\nabla^2\mathcal{L}(\theta)\,u
  =\E\!\left[\sigma(\ell_\theta)(1-\sigma(\ell_\theta))(v^\top X+c)^2\right].
\end{equation}
Since $\sigma(t)(1-\sigma(t))=e^{-t}/(1+e^{-t})^2>0$ for all $t\in\R$, and
since under $X\sim\mathcal{N}(0,I_d)$ the random variable $v^\top X+c$ has
variance $\|v\|^2>0$ (for $(v,c)\ne 0$; if $v=0$ then $c\ne 0$ and $v^\top X+c=c\ne 0$ a.s.):
the expectation in~\eqref{eq:hessian_pos} is strictly positive.
Hence $\mathcal{L}$ is strictly convex on $\R^{d+1}$.

\textbf{Part B: Coercivity and existence.}
Write $m(x)=\Prob(Y=1\mid X=x)\in\{\eta,1-\eta\}$ and $u=w^\top x+b$. Conditional logistic risk satisfies
\[
m(x)\log(1+e^{-u})+(1-m(x))\log(1+e^u)\ge\eta|u|.
\]
For $u\ge0$, use $1-m(x)\ge\eta$ and $\log(1+e^u)\ge u$; for $u\le0$, use the first term. Hence
\begin{equation}\label{eq:logistic_coercivity}
\mathcal L(w,b)\ge\eta\E|w^\top X+b|
\ge\eta a_d\|(w,b)\|,
\quad a_d=\min_{\|(v,c)\|=1}\E|v^\top X+c|>0.
\end{equation}
The expectation is continuous on the unit sphere and positive at every point: Gaussian affine functionals with nonzero coefficient vector have positive absolute first moment. Compactness gives $a_d>0$. Thus the sublevel sets of $\mathcal L$ are compact. Continuity gives existence of a finite minimizer and strict convexity gives uniqueness.

\textbf{Part C: Direction $\hat w\parallel w_*$.}
Write $w=\alpha w_*+w_\perp$ with $w_\perp\perp w_*$ (orthogonal decomposition).
Under $P=\mathcal{N}(0,I_d)$: let $V:=w_*^\top X\sim\mathcal{N}(0,1)$ and
$Z:=w_\perp^\top X\sim\mathcal{N}(0,\|w_\perp\|^2)$.
Since $w_*\perp w_\perp$: $\Cov(V,Z)=w_*^\top w_\perp=0$ and, since
jointly Gaussian, $V\perp Z$.

The regression function $m(x):=\Prob(Y=1\mid X=x)$ satisfies:
$m(x)=(1-\eta)\1\{w_*^\top x+b_*\ge0\}+\eta\1\{w_*^\top x+b_*<0\}=m(w_*^\top x)=m(V)$.
(It depends on $x$ only through $V=w_*^\top x$.)

Expand the loss:
\begin{align}
  \mathcal{L}(\alpha w_*+w_\perp,b)
  &=\E\bigl[m(V)\ell_{\rm log}(\alpha V+Z+b)+(1-m(V))\ell_{\rm log}(-\alpha V-Z-b)\bigr],
  \label{eq:loss_expanded}
\end{align}
where $\ell_{\rm log}(s):=\log(1+e^{-s})$.

For fixed $v$ (i.e., fixed $V=v$), the inner conditional expectation over $Z|V=v$
(note $Z\perp V$, so the conditional distribution is just the marginal of $Z$):
\[
  F(v):=\E_Z\bigl[m(v)\ell_{\rm log}(\alpha v+Z+b)+(1-m(v))\ell_{\rm log}(-\alpha v-Z-b)\bigr].
\]
The function $s\mapsto \mathcal J_v(s):=m(v)\ell_{\rm log}(\alpha v+s+b)+(1-m(v))\ell_{\rm log}(-\alpha v-s-b)$
is strictly convex in $s$:
\begin{align}
  \mathcal J_v''(s)
  &=m(v)\ell_{\rm log}''(\alpha v+s+b)+(1-m(v))\ell_{\rm log}''(-\alpha v-s-b)
  \notag\\
  &=m(v)\sigma(|\alpha v+s+b|)(1-\sigma(|\alpha v+s+b|))
  \notag\\
  &\quad+(1-m(v))\sigma(|\alpha v+s+b|)(1-\sigma(|\alpha v+s+b|))
  =\sigma(\ldots)(1-\sigma(\ldots))>0.
  \label{eq:G_convex}
\end{align}
(We used $\ell_{\rm log}''(t)=e^{-t}/(1+e^{-t})^2=\sigma(t)(1-\sigma(t))>0$.)

By Jensen's inequality (strictly convex $\mathcal J_v$, $Z$ has mean $\E[Z|V]=0$):
\begin{equation}
\label{eq:jensen_dir}
  F(v)=\E_Z[\mathcal J_v(Z)]\ge \mathcal J_v(\E[Z|V])=\mathcal J_v(0)=m(v)\ell_{\rm log}(\alpha v+b)+(1-m(v))\ell_{\rm log}(-\alpha v-b),
\end{equation}
with strict inequality whenever $\Var(Z|V)=\|w_\perp\|^2>0$, i.e., whenever $w_\perp\ne 0$.

Taking $\E_V$ over both sides:
\[
  \mathcal{L}(\alpha w_*+w_\perp,b)\ge\mathcal{L}(\alpha w_*,b),
\]
with strict inequality if $w_\perp\ne 0$.
Hence the unique minimizer satisfies $\hat w_\perp=0$. Moreover, for every finite $b$,
\[
\left.\partial_a\mathcal L(a w_*,b)\right|_{a=0}
=-\E[m(V)V]=-(1-2\eta)\gauss(b_*)<0.
\]
Strict convexity in $a$ therefore gives $\hat a>0$ and $\hat w=\hat a w_*$, in particular $\|\hat w\|>0$.
\end{proof}

\begin{proof}[Complete proof of Lemma~\ref{lem:sym}]

\textbf{Part~(i): Symmetry $\mathcal{L}(w,b)=\mathcal{L}(w,-b)$ for $b_*=0$, $\eta\in[0,1/2)$.}

\textbf{Joint-distribution symmetry.}
For $b_*=0$, the target satisfies $f^*(-x)=1-f^*(x)$ for $P$-almost every $x$. Thus the regression function $m(x)=\Prob(Y=1\mid X=x)$ obeys $m(-x)=1-m(x)$. Since $X$ is centered Gaussian and $\tilde Y=2Y-1$, this gives $(X,\tilde Y)\overset{d}{=}(-X,-\tilde Y)$.

\textbf{Substitution.}
Using this joint symmetry,
\begin{align}
\mathcal L(w,b)
&=\E\log(1+e^{-\tilde Y(w^\top X+b)})\notag\\
&=\E\log(1+e^{-(-\tilde Y)(w^\top(-X)+b)})\notag\\
&=\E\log(1+e^{-\tilde Y(w^\top X-b)})=\mathcal L(w,-b).
\label{eq:sym_derivation}
\end{align}

\textbf{Part~(ii): Unique $b$-minimizer at $0$.}
From Part~(i): $b\mapsto\mathcal{L}(w,b)$ is an even function.
From Lemma~\ref{lem:dir} (strict convexity, valid for all $\eta\in[0,1/2)$ since
$\sigma(1-\sigma)>0$ regardless of $\eta$): $b\mapsto\mathcal{L}(w,b)$ is strictly convex.
A strictly convex even function on $\R$ attains its unique global minimum at $0$
(since: (a) by parity $b^*=-b^*$, so $b^*=0$; (b) uniqueness from strict convexity).

\textbf{Part~(iii): Global minimizer has $\hat b=0$.}
For $\eta\in(0,1/2)$: Lemma~\ref{lem:dir} guarantees existence and uniqueness of
$(\hat w,\hat b)$.  The global minimizer must satisfy the marginal optimality
condition: $\hat b=\arg\min_{b}\mathcal{L}(\hat w,b)$.
By Part~(ii), this unique minimizer is $b=0$.  Hence $\hat b=0$.
\end{proof}

%=======================================================================
\subsection{Proof of Proposition~\ref{prop:gradient} (Gradient Formula)}
\label{app:grad_proof}
%=======================================================================

\begin{proof}[Complete proof]

\textbf{Step 1: Simplification.}
Since $P=\mathcal{N}(0,I_d)$ and $P(A)=1/2$: the pushforward
$(T_W)_\#P=\mathcal{N}(0,WW^\top)$ is symmetric about $0$, so
$(T_W)_\#P(T_W(A))=P(A)=1/2$ for any $W\in GL(d)$.
Put $Q_W=(T_W)_\#P$ and $g_W=\1_{T_W(A)}$. Then $\min(Q_W(g_W=1),Q_W(g_W=0))=1/2$, so
\[
  \psi_W=\mup_{Q_W}(g_W)/(1/2)
  =4\int_{\partial A}p(x)|W^{-\top}n(x)|\,d\Hd(x).
\]

\textbf{Step 2: Differentiate $W^{-\top}$.}
From the identity $(W^{-\top})W^\top=I_d$, differentiate both sides with respect to $W_{ij}$:
\begin{equation}
\label{eq:wtop_diff}
  \frac{\partial W^{-\top}}{\partial W_{ij}}\cdot W^\top
  +W^{-\top}\cdot\frac{\partial W^\top}{\partial W_{ij}}=0.
\end{equation}
The matrix $\partial W^\top/\partial W_{ij}$ has $(k,l)$-entry
$\partial(W^\top)_{kl}/\partial W_{ij}=\partial W_{lk}/\partial W_{ij}=\delta_{li}\delta_{kj}$, where $\delta_{ab}=\1\{a=b\}$ is the Kronecker delta,
i.e., $\partial W^\top/\partial W_{ij}=E_{ji}$ (the matrix with $1$ at position $(j,i)$).
Solving~\eqref{eq:wtop_diff}:
\begin{equation}
\label{eq:inv_trans_diff}
  \frac{\partial W^{-\top}}{\partial W_{ij}}=-W^{-\top}E_{ji}W^{-\top}.
\end{equation}
Entry-wise:
$\left[\partial W^{-\top}/\partial W_{ij}\right]_{kl}
=-\sum_{m,p}(W^{-\top})_{km}(E_{ji})_{mp}(W^{-\top})_{pl}
=-(W^{-\top})_{kj}(W^{-\top})_{il}$.

\textbf{Step 3: Differentiate $a_k=(W^{-\top}n)_k$.}
For fixed $n\in\R^d$, let $a(W):=W^{-\top}n$, so $a_k=\sum_l(W^{-\top})_{kl}n_l$.
\begin{align}
  \frac{\partial a_k}{\partial W_{ij}}
  &=\sum_l\left[\frac{\partial W^{-\top}}{\partial W_{ij}}\right]_{kl}n_l
  \notag\\
  &=\sum_l\bigl[-(W^{-\top})_{kj}(W^{-\top})_{il}\bigr]n_l
  =-(W^{-\top})_{kj}(W^{-\top}n)_i
  \notag\\
  &=-(W^{-\top})_{kj}a_i.
  \label{eq:da_dW}
\end{align}

\textbf{Step 4: Differentiate $|a|=(\sum_k a_k^2)^{1/2}$.}
\begin{align}
  \frac{\partial|a|}{\partial W_{ij}}
  =\frac{1}{|a|}\sum_k a_k\frac{\partial a_k}{\partial W_{ij}}
  =\frac{1}{|a|}\sum_k a_k\bigl[-(W^{-\top})_{kj}a_i\bigr]
  =-\frac{a_i}{|a|}\sum_k a_k(W^{-\top})_{kj}.
  \label{eq:dnorm_comp}
\end{align}
Note: $\sum_k a_k(W^{-\top})_{kj}=(a^\top W^{-\top})_j=(W^{-1}a)_j$
since $(W^{-\top})_{kj}=(W^{-1})_{jk}$.
Also: $W^{-1}a=W^{-1}W^{-\top}n=(W^\top W)^{-1}n$.
Hence:
\begin{equation}
\label{eq:dnorm_final}
  \frac{\partial|W^{-\top}n|}{\partial W_{ij}}
  =-\frac{(W^{-\top}n)_i\,[(W^\top W)^{-1}n]_j}{|W^{-\top}n|}.
\end{equation}
In matrix form: the $(i,j)$-entry of $\nabla_W|W^{-\top}n|$ is~\eqref{eq:dnorm_final},
giving the rank-1 outer product:
\begin{equation}
\label{eq:matrix_grad}
  \nabla_W|W^{-\top}n|=-\frac{(W^{-\top}n)[(W^\top W)^{-1}n]^\top}{|W^{-\top}n|}.
\end{equation}

\textbf{Step 5: Interchange differentiation and integration.}
The integrand $x\mapsto p(x)|W^{-\top}n(x)|$ is $C^\infty$ in $W\in GL(d)$
for each fixed $x$ (since $W\mapsto W^{-\top}$ is $C^\infty$ on invertible matrices
and $n(x)$ is fixed).  The derivative satisfies
$|\partial_W p(x)|W^{-\top}n(x)||\le p(x)\cdot C_W\cdot|n(x)|$
for a locally bounded $C_W$ (uniform on compact subsets of $GL(d)$).
The assumed finite integral $\int_{\partial A}p\,d\Hd$ supplies an integrable dominating function:
dominated convergence justifies differentiating under the integral sign.

\textbf{Conclusion.}
\begin{align*}
  \nabla_W\psi_W
  &=4\nabla_W\int_{\partial A}p(x)|W^{-\top}n(x)|\,d\Hd(x)
  \\
  &=4\int_{\partial A}p(x)\nabla_W|W^{-\top}n(x)|\,d\Hd(x)
  \\
  &=-4\int_{\partial A}p\,
  \frac{(W^{-\top}n)[(W^\top W)^{-1}n]^\top}{|W^{-\top}n|}\,d\Hd.
  \qedhere
\end{align*}
\end{proof}

%=======================================================================
\subsection{Proofs for Oscillatory Feature Geometry}\label{app:family}
\begin{lemma}[Periodic zero-set area]\label{lem:patch}
Let $\Sigma^w=\{u\in(-\pi,\pi)^d:\sum_iw_i\sin u_i=0\}$ and suppose $w$ has at least three nonzero coordinates. Then
\[
(2d)^{-(d-1)}\le\mathcal H^{d-1}(\Sigma^w)\le2d(2\pi)^{d-1}.
\]
\end{lemma}
\begin{proof}
Choose $j$ with $|w_j|=\max_i|w_i|$ and put $a=1/(4d)$. For $|u_i|<a$, $i\ne j$,
$|\sum_{i\ne j}w_i\sin u_i|\le(d-1)a|w_j|<|w_j|/4$.
Thus the principal arcsine defines a smooth zero-set graph over $(-a,a)^{d-1}$, giving area at least $(2a)^{d-1}$.
For the upper bound, each line parallel to an active coordinate has at most two regular zero crossings per period. The area formula for coordinate projections gives
$\int_{\Sigma^w}|n_j|\,d\Hd\le2(2\pi)^{d-1}$ for each active $j$.
On the regular part, $\sum_j|n_j|\ge1$, so summing gives the bound. At a singular point, all active cosines vanish. At least three active coordinates therefore make this singular set dimension at most $d-3$, with zero $\Hd$ measure.
\end{proof}

\begin{lemma}[Periodized Gaussian density]\label{lem:poisson}
For $\Theta_{\omega,d}(u)=\sum_{k\in\Z^d}\gauss_d((u+2\pi k)/\omega)$,
\begin{equation}\label{eq:theta_formula}
\Theta_{\omega,d}(u)=\left(\frac{\omega}{2\pi}\right)^d(1+R_{\omega,d}(u)),
\qquad |R_{\omega,d}(u)|\le\tfrac12
\end{equation}
when $\omega\ge\sqrt{2\log(16d)}$.
\end{lemma}
\begin{proof}
Poisson summation and the Gaussian Fourier transform give
\[
\Theta_{\omega,d}(u)=\left(\frac{\omega}{2\pi}\right)^d
\sum_{m\in\Z^d}e^{-\omega^2\|m\|^2/2}e^{im\cdot u}.
\]
Set $q=e^{-\omega^2/2}\le1/(16d)$. Since $\sum_{n\ge1}q^{n^2}\le q/(1-q)\le1/(15d)$, the absolute remainder is at most
$(1+2/(15d))^d-1\le e^{2/15}-1<1/2$.
\end{proof}

\begin{lemma}[Density bounds for three controlled coefficients]\label{lem:density_bound}
Let $w\in\mathcal W_{d,\kappa}$ and $\omega\ge\omega_0(d)$. Define the scalar score $S_{w,\omega}=\sum_iw_i\sin(\omega X_i)$ and its density $p_{S_{w,\omega}}$. This density is continuous with
$0<c_d\le p_{S_{w,\omega}}(0)\le C\kappa^{-1}$.
\end{lemma}
\begin{proof}
Proposition~\ref{prop:rdensity} bounds $r_\omega$ above and below by fixed positive multiples of the arcsine density $r_{\rm arc}(z)=(\pi\sqrt{1-z^2})^{-1}$ on $(-1,1)$. This arcsine density is in $L^{3/2}$. The density of $w_iZ_i$ has $L^{3/2}$ norm $|w_i|^{-1/3}\|r_\omega\|_{3/2}$. For $c\ne0$, write $r_c(t)=|c|^{-1}r_\omega(t/c)$ for the density of $cZ_i$. Relabel three coordinates with $|w_i|\ge\kappa$ as $1,2,3$. Young's convolution inequality gives
\[
\|r_{w_1}*r_{w_2}*r_{w_3}\|_\infty
\le\prod_{j=1}^3\|r_{w_j}\|_{3/2}\le C\kappa^{-1}.
\]
Approximation in $L^{3/2}$ by smooth compactly supported functions gives continuity of this convolution. Convolving with the remaining probability measures preserves its bound and continuity.
For a lower bound, take $j$ with $|w_j|=\max_i|w_i|$ and integrate the density formula
\[
p_{S_{w,\omega}}(0)=\frac1{|w_j|}\int
r_\omega\!\left(-\frac{\sum_{i\ne j}w_i z_i}{w_j}\right)
\prod_{i\ne j}r_\omega(z_i)\,dz_{-j}
\]
over $|z_i|<1/(4d)$ for $i\ne j$, where $dz_{-j}=\prod_{i\ne j}dz_i$. Its first argument lies in $(-1/4,1/4)$ and every density factor has a uniform positive lower bound there. The box has positive $(d-1)$-volume depending only on $d$, proving $p_{S_{w,\omega}}(0)\ge c_d$.
\end{proof}

\begin{proposition}[Density of the arcsine-approximation]
\label{prop:rdensity}
For $X\sim\mathcal N(0,1)$, $Z_\omega:=\sin(\omega X)$ has density:
\[
  r_\omega(z)=\frac{1}{\pi\sqrt{1-z^2}}\!\left[1+2\sum_{\ell=1}^\infty e^{-2\ell^2\omega^2}
  \cos(2\ell\arcsin z)\right]\cdot\1_{(-1,1)}(z).
\]
Consequently $(1-\varepsilon_\omega)r_{\rm arc}\le r_\omega\le(1+\varepsilon_\omega)r_{\rm arc}$,
$\varepsilon_\omega:=2\sum_{\ell=1}^\infty e^{-2\ell^2\omega^2}\to 0$ exponentially.
\end{proposition}

\begin{proof}[Complete proof]
The preimages of $z\in(-1,1)$ under $x\mapsto\sin(\omega x)$ form two families:
$x_{k,+}:=(\arcsin z+2\pi k)/\omega$ and $x_{k,-}:=(\pi-\arcsin z+2\pi k)/\omega$, $k\in\Z$.
The density is obtained by summing Gaussian density over preimages:
\begin{align}
  r_\omega(z)
  &=\frac{1}{\omega\sqrt{1-z^2}}\sum_{k\in\Z}\!\left[\gauss\!\left(\frac{\arcsin z+2\pi k}{\omega}\right)
  +\gauss\!\left(\frac{\pi-\arcsin z+2\pi k}{\omega}\right)\right].
  \label{eq:density_sum}
\end{align}
Apply the one-dimensional Poisson summation formula (Lemma~\ref{lem:poisson} with $d=1$):
\begin{equation}
  \sum_{k\in\Z}\gauss\!\left(\frac{\alpha+2\pi k}{\omega}\right)
  =\frac{\omega}{2\pi}\sum_{m\in\Z}e^{-m^2\omega^2/2}e^{im\alpha}.
  \label{eq:1d_poisson}
\end{equation}
Set $\alpha=\arcsin z$ for the first sum and $\alpha=\pi-\arcsin z$ for the second.
Substitute into~\eqref{eq:density_sum}:
\begin{align}
  r_\omega(z)
  &=\frac{1}{2\pi\sqrt{1-z^2}}\sum_{m\in\Z}e^{-m^2\omega^2/2}
  \!\left[e^{im\arcsin z}+e^{im(\pi-\arcsin z)}\right].
  \label{eq:rdensity_fourier}
\end{align}
Using $e^{im(\pi-\alpha)}=e^{im\pi}e^{-im\alpha}=(-1)^m e^{-im\alpha}$:
\begin{itemize}[itemsep=0pt]
\item For even $m=2\ell$: $e^{2i\ell\alpha}+e^{-2i\ell\alpha}=2\cos(2\ell\alpha)$.
\item For odd $m$: $e^{im\alpha}+(-1)^m e^{-im\alpha}=e^{im\alpha}-e^{-im\alpha}=2i\sin(m\alpha)$.
  The contribution of $+m$ and $-m$ for odd $m$: $e^{-m^2\omega^2/2}2i\sin(m\alpha)
  +e^{-m^2\omega^2/2}2i\sin(-m\alpha)=0$.  So odd terms cancel.
\end{itemize}
Only even terms survive:
\[
  r_\omega(z)=\frac{1}{\pi\sqrt{1-z^2}}\!\left[1+2\sum_{\ell=1}^\infty e^{-2\ell^2\omega^2}
  \cos(2\ell\arcsin z)\right].
\]
Since $|\cos(\cdot)|\le 1$: $r_\omega\le r_{\rm arc}(1+2\sum_{\ell\ge 1}e^{-2\ell^2\omega^2})
=r_{\rm arc}(1+\varepsilon_\omega)$ and similarly $r_\omega\ge r_{\rm arc}(1-\varepsilon_\omega)$.
\end{proof}

\begin{proof}[Proof of Theorem~\ref{thm:family_reduction}]
Under $x=(u+2\pi k)/\omega$, surface measure scales by $\omega^{-(d-1)}$. Summing period cells gives
\begin{equation}\label{eq:mup_poisson_sum}
\mup_P(f_{w,\omega})=2\omega^{-(d-1)}
\int_{\Sigma^w}\Theta_{\omega,d}(u)\,d\Hd(u).
\end{equation}
Lemmas~\ref{lem:patch} and~\ref{lem:poisson} give positive multiples of $\omega$ as upper and lower bounds. Gaussian symmetry and the continuous score distribution imply class probability $1/2$, hence $\psieff=2\mup_P$.
In represented coordinates, $\|w\|=1$ and score density continuity give
$\mup_{Q_\omega}(h_w)=2p_{S_{w,\omega}}(0)$ and $\psieff(h_w,Q_\omega)=4p_{S_{w,\omega}}(0)$. Lemma~\ref{lem:density_bound} proves part~(ii). The map is common to all $w$. For $\bar w=(1,1,1,0,\ldots,0)/\sqrt3$, all quantities reduce exactly to their three-dimensional counterparts after integrating the other Gaussian coordinates. The constants in the ratio are therefore independent of ambient $d$, proving part~(iv).
\end{proof}

\subsection{Explicit Packing and Information Bounds}\label{app:stat_lb}
\begin{lemma}[Fano bound with separate separation and information scales]\label{lem:fano}
Let $f_1,\ldots,f_M$, $M\ge2$, satisfy
$\Delta\le P(f_j\ne f_k)\le D$ for distinct $j,k$. Under independent symmetric label noise $\eta\in(0,1/2)$ and an index $J$ uniform on $\{1,\ldots,M\}$, every decoder $\hat J$ based on $n$ observations satisfies
\[
\Prob(\hat J\ne J)\ge1-\frac{nD_\eta D+\log2}{\log M},
\quad D_\eta=(1-2\eta)\log\frac{1-\eta}{\eta}.
\]
For classifiers, nearest-hypothesis decoding gives the same lower bound on the probability that $P(\hat f\ne f_J)\ge\Delta/2$.
\end{lemma}
\begin{proof}
The one-sample KL divergence is exactly $D_\eta P(f_j\ne f_k)$. Its $n$-sample value is at most $nD_\eta D$, which also bounds the mutual information of a uniform hypothesis index. Fano's inequality~\citep{tsybakov2009introduction} proves the statement. The disagreement triangle inequality supplies the decoding step. The related Bernoulli bound is
$D_\eta\le(1-2\eta)^2/(\eta(1-\eta))$ by the KL--$\chi^2$ inequality.
\end{proof}

\begin{lemma}[Phase-flip packing]\label{lem:packing}
There are absolute constants $c_0>0,d_0$ such that for $d\ge d_0$ and $\omega\ge1$, one can choose $\Theta\subset\{0,\pi\}^d$ with $\log|\Theta|\ge c_0d$ and pairwise disagreement at least $1/4$ for
\[
g_\theta(x)=\1\left\{\sum_{i=1}^d\sin(\omega x_i+\theta_i)\ge0\right\}.
\]
\end{lemma}
\begin{proof}
A greedy binary code construction removes Hamming balls of radius $\lfloor d/4\rfloor$ around both each chosen word and its complement. It yields at least
$2^d/(2\sum_{j\le d/4}\binom{d}{j})$ words, with pairwise distances between $d/4$ and $3d/4$ up to rounding. The binary entropy bound gives $\log M\ge c_0d$ for an absolute $c_0>0$ and large $d$.
Map a word $v$ to $\theta_i=\pi v_i$. For two words, split the sum into the flipped-coordinate sum $A$ and shared-coordinate sum $B$. Independence and symmetry give disagreement event $|A|>|B|$. Each summand is bounded, has mean zero and variance
$\sigma_\omega^2=(1-e^{-2\omega^2})/2$, uniformly positive for $\omega\ge1$.
Both sums contain at least a fixed fraction of $d$ coordinates. Berry--Esseen applied to the two independent sums bounds the difference from the Gaussian comparison by $C/\sqrt d$~\citep{petrov1975sums}. If $m$ coordinates are flipped, the Gaussian disagreement is
$\frac2\pi\arctan\sqrt{m/(d-m)}\ge1/3+O(1/d)$.
For sufficiently large $d$ the comparison error is at most $1/12$, proving the claimed separation.
\end{proof}

\begin{theorem}[Phase-family lower bound]\label{thm:minimax_family}
Let $\mathcal F_\omega^\pm=\{g_\theta:\theta\in\{0,\pi\}^d\}\subset\mathcal F_\omega$. For $d\ge d_0$ enlarged if necessary, $\omega\ge1$, and $n\le c_{\eta,\rm pack} d$ with $c_{\eta,\rm pack}=c_0/(4D_\eta)$,
\[
\inf_{\hat f}\sup_{f\in\mathcal F_\omega^\pm}
\Prob_f\{P(\hat f\ne f)\ge1/8\}\ge1/2.
\]
\end{theorem}
\begin{proof}
Use Lemma~\ref{lem:packing} with $\Delta=1/4$, the universal upper bound $D=1$, and $\log M\ge c_0d$. Then $nD_\eta\le c_0d/4$ and, for large $d$, $\log2\le c_0d/4$. Lemma~\ref{lem:fano} gives the result. The constant $c_{\eta,\rm pack}$ depends on the noise rate and the absolute packing constant.
\end{proof}

\begin{theorem}[An interval-modulated oscillatory family]\label{thm:minimax_local}
Let $P=\mathcal N(0,I_d)$, $m=\lfloor\omega/\pi\rfloor$, $a_j=j\pi/\omega$ for $1\le j\le m$, and $h=\pi/(8\omega)$. For a label-flip vector $v\in\{0,1\}^m$, put
\[
B_j=\{x:|x_1-a_j|<h\}\cup\{x:|x_1+a_j|<h\},\qquad
f_v(x)=\1\{\sin(\omega x_1)\ge0\}\mathbin\oplus
\1_{\cup_{j:v_j=1}B_j}(x).
\]
For all sufficiently large $\omega$, the family $\mathcal G_\omega=\{f_v:v\in\{0,1\}^m\}$ has balanced classes and $\psieff(f_v,P)=\Theta(\omega)$ uniformly in $v$. There are constants $c,c_\eta'>0$ such that
\[
n\le c_\eta'\omega\quad\Longrightarrow\quad
\inf_{\hat f}\sup_{f\in\mathcal G_\omega}
\Prob_f\{P(\hat f\ne f)\ge c\}\ge1/2.
\]
\end{theorem}
\begin{proof}
The sets $B_j$ are disjoint and invariant under $x\mapsto-x$. Therefore $f_v(-x)=1-f_v(x)$ almost surely, giving balance. Every original zero crossing $x_1=k\pi/\omega$ remains a boundary after these flips, and each active $B_j$ adds four endpoint hyperplanes. The original Gaussian-weighted boundary sum is $\Theta(\omega)$ by one-dimensional Poisson summation; the added surface content is at most $4m\gauss(0)$. Thus all family members have complexity $\Theta(\omega)$.

A binary code of length $m$ gives $\log M\ge c_0m$ and pairwise Hamming distance at least $m/4$. Since $a_j\le1$ and $h\le1$, each differing coordinate contributes mass at least $4h\gauss(2)=\pi\gauss(2)/(2\omega)$. For $\omega\ge2\pi$, $m\ge\omega/(2\pi)$, so the disagreement is at least $\Delta_0=\gauss(2)/16$. Apply Lemma~\ref{lem:fano} with this separation, $D=1$, and $n\le c_0m/(4D_\eta)$. For sufficiently large $m$, the error probability is at least $1/2$. Taking $c=\Delta_0/2$ and $c_\eta'=c_0/(8\pi D_\eta)$ proves the claim.
\end{proof}

\paragraph{Family structure.}
The phase family varies one sign per coordinate and has a fixed halfspace representation. The interval-modulated family varies labels independently on $m$ symmetric slab pairs, which supplies the frequency-dependent packing size. Each lower bound uses the family explicitly constructed in its statement.

\subsection{Proof of Theorem~\ref{thm:bv_bilateral} (BV Extension)}
\label{app:bv_bilateral}
%=======================================================================

\begin{proof}
The weighted-content assumption gives
$P(\operatorname{dist}(X,\partial^*A)\le t)=\mup_P(f)t+o(t)$ directly. A learner whose geometric errors lie in this tube therefore has bias at most this probability. Since $\mup_P(f)=\psieff(f,P)\min(P(A),P(A^c))$, the boundary complexity determines the stated first-order bias coefficient.
\end{proof}

\subsection{Proof of Theorem~\ref{thm:manifold_rate}}\label{app:manifold_rate}
\begin{proof}
Use intrinsic normal coordinates around the compact $C^2$ decision boundary in $\mathcal M$. The intrinsic volume Jacobian and the $C^1$ density give tube probability $\alpha\psi r+O(r^2)$, uniformly for sufficiently small geodesic radius. Outside this tube a geodesic ball lies within one decision region. Its sample count is binomial with success probability at least $a r^k$. The binomial-Hoeffding argument of Appendix~\ref{app:ub} therefore gives $e^{-c_\eta a n r^k}$, and addition of the tube term proves the result.
\end{proof}

\subsection{Proof of Theorem~\ref{thm:opt_rep} (Optimal Linear Representation)}
\label{app:opt_rep}
%=======================================================================

\begin{proof}[Complete proof]

\textbf{Part~(i): Stationarity condition.}

Recall $\psi_W=4\int_{\partial A}p(x)|W^{-\top}n(x)|\,d\mathcal{H}^{d-1}(x)$
for the balanced case.  From Proposition~\ref{prop:gradient}:
\[
  \nabla_W\psi_W=-4\int_{\partial A}p(x)\frac{(W^{-\top}n)[(W^\top W)^{-1}n]^\top}{|W^{-\top}n|}\,d\mathcal{H}^{d-1}(x).
\]
With the constraint $\det W=1$, stationarity requires
$\nabla_W\psi_W+4\lambda W^{-\top}=0$ for a scaled Lagrange multiplier $\lambda$. Rearranging gives:
\[
  W^{-\top}\!\left[\int_{\partial A}p(x)\frac{n(x)n(x)^\top}{\|W^{-\top}n(x)\|}\,d\mathcal{H}^{d-1}(x)\right]W^{-1}
  =\lambda I .
\]
The scalar $\|W^{-\top}n(x)\|^{-1}$ remains inside the integral because it depends on the boundary normal. These weights give the stationarity equation its boundary-dependent form.

\textbf{Part~(ii): Minimum on $O(d)$.}

For $W=R\in O(d)$ (orthogonal): $|R^{-\top}n|=|R^{-1}n|=|n|=1$ (since orthogonal
matrices preserve norms).  Hence:
\[
  \psi_R=4\int_{\partial A}p(x)\,d\mathcal{H}^{d-1}(x)=\psieff(f,P)
\]
is \textbf{constant} over all $R\in O(d)$.  Thus any $R\in O(d)$ achieves the
same $\psieff$.  All orthogonal directions give this common value.

\begin{remark}
This shows that orthogonal transformations preserve $\psieff$ exactly (as expected
from the isometry invariance of Hausdorff measure). Under the $\det W=1$
normalization, beneficial representations must be non-orthogonal anisotropic maps.
\end{remark}

\textbf{Part~(iii): Second-moment preconditioner.}

Let $\Sigma_f=V\Lambda V^\top$ with $\Lambda=\mathrm{diag}(\lambda_1,\ldots,\lambda_d)$
($\lambda_1\ge\cdots\ge\lambda_d>0$; assume $\Sigma_f$ is positive definite, otherwise
project onto its support). The volume-normalized map
$W_{\rm pre}=(\det\Sigma_f)^{-1/(2d)}V\Lambda^{1/2}V^\top$
satisfies $W_{\rm pre}^{-\top}\Sigma_fW_{\rm pre}^{-1}=(\det\Sigma_f)^{1/d}I$ and $\det W_{\rm pre}=1$. Because the true stationarity equation contains
$\|W^{-\top}n(x)\|^{-1}$ inside the integral, $W_{\rm pre}$ serves as a preconditioner or initialization for numerical optimization.

\textbf{Proof of Corollary~\ref{cor:opt_rep_reduction}.}
The trace of $H$ is zero, so $\det\exp(tH)=\exp(t\operatorname{tr}H)=1$. At $W=I$, Proposition~\ref{prop:gradient} gives $\nabla_W\psi=-4\Sigma_f$. Since $\dot W_0=H$ and $\langle\Sigma_f,H\rangle_F=\|H\|_F^2$, the claimed derivative follows. Anisotropy gives $H\ne0$, hence strict descent for sufficiently small positive $t$.
\end{proof}

%=======================================================================
\subsection{Proof of Proposition~\ref{prop:sep_ns} (Noise Sensitivity)}
\label{app:ns_expansion}
\begin{proof}
Write $s$ for signed distance to the compact $C^2$ boundary and $x$ for its closest boundary point. On this compact tubular neighborhood, the Gaussian transition has normal displacement $\sqrt\delta U_n+O(\delta)$, where $U_n=\langle U,n(x)\rangle\sim\mathcal N(0,1)$. For $s>0$, its leading crossing probability is $\Phi(-s/\sqrt\delta)$; the other side contributes the same leading term. The density and area element equal $p(x)+O(|s|)$ times $d\Hd(x)\,ds$. Gaussian tail bounds control the part away from this neighborhood. Integrating the normal displacement and the $O(|s|)$ corrections yields
\[
\mathrm{NS}_\delta(f)=2\int_{\partial A}p(x)\,d\Hd(x)
\int_0^\infty\Phi(-s/\sqrt\delta)\,ds+O(\delta).
\]
Since $\int_0^\infty\Phi(-u)\,du=(2\pi)^{-1/2}$ and $\psieff=4\int_{\partial A}p\,d\Hd$ in the balanced case, Eq.~\eqref{eq:ns_expansion} follows. As a normalization check, a centered Gaussian halfspace has
\[
\mathrm{NS}_\delta=\pi^{-1}\arccos\sqrt{1-\delta}
=\pi^{-1}\sqrt\delta+O(\delta^{3/2}),\qquad
\psieff=2\sqrt{2/\pi}.
\]
\end{proof}

\subsection{Proofs of Theorems~\ref{thm:nonasymptotic_ub} and~\ref{thm:nonasymptotic_lb}}
\label{app:nonasymptotic}\label{app:nonasymptotic_lb}
\begin{proof}[High-probability upper bound]
For a fitted classifier, its error on the boundary tube is at most $\alpha\psi r+Lr^2$ and its error on $G^c$ is at most $\xi$. The nonnegative remaining random error has expectation at most $e^{-c_\eta a n r^d}$ by Appendix~\ref{app:ub}. Markov's inequality bounds it by $\delta^{-1}e^{-c_\eta a n r^d}$ with probability at least $1-\delta$. This proves Eq.~\eqref{eq:nonasymptotic_ub}. The stated bandwidth gives Eq.~\eqref{eq:nonasymptotic_explicit}; the choice in Corollary~\ref{cor:sample_size} bounds each of the four terms by $\varepsilon/4$.
\end{proof}
\begin{proof}[Finite-family lower bound]
For each pair, the one-sample KL divergence is $D_\eta P(f_j\ne f_k)\le D_\eta D$. Fano's inequality with a uniform hypothesis index gives testing error at least $1-(nD_\eta D+\log2)/\log M$. Nearest-hypothesis decoding turns disagreement below $\Delta/2$ into a correct index. Multiplication by $\Delta/2$ proves the expected-risk bound. Lemma~\ref{lem:fano} states the same information calculation explicitly.
\end{proof}

\subsection{Proofs for Population and Kernel Dynamics}
\label{app:gf_psieff}\label{app:lazy_psieff}
\begin{proof}[Proof of Theorem~\ref{thm:gf_psieff}]
Let $\hat\theta=(\hat w,0)$ and $\mathcal L_0=\{\theta:\mathcal L(\theta)\le\mathcal L(\theta_0)\}$. Coercivity makes $\mathcal L_0$ compact, and convexity makes it convex. The continuous logistic Hessian is positive definite at every finite point, so its smallest eigenvalue on $\mathcal L_0$ has a positive minimum $\lambda_{\mathcal L_0}$. Along the flow, $\mathcal L$ decreases and the trajectory stays in $\mathcal L_0$. The Hessian also has a global finite upper bound because Gaussian augmented features have finite second moment. The flow is global, and
\[
\frac{d}{dt}\frac12\|\theta_t-\hat\theta\|^2
=-\langle\theta_t-\hat\theta,\nabla\mathcal L(\theta_t)-\nabla\mathcal L(\hat\theta)\rangle
\le-\lambda_{\mathcal L_0}\|\theta_t-\hat\theta\|^2.
\]
Thus $\theta_t\to\hat\theta=(\hat w,0)$ exponentially. Theorem~\ref{thm:gf} gives $\|\hat w\|>0$, so $r_t=\|w_t\|$ is bounded below for large $t$ and $\dot r_t/r_t\to0$.

Evenness in $b$ gives $\partial_b\mathcal L(w,0)=0$, hence
\[
\dot b_t=-a_t b_t,\qquad
a_t=\int_0^1\partial_{bb}\mathcal L(w_t,sb_t)\,ds.
\]
Near the limiting point, $a_t\ge a_0>0$. If $b_0=0$, uniqueness of the flow preserves $b_t=0$. Otherwise its sign is constant and $|b_t|>0$ at finite times. For $\zeta_t=|b_t|/r_t$, choose $T_0$ so that $|\dot r_t/r_t|\le a_0/2$. Then
\[
\dot\zeta_t=\zeta_t(-a_t-\dot r_t/r_t)\le-(a_0/2)\zeta_t,\qquad t\ge T_0.
\]
Write $\mathfrak g_+(u)=2\gauss(u)/\Phi(-u)$ for $u\ge0$. On a sufficiently small interval $[0,u_0]$, its right-continuous derivative satisfies $0<m\le \mathfrak g_+'(u)\le M$ because $\mathfrak g_+'(0)=4/\pi$. Therefore
\[
m\zeta\le \mathfrak g_+(\zeta)-\mathfrak g_+(0)\le M\zeta,
\quad
\dot\delta_t=\mathfrak g_+'(\zeta_t)\dot\zeta_t
\le-\frac{a_0m}{2M}\delta_t.
\]
This proves eventual monotonicity and the exponential bound. The balanced trajectory $b_t=0$ gives $\delta_t=0$ whenever $w_t\ne0$ and obeys the same conclusion.
\end{proof}

\begin{proof}[Proof of Proposition~\ref{prop:lazy_psieff}]
Put $E_t=F_t-F^*$. Self-adjointness and the positive spectrum give
$\frac12\partial_t\|E_t\|_{L^2(Q)}^2=-\langle E_t,\mathcal KE_t\rangle\le-\lambda_0\|E_t\|_{L^2(Q)}^2$.
Finite-dimensional norm equivalence gives $\|E_t\|_{C^1}\le C_V\|E_t\|_{L^2(Q)}$.

For sufficiently small $\|E_t\|_{C^1}$, the normal-coordinate implicit function theorem represents the zero set as $x+u_t(x)n_*(x)$ for $x\in\Sigma_*$, where $n_*=\nabla F^*/\|\nabla F^*\|$ is a unit normal and $u_t$ is the normal displacement. The lower gradient bound and the positive gap away from the tubular neighborhood give
$\|u_t\|_{C^1(\Sigma_*)}\le C\|E_t\|_{C^1}$ and account for the full zero set. The surface Jacobian $J_t$ of this graph satisfies $|J_t-1|\le C\|u_t\|_{C^1}$, and density regularity gives $|q(x+u_tn_*)-q(x)|\le\|\nabla q\|_\infty\|u_t\|_\infty$. Consequently,
\[
|\mu_Q^+(\1\{F_t\ge0\})-\mu_Q^+(\1\{F^*\ge0\})|
\le C\|u_t\|_{C^1}.
\]
The symmetric-difference region lies between these normal graphs, so the class-probability change is at most $C\|u_t\|_\infty$. The minimum of the two class probabilities is Lipschitz and remains at least $\beta_0/2$. Applying the quotient bound to numerator and denominator yields part~(ii). Combining with the spectral estimate proves part~(iii).
\end{proof}

\subsection{Formal Comparison with Existing Complexity Measures}
\label{app:comparison}
%=======================================================================

We formally compare $\psieff$ with three existing notions used in the learning
theory literature: Gaussian surface area, noise sensitivity, and margin-based
VC complexity.

\subsubsection{Gaussian surface area}

\begin{definition}[Gaussian surface area, \citealt{ledoux2001concentration}]
For a measurable set $A\subset\R^d$ with $C^1$ boundary:
\[
  \gamma^+(A):=\liminf_{\varepsilon\to 0}\frac{\mathcal{N}(A_\varepsilon\setminus A)}{\varepsilon},
\]
where $A_\varepsilon:=\{x:\operatorname{dist}(x,A)\le\varepsilon\}$ is the $\varepsilon$-enlargement
and $\mathcal{N}=\mathcal{N}(0,I_d)$.
\end{definition}

\begin{proposition}[Exact equivalence for balanced Gaussian case]
\label{prop:gsa_equiv}
For $P=\mathcal{N}(0,I_d)$, a compact $C^2$ boundary, and $P(A)=1/2$:
\[
  \psieff(\1_A,P)=4\gamma^+(A)=4\int_{\partial A}\gauss_d\,d\mathcal{H}^{d-1}.
\]
\end{proposition}

\begin{proof}
By Definition~\eqref{eq:def} and $P(A)=1/2$:
$\psieff(\1_A,P)=\mup_P(\1_A)/(1/2)=2\mup_P(\1_A)$.
Since $\mup_P(\1_A)=\lim_{\varepsilon\to 0}P(\operatorname{dist}(X,\partial A)\le\varepsilon)/\varepsilon
=2\int_{\partial A}\gauss_d\,d\mathcal{H}^{d-1}$:
$\psieff=4\int_{\partial A}\gauss_d\,d\mathcal{H}^{d-1}=4\gamma^+(A)$.
\end{proof}

\begin{remark}[Differences from Gaussian surface area]
Three key properties distinguish $\psieff$ from Gaussian surface area:
\begin{enumerate}[itemsep=2pt,leftmargin=*]
\item \textbf{Generality}: $\psieff(f,P)$ is defined for any measure $P$
  with an integrable density trace on the decision boundary.
  This enables the transformation formula Theorem~\ref{thm:pushforward}
  and the case-study analysis under $Q_\omega=(T_\omega)_\#P$.

\item \textbf{Class imbalance}: the normalization by $\min(P(f=1),P(f=0))$
  makes $\psieff$ sensitive to class imbalance, which is absent in $\gamma^+(A)$.
  For a halfspace $f_a$ with $a\ne 0$: $\gamma^+(A_a)=\gauss(a)$ (a single number)
  while $\psieff(f_a,P)=2\gauss(a)/\Phi(-|a|)\to\infty$ as $|a|\to\infty$,
  measuring the relative boundary concentration of increasingly imbalanced cuts.

\item \textbf{Transformation law}: Theorem~\ref{thm:pushforward} gives
  $\mup_{T_\#P}(g)=2\int_{\partial A}p|DT^{-\top}n|d\mathcal{H}^{d-1}$,
  an exact formula for how $\psieff$ changes under any $C^1$ diffeomorphism.
  For general non-isometric transformations, $\psieff$ follows the pushed data distribution through the displayed normal factor.
\end{enumerate}
\end{remark}

\subsubsection{Noise sensitivity}

\begin{definition}[Noise sensitivity]
\label{def:ns}
For $\delta\in(0,1)$ and $X_\delta:=\sqrt{1-\delta}X+\sqrt{\delta}U$ with
$U\sim\mathcal{N}(0,I_d)$ independent of $X\sim\mathcal{N}(0,I_d)$:
\[
  \mathrm{NS}_\delta(f):=\Prob(f(X)\ne f(X_\delta)).
\]
(See e.g.\ \citealt{ledoux2001concentration} for the Gaussian setting.)
\end{definition}

\begin{proposition}[$\psieff$ as infinitesimal noise sensitivity]
\label{prop:ns_equiv}
For $P=\mathcal N(0,I_d)$ and $f=\1_A$ with compact $C^2$ boundary and $P(A)=1/2$:
\[
  \psieff(f,P)=2\sqrt{2\pi}\lim_{\delta\to 0}\frac{\mathrm{NS}_\delta(f)}{\sqrt{\delta}}.
\]
\end{proposition}

\begin{proof}
The Ornstein--Uhlenbeck semigroup $\mathsf P_t f(x):=\E[f(e^{-t}x+\sqrt{1-e^{-2t}}U)]$, $t\ge0$,
satisfies $\mathsf P_t\1_A(x)=\Prob(f(X_\delta)=1\mid X=x)$ for $\delta=1-e^{-2t}$.
Gaussian stationarity and reversibility give:
\begin{align*}
  \mathrm{NS}_\delta(f)=\E[\1_A(X)\ne\1_A(X_\delta)]
  &=2\E[\1_A(X)(1-\mathsf P_t\1_A(X))]
  \\
  &=2\E[\1_A(X)]-2\E[\1_A(X)\mathsf P_t\1_A(X)].
\end{align*}
For $C^2$ boundary and small $\delta$, a Taylor expansion of the transition kernel
(by the tubular neighborhood theorem applied to the boundary-tube probability) gives:
\[
  \mathrm{NS}_\delta(f)=\sqrt{2\delta/\pi}\int_{\partial A}\gauss_d\,d\mathcal{H}^{d-1}+O(\delta)
  =\sqrt{2\delta/\pi}\cdot\frac{\gamma^+(A)}{1}+O(\delta).
\]
Since $\psieff(f,P)=4\gamma^+(A)$ (Proposition~\ref{prop:gsa_equiv}):
$\mathrm{NS}_\delta(f)=\sqrt{2\delta/\pi}\cdot\psieff(f,P)/4+O(\delta)$,
giving $\lim_{\delta\to 0}\mathrm{NS}_\delta/\sqrt{\delta}=\psieff/(2\sqrt{2\pi})$,
equivalently $\psieff=\sqrt{2\pi}\cdot 2\cdot\lim_{\delta\to 0}\mathrm{NS}_\delta/\sqrt{\delta}$.
\end{proof}

\begin{remark}
This shows that $\psieff$ is the leading-order coefficient of noise sensitivity
as a function of $\sqrt{\delta}$.  In this sense, $\psieff$ is the \textbf{infinitesimal
noise sensitivity constant}: small $\psieff$ means the classifier is robust to small
Gaussian perturbations.
\end{remark}

\subsubsection{Margin complexity}

Margin-based generalization bounds express classification error through normalized score separation and function-class norm control~\citep{bartlett2017spectrally}. For a balanced Gaussian halfspace, $\psieff=2\sqrt{2/\pi}$ is independent of dimension. Learning over the affine-halfspace family under fixed symmetric noise uses its VC dimension $d+1$ and needs $\widetilde O(d/\varepsilon)$ samples for clean disagreement $\varepsilon$ (Theorem~\ref{thm:halfspace_tight}). The local-vote bound in Theorem~\ref{thm:upper_bound} expresses the performance of its specified neighborhood rule through tube mass and sample coverage. These formulas describe how the hypothesis family and learning procedure enter the statistical guarantee.

\subsubsection{Comparison with other regularization notions}
\label{app:reg_comparison}

We compare $\psieff$-regularization with three common regularizers: spectral
normalization, Fisher information regularization, and margin regularization.
These comparisons identify the distinct geometric information captured by $\psieff$.

\paragraph{Spectral normalization \citep{bartlett2017spectrally}.}
Let $W\in\R^{m\times d}$ be a weight matrix.
Spectral normalization controls $\|W\|_{\mathrm{op}}:=\sigma_{\max}(W)$, where $\sigma_{\max}$ is the largest singular value; $\sigma_{\min}$ denotes the smallest.
For linear representations $T_W(x)=Wx$ with $W\in GL(d)$ and balanced classes:
\begin{equation}
\label{eq:spectral_psi}
  \psieff(f\circ T_W^{-1},(T_W)_\#P)
  =4\int_{\partial A}p(x)\frac{|W^{-\top}n(x)|}{1}\,d\mathcal{H}^{d-1}(x).
\end{equation}
The integrand $|W^{-\top}n|$ depends on the inverse transpose $W^{-\top}$.
For the isotropic case $W=\lambda I$: $|W^{-\top}n|=\lambda^{-1}|n|=\lambda^{-1}$,
giving $\psieff\propto\lambda^{-1}=\|W\|_{\mathrm{op}}^{-1}$.  In this case
$\psieff$ and spectral norm are equivalent (up to constants).

However, for non-isotropic $W$: the formula~\eqref{eq:spectral_psi} shows that
$\psieff$ depends on $W$ through the boundary-normal directions $\{n(x):x\in\partial A\}$,
which are classifier-specific.  In particular:
\begin{itemize}[itemsep=2pt,leftmargin=*]
\item Spectral normalization controls $\|W\|_{\mathrm{op}}$ globally, regardless of
  where the decision boundary lies.
\item $\psieff$ penalizes the effective boundary complexity in the
  \textbf{transformed space}, weighted by the density $p$ at the boundary.
  If the boundary is in a low-density region, $\psieff$ may be small even
  for large $\|W\|_{\mathrm{op}}$.
\end{itemize}
Thus $\psieff$-regularization is more \textbf{boundary-aware} than spectral normalization.

\paragraph{Margin regularization.}
For $h_{w,b}$ under $P=\mathcal N(0,I_d)$, define the boundary offset from the origin by $a_{\rm off}:=|b|/\|w\|$. The signed sample margin is $s(x)=(w^\top x+b)/\|w\|$. Proposition~\ref{prop:margin} gives
\begin{equation}\label{eq:margin_psi}
\psieff(h_{w,b},P)=\mathfrak g(a_{\rm off})\sim2a_{\rm off}
\qquad(a_{\rm off}\to\infty).
\end{equation}
Thus increasing the offset increases class imbalance and effective boundary complexity. The sample margins describe individual distances to the fitted boundary, while $a_{\rm off}$ locates that boundary under the Gaussian input law. The density-weighted surface integral extends the coefficient to curved decision boundaries.

\paragraph{Fisher information regularization.}
For model parameters $\theta$ and conditional label model $p_\theta(y\mid x)$, define the population Fisher matrix $F(\theta):=\E_{X\sim P,\,Y\sim p_\theta(\cdot\mid X)}[\nabla_\theta\log p_\theta(Y\mid X)(\nabla_\theta\log p_\theta(Y\mid X))^\top]$. For logistic models with $\theta=(w,b)$ and augmented vector $\tilde X=(X,1)$, $F(w,b)=\E[\sigma(w^\top X+b)(1-\sigma(w^\top X+b))\tilde X\tilde X^\top]$, the population cross-entropy Hessian.

The key difference:
\begin{itemize}[itemsep=2pt,leftmargin=*]
\item Fisher regularization penalizes uncertainty in \textbf{parameter space}.
\item $\psieff$ penalizes the \textbf{geometric complexity} of the decision boundary
  in \textbf{feature space}.
\end{itemize}
Two models with the same Fisher matrix can have different $\psieff$ through differences in boundary curvature or density weighting. The two diagnostics therefore capture complementary information.

\begin{remark}[$\psieff$-regularization uniqueness]
None of the three regularizers above---spectral normalization, margin, or Fisher---
explicitly minimizes the representation-dependent geometric complexity
$\psieff(f\circ T^{-1},T_\#P)$ under an invertible representation $T$.  The surrogate $\hat\psi_{\tau,\rho}$
(Section~\ref{sec:surrogate}) directly targets this quantity and,
by Proposition~\ref{prop:bridge}, targets the corresponding feature-space
halfspace complexity in the linear-head limit.
The fixed-backbone low-shot intervention summary
(Table~\ref{tab:intervention_summary}) shows that $\psieff$ intervention
consistently lowers feature-space $\hat\psieff$ (roughly $16\%$--$36\%$ versus
standard across settings) while keeping Top-1 close to standard and weight
decay.  This pattern is consistent with $\psieff$ intervention providing a
more targeted control of feature-space geometry than isotropic parameter
shrinkage.
\end{remark}

%=======================================================================
\subsection{Geometric Bias and Statistical Learning Rates}\label{app:bilateral_remarks}
The local-vote calculation separates a geometric contribution, $\alpha\psieff r+Lr^2$, from the probability of obtaining enough nearby labels. With uniform small-ball mass $a r^d$ and $r_n=(C\log n/n)^{1/d}$, the leading term is $\widetilde O(\psieff n^{-1/d})$. Its sufficient sample-size contribution is $\widetilde O((\psieff/\varepsilon)^d)$ with fixed balance and coverage constants.

Classical smooth-boundary minimax theory specifies a decision-set class through smoothness and entropy, together with density and noise assumptions~\citep{mammen1999smooth,tsybakov2009introduction}. The bounds in this appendix give their own coverage, packing, and KL quantities explicitly. A geometric transformation changes the tube coefficient, while the learning protocol and hypothesis family determine how that change enters a label-budget guarantee.

For the oscillatory feature family with known $T_\omega$, both $x\mapsto\1\{w^\top T_\omega(x)\ge0\}$ and $z\mapsto\1\{w^\top z\ge0\}$ have VC dimension at most $d+1$. Theorem~\ref{thm:halfspace_tight} applies to either parameterization under fixed symmetric label noise. The geometric comparison in Theorem~\ref{thm:family_reduction} describes the change in boundary mass under this explicit feature map.

\subsection{Proof of Corollary~\ref{cor:psieff_min} and Extended Gradient Flow Discussion}
\label{app:psieff_min}
%=======================================================================

\subsubsection{Proof of Corollary~\ref{cor:psieff_min}}

\begin{proof}
From Proposition~\ref{prop:margin}: for any halfspace $h_{w,b}$ under $P=\mathcal{N}(0,I_d)$,
$\psieff(h_{w,b},P)=\mathfrak g(b/\|w\|)$ with the Gaussian profile $\mathfrak g$ from Proposition~\ref{prop:margin}.
Since $\mathfrak g(a)>\mathfrak g(0)=2\sqrt{2/\pi}$ for all $a\ne 0$ (strict monotonicity,
Proposition~\ref{prop:margin}), the balanced halfspace $h_{w,0}$ with $b=0$
attains the minimum value $2\sqrt{2/\pi}$ for every nonzero normal $w$.
The Gaussian isoperimetric theorem \citep{borell1975brunn,sudakov1978extremal}
states that halfspaces achieve the minimum Gaussian surface area among all sets
with prescribed Gaussian measure.  For $P(A)=1/2$, the minimum is $\gamma^+(B_{0})$
where $B_0=\{x:x_1\ge 0\}$.  By Proposition~\ref{prop:gsa_equiv}:
$\psieff(h_{w,0},P)=4\gamma^+(B_0)=2\sqrt{2/\pi}$, consistent.
\end{proof}

\subsubsection{Population optimization and the boundary diagnostic}
The positive-definite Hessian establishes strict convexity, the lower bound $\mathcal L(w,b)\ge\eta a_d\|(w,b)\|$ establishes coercivity, and Gaussian conditional independence aligns the optimizer with the target direction. For a balanced target, symmetry fixes the optimal intercept at zero. These facts give the unique finite population optimizer and the geometric minimum.

\begin{proposition}[Convergence of boundary complexity for a balanced target]\label{prop:gf_final}
Under Theorem~\ref{thm:gf_psieff}, the population logistic trajectory has a finite time $T_0$ after which its boundary complexity is well defined and
\[
0\le\psieff(h_{w_t,b_t},P)-2\sqrt{2/\pi}
\le C e^{-c(t-T_0)}.
\]
\end{proposition}
\begin{proof}
Apply the convergence and differential inequality established in Appendix~\ref{app:gf_psieff}. The one-sided linear expansion $\mathfrak g(b)=\mathfrak g(0)+(4/\pi)|b|+O(b^2)$ gives the required local control at the balanced minimum.
\end{proof}

\subsection{Proofs for Section~\ref{sec:surrogate} (Surrogate and Bridge)}
\label{app:surrogate}
\label{app:bridge}
%=======================================================================

\begin{proof}[Complete proof of Proposition~\ref{prop:surrogate}]

\textbf{Part~(i): Smoothness.}
Define $F_i(w,b):=\sigma_\tau(\varepsilon_0-|s_i(w,b)|)$ where
$s_i(w,b)=(w^\top Z_i+b)/\|w\|$.
The map $w\mapsto\|w\|^{-1}$ is $C^\infty$ on $\R^m\setminus\{0\}$;
$w\mapsto w^\top Z_i$ is linear, hence $C^\infty$; composition gives $s_i\in C^\infty$.
On any open region where $s_i\ne0$, the absolute value is $C^\infty$, and since $\sigma_\tau$ is analytic, $F_i$ is $C^\infty$ there. At $s_i=0$, the cusp in $|s_i|$ gives local Lipschitz continuity and almost-everywhere differentiability globally. Likewise,
$\hat u_\tau=N^{-1}\sum_i\sigma_\tau(s_i)$ is $C^\infty$.  For every
$\rho>0$, the balance function
\[
  m_\rho(u)=\frac{1-\sqrt{(2u-1)^2+\rho^2}+\rho}{2}
\]
is analytic and strictly positive on $[0,1]$. Therefore
$\hat\psi_{\tau,\rho}$ is $C^\infty$ on the stated open regions. Replacing $|s_i|$ by $(s_i^2+\nu^2)^{1/2}$ gives a globally $C^\infty$ surrogate for fixed $\nu>0$.

\textbf{Part~(ii): Hard-limit consistency.}
For fixed $\{Z_i\}$ and parameters $(w,b)$ away from the threshold events,
$\sigma_\tau(u)\to\1\{u>0\}$ for $u\neq 0$ as $\tau\to 0$.  Hence
\[
  \widehat B_{\varepsilon_0,\tau}\to\frac{1}{N}\sum_i\1\{|s_i|\le\varepsilon_0\},
  \qquad
  \hat u_\tau\to \hat u:=\frac{1}{N}\sum_i\1\{s_i\ge0\}.
\]
Because $m_\rho(u)\to\min(u,1-u)$ uniformly on $[0,1]$ as $\rho\to 0$,
we obtain $\hat\psi_{\tau,\rho}\to\widehat{\psi}_{\mathrm{eff},\varepsilon_0}^{\mathrm{hard}}$.

\textbf{Part~(iii): Population consistency.}
For fixed positive $\tau,\rho,\varepsilon_0$, the law of large numbers gives
\[
\widehat B_{\varepsilon_0,\tau}\xrightarrow{P}\E_Q\sigma_\tau(\varepsilon_0-|s|),\qquad
\hat u_\tau\xrightarrow{P}\E_Q\sigma_\tau(s).
\]
The smoothed balance is positive, so continuous mapping gives convergence to the ratio of these population quantities. Absolute continuity of $Q$ assigns probability zero to the three hyperplanes $s=0,\pm\varepsilon_0$. Dominated convergence as $\tau\to0$, followed by uniform convergence of $m_\rho$ as $\rho\to0$, yields
\[
\frac{Q(|s|\le\varepsilon_0)}{\varepsilon_0\min(Q(h=1),Q(h=0))}.
\]
When the slice integral $r_s(t)$ is finite and continuous at zero,
$Q(|s|\le\varepsilon_0)=2\varepsilon_0 r_s(0)+o(\varepsilon_0)$ and
$2r_s(0)=2\int_{\partial h}q\,d\mathcal H^{m-1}=\mup_Q(h)$.
Sending $\varepsilon_0\to0$ therefore gives $\psieff(h,Q)$.
\end{proof}

\begin{proof}[Complete proof of Proposition~\ref{prop:bridge}]

\textbf{Part~(i): Definition of the target.}
For a fixed backbone $\phi$, the linear head acts on features
$Z=\phi(X)\sim Q$.  The target of interest is therefore the feature-space
halfspace complexity
\[
  G(w,b;Q):=\psieff(h_{w,b},Q),
  \qquad
  h_{w,b}(z)=\1\{w^\top z+b\ge 0\}.
\]

\textbf{Part~(ii): Surrogate consistency in feature space.}
Apply Proposition~\ref{prop:surrogate} directly to the feature sample
$Z_i=\phi(X_i)$.  This yields
\[
  \hat\psi_{\tau,\rho}(w,b;\{Z_i\}) \xrightarrow{\,P\,} G(w,b;Q)
\]
as $N\to\infty$, then $\tau,\rho\to 0$, and then $\varepsilon_0\to 0$.

\textbf{Part~(iii): Balanced zero-bias specialization.}
If $b=0$ and $Q(h_{w,0}=1)=Q(h_{w,0}=0)=1/2$, then Definition~\eqref{eq:def}
gives
\[
  G(w,0;Q)
  =\frac{2\int_{\{w^\top z=0\}} q(z)\,d\mathcal{H}^{m-1}(z)}{1/2}
  =4\int_{\{w^\top z=0\}} q(z)\,d\mathcal{H}^{m-1}(z).
\]
\end{proof}

%=======================================================================
\subsection[Multi-class psieff: Complete Proofs and Generalization Bounds]{Multi-class $\psieff$: Complete Proofs and Generalization Bounds}
\label{app:multiclass}
%=======================================================================

We develop $\psieff$ for $K$-class classification:
strict definitions, exact pushforward, upper and lower bounds.

\subsubsection{Strict definition and boundary structure}

\begin{definition}[Multi-class $\psieff$, extended]
\label{def:mc_extended}
Let $P$ have locally bounded density $p$ and
$f:\R^d\to[K]$ have decision regions $A_k:=f^{-1}(k)$,
each with $C^1$ boundary and $0<P(A_k)<1$. Write $f_k=\1_{A_k}$.
\begin{enumerate}[label=(\roman*),leftmargin=*,itemsep=2pt]
\item \textbf{Per-class (OvR):}
\begin{equation}
\label{eq:mc_per_class}
  \psieff^{(k)}(f,P):=\frac{2\int_{\partial A_k}p\,d\Hd}{\min(P(A_k),1-P(A_k))}.
\end{equation}

\item \textbf{Total complexity:}
\begin{equation}
\label{eq:mc_total}
  \psieff(f,P):=\sum_{k=1}^K\psieff^{(k)}(f,P).
\end{equation}

\item \textbf{Pair-based:}
Let $\Gamma_{jk}:=\partial A_j\cap\partial A_k$ (shared boundary, $j\ne k$).
Then $\partial A_k=\bigcup_{j\ne k}\Gamma_{jk}$ ($\Hd$-a.e.), so
\begin{equation}
\label{eq:mc_pair}
  \mup_P(f_k)=2\sum_{j\ne k}\int_{\Gamma_{jk}}p\,d\Hd.
\end{equation}
\end{enumerate}
For balanced classes $P(A_k)=1/K$: $\psieff(f,P)=K\sum_{k=1}^K\mup_P(f_k)$.
\end{definition}

\begin{remark}[OvR vs pair-based]
The OvR total counts each $\Gamma_{jk}$ twice (for class $j$ and class $k$),
weighted by minority-class probabilities.
For $K=2$ and $A=A_1$: $\psieff(f,P)=\psieff^{(1)}(f,P)+\psieff^{(2)}(f,P)=2\psieff(\1_A,P)$ (factor $2$ from counting both classes).
\end{remark}

\subsubsection{Complete proof of Proposition~\ref{prop:mc_pushforward}}

\begin{proof}[Complete proof]

\textbf{Step~1: Per-class pushforward.}
Fix $k\in[K]$.  Apply Theorem~\ref{thm:pushforward} to $f_k=\1_{A_k}$:
\begin{equation}
\label{eq:mc_pf_k}
  \mup_{T_\#P}(\1_{T(A_k)})
  =2\int_{\partial A_k}p(x)\,|(DT^{-\top}(x))n_k(x)|\,d\Hd(x).
\end{equation}

\textbf{Step~2: Class probability.}
$(T_\#P)(T(A_k))=P(A_k)$, so class priors are preserved.

\textbf{Step~3: Per-class transformed complexity.}
\begin{equation}
\label{eq:mc_per_class_pf}
  \psieff^{(k)}(f\circ T^{-1},T_\#P)
  =\frac{2\int_{\partial A_k}p|(DT^{-\top})n_k|\,d\Hd}{\min(P(A_k),1-P(A_k))}.
\end{equation}

\textbf{Step~4: Bound by $C_T$.}
$C_T:=\sup_{k,x\in\partial A_k}|(DT^{-\top}(x))n_k(x)|$.
From~\eqref{eq:mc_per_class_pf}:
$\psieff^{(k)}(f\circ T^{-1},T_\#P)\le C_T\cdot\psieff^{(k)}(f,P)$.

\textbf{Step~5: Sum over classes.}
$\psieff(f\circ T^{-1},T_\#P)=\sum_k\psieff^{(k)}(f\circ T^{-1},T_\#P)\le C_T\psieff(f,P)$.
\end{proof}

\begin{remark}[Tightness of $C_T$]
For linear $T_W(x)=Wx$, $W\in GL(d)$:
$C_T\le\|W^{-\top}\|_{\rm op}=\sigma_{\min}(W)^{-1}$.
Equality holds when $n_k$ aligns with the right singular vector of $W$ associated with its smallest singular value.
For orthogonal $W$: $C_T=1$ (isometry preserves $\psieff$).
\end{remark}

\subsubsection{Proof of Proposition~\ref{prop:mc_sample}}\label{app:mc_sample_proof}
\begin{proof}
Train binary local-majority rules $\hat f_k$ and predict a class attaining $\max_k\hat f_k$, with a fixed tie rule. When all binary predictions are correct, exactly the true class is positive. Thus
\begin{equation}\label{eq:mc_union}
P(\hat f\ne f)\le\sum_kP(\hat f_k\ne f_k).
\end{equation}
For each $k$, the correct-label probability in a ball contained in one classwise decision region is at least $1-\eta$. The binomial-Hoeffding argument gives
\begin{equation}\label{eq:mc_per_class_ub}
\E[P(\hat f_k\ne f_k)]\le\alpha_k\psi_k r+L_kr^2
+\int e^{-c_\eta nP(B(x,r))}\,dP(x).
\end{equation}
Sum this expression over classes. Since $\alpha_k\le1/2$, the leading term is at most $\psieff(f,P)r/2$. For standard symmetric $K$-class corruption of total rate $\eta<1/2$, the OvR conditional error is $\eta$ on class $k$ and $\eta/(K-1)$ on its complement, so the assumed common upper bound applies.
\end{proof}

\subsubsection{A multiclass packing lower bound}
\begin{theorem}[Multiclass information lower bound]\label{thm:mc_lb}
Let $\mathcal F$ be a family of $K$-class partitions under a fixed distribution $P$, and suppose it contains $M\ge2$ hypotheses satisfying
$\Delta\le P(f_j\ne f_\ell)\le D$ for distinct $j,\ell$. Labels equal the clean class with probability $a_\eta=1-\eta$ and each other class with probability $b_{\eta,K}=\eta/(K-1)$, independently over observations, where $0<\eta<(K-1)/K$. Then
\[
\inf_{\hat f}\sup_{f\in\mathcal F}\E[P(\hat f\ne f)]
\ge\frac\Delta2\left(1-\frac{nD_{\eta,K}D+\log2}{\log M}\right)_+,
\quad D_{\eta,K}=(a_\eta-b_{\eta,K})\log(a_\eta/b_{\eta,K}).
\]
This applies to any specified complexity-bounded family containing the stated packing.
\end{theorem}
\begin{proof}
At a point where two clean hypotheses agree, their categorical observation laws coincide. At a disagreement point the two laws exchange the entries $a_\eta$ and $b_{\eta,K}$, so their KL divergence is
$a_\eta\log(a_\eta/b_{\eta,K})+b_{\eta,K}\log(b_{\eta,K}/a_\eta)=(a_\eta-b_{\eta,K})\log(a_\eta/b_{\eta,K})$.
Integration under the original distribution gives the exact one-sample identity
\[
\KL(\mathbb P_{f_j}\Vert\mathbb P_{f_\ell})
=D_{\eta,K}P(f_j\ne f_\ell)\le D_{\eta,K}D.
\]
Independence gives the factor $n$. Apply Fano's inequality and nearest-hypothesis decoding in the full-distribution disagreement metric, exactly as in Lemma~\ref{lem:fano}. This proves the result with the specified packing cardinality, separation, and information bound.
\end{proof}
\begin{remark}[Partitions and packing]
The per-class normalization and pushforward laws describe each partition geometrically. A statistical lower bound additionally specifies a family of distinct partitions through $M$, $\Delta$, and $D$, all measured under the same $P$.
\end{remark}

\subsubsection{Multi-class surrogate convergence}

\begin{proposition}[Multi-class surrogate convergence]
\label{prop:mc_surrogate}
For $K$-class OvR with linear heads $(w_k,b_k)_{k=1}^K$:
(i)~$\hat\psi_{\tau,\rho}:=\sum_k\hat\psi_{\tau,\rho}^{(k)}$ is $C^\infty$ on open regions where every per-class signed distance is nonzero, and is differentiable almost everywhere globally;
(ii)~$\hat\psi_{\tau,\rho}$ converges in probability to the summed per-class
feature-space complexity as $N\to\infty$, then $\tau,\rho\to 0$, and then $\varepsilon_0\to 0$.
\end{proposition}

\begin{proof}
(i): Finite sum of the piecewise smooth functions from Proposition~\ref{prop:surrogate}(i).
(ii): Apply Proposition~\ref{prop:surrogate}(iii) per class and sum the limits.
\end{proof}

%=======================================================================
\subsection[Monitoring During CNN Training]{Monitoring During CNN Training}\label{app:exp5}
\begin{figure}[!htbp]
\centering
\includegraphics[width=0.65\textwidth]{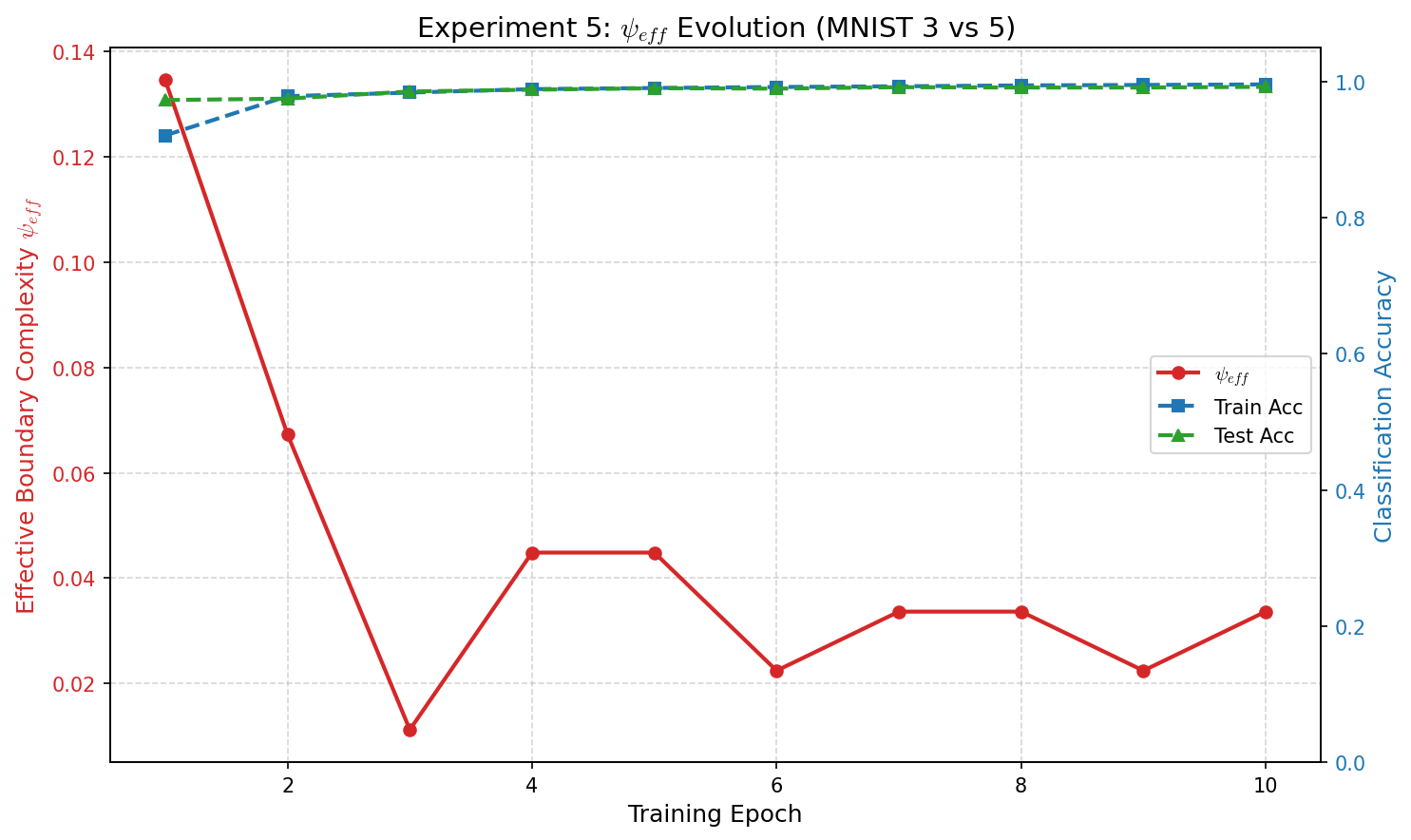}
\caption{CNN training monitoring on MNIST digits 3 and 5. The curves display the feature-space boundary readout and test accuracy over training, illustrating joint changes in representation geometry and predictive performance.}
\label{fig:exp5}
\end{figure}
\paragraph{Architecture and monitoring protocol.}
The network has two convolutional layers with 32 and 64 filters, ReLU activations, max pooling, a 128-unit fully connected layer, and a binary head. Training uses Adam with learning rate $10^{-3}$ and batch size 64. At each training stage, a fixed set of 512 test images supplies a supplementary boundary readout; test accuracy summarizes predictive performance. These measurements monitor a trained network. The prospective backbone-selection protocol computes its diagnostics on independent held-out training features.
\paragraph{Geometric reading.}
The plotted training stages associate increasing accuracy with lower boundary readouts. A fitted head at each training stage defines the decision surface and its feature-space normal. This viewpoint follows the fixed-probe target in Proposition~\ref{prop:bridge}, with the representation and head updated between training stages.

\subsection{Supplementary End-to-End Controllability Check}\label{app:ext_endtoend}
\label{app:d1}
%=======================================================================

\paragraph{Architecture and data.} The main end-to-end check uses a CIFAR-style
ResNet-18: a $3\times3$ stride-1 input convolution followed by
residual stages with a 512-dimensional penultimate representation, and a dataset-specific linear classifier.
CIFAR-100 and CIFAR-10 use the standard 50K/10K train/test split with random
crop, horizontal flip, and per-channel normalization.

\paragraph{Optimization.}
SGD with momentum $0.9$, 40 epochs, batch size 256, and multistep learning-rate
decay. Standard training uses learning rate $3\times10^{-3}$; the
boundary-pressure run uses $3.5\times10^{-3}$; the weight-decay baseline
uses SGD weight decay $5\times10^{-4}$. Margin regularization uses
$\lambda=0.1$ on a top-2 hinge margin, and entropy regularization uses a
$0.01$ confidence penalty. Each training seed starts all conditions from the
same random initialization; the independent training seeds are $\{42,43,44\}$.

\paragraph{Measured $\hat\psieff$.}
At evaluation time we estimate the one-vs-rest boundary mass of the dataset-specific
linear head on 2000 held-out test examples with tube width $\varepsilon_0=0.1$.
The table symbols $\hat\psi_{\rm hard}$ and $\hat\psi_{\rm kern}$ denote the class-averaged hard and triangular-kernel readouts with the observed-class normalization specified in Appendix~\ref{app:compute}. These test-set measurements monitor the training trajectory and final model. The backbone-selection diagnostics use the independent training-heldout split specified in Section~5.1.

\paragraph{Regularized loss.}
The end-to-end run uses a differentiable top-2 boundary-pressure proxy during
training,
\[
\mathcal L_{\rm reg}
=\mathcal L_{\rm CE}
+\lambda\left[0.5\,\mathrm{ReLU}\{0.2-(g_y-\max_{c\ne y}g_c)\}
+0.05\|W\|_F\right],
\]
with penalty weight $\lambda\in\{0.06,0.12,0.24\}$. Here $y$ is the true class, $g_c(z)=w_c^\top z+b_c$ its class-$c$ logit, $W$ stacks $w_c^\top$, and $\mathcal L_{\rm CE}$ is the minibatch cross-entropy; the bracketed expression is averaged over that minibatch. The optimization proxy supplies the training signal, and the reported $\hat\psieff$ is the held-out finite-tube estimate above. The proxy gives a
lightweight end-to-end check that boundary pressure can move the measured state.

\paragraph{Results.}
This end-to-end 40-epoch setting examines the boundary readout when the
representation itself is changing. Downstream probes freeze the epoch-40
representation and train new linear heads with balanced low-label subsets. The
three-seed average shows lower measured boundary mass and higher low-label probe
accuracy for the boundary-pressure run. Calibration varies across methods, while the boundary-mass and low-label results support controllability of the measured boundary state.

\begin{table}[!htbp]
\centering\scriptsize
\caption{CIFAR-100 ResNet-18 end-to-end controllability check. Entries are mean$\pm$standard deviation over three training seeds. Low-label columns are frozen linear-probe top-1 accuracies on the epoch-40 representations, averaged over three label-subsampling seeds inside each training seed.}
\label{tab:endtoend_extra_metrics}
\resizebox{\linewidth}{!}{%
\begin{tabular}{lcccccccc}
\toprule
Method & Test Acc.\upgood & $\hat\psi_{\rm hard}$\downgood & $\hat\psi_{\rm kern}$\downgood & ECE\downgood & $\|W\|_F$ & feat. norm & margin & 20/50-spc probe\upgood\\
\midrule
Standard & $53.68\pm0.25\%$ & $23.57\pm0.33$ & $11.88$ & $0.060$ & $10.38$ & $31.60$ & $0.322$ & $44.46/47.47$\\
Weight decay & $54.55\pm0.20\%$ & $23.61\pm0.23$ & $11.77$ & $0.055$ & $10.03$ & $30.21$ & $0.361$ & $45.53/48.43$\\
Margin reg. & $54.55\pm0.18\%$ & $23.82\pm0.52$ & $11.80$ & $0.046$ & $10.48$ & $31.42$ & $0.371$ & $44.96/47.96$\\
Entropy reg. & $53.98\pm0.13\%$ & $24.01\pm0.41$ & $12.05$ & $0.051$ & $10.34$ & $31.40$ & $0.304$ & $44.49/47.43$\\
\psimethod{$\psieff$-reg, $\lambda=0.24$} & \psimethod{$57.09\pm0.42\%$} & \psimethod{$22.07\pm0.35$} & \psimethod{$10.95$} & \psimethod{$0.048$} & \psimethod{$9.69$} & \psimethod{$32.48$} & \psimethod{$0.592$} & \psimethod{$48.26/51.07$}\\
\bottomrule
\end{tabular}
}
\end{table}

\begin{table}[!htbp]
\centering\scriptsize
\caption{ResNet-18 $\lambda$ sweep on CIFAR-100. The sweep shows a Pareto-safe region: larger boundary pressure lowers both hard and kernel tube estimates while preserving or improving test accuracy and low-label probes.}
\label{tab:lambda_sweep_resnet}
\begin{tabular}{lccccc}
\toprule
$\lambda$ & Test Acc.\upgood & $\hat\psi_{\rm hard}$\downgood & $\hat\psi_{\rm kern}$\downgood & ECE\downgood & 20/50-spc probe\upgood\\
\midrule
$0.06$ & $56.27\pm0.29\%$ & $23.56\pm0.19$ & $11.77$ & $0.069$ & $47.78/50.55$\\
$0.12$ & $57.05\pm0.27\%$ & $23.22\pm0.55$ & $11.59$ & $0.057$ & $48.09/50.63$\\
$0.24$ & $57.09\pm0.42\%$ & $22.07\pm0.35$ & $10.95$ & $0.048$ & $48.26/51.07$\\
\bottomrule
\end{tabular}
\end{table}

\begin{table}[!htbp]
\centering\scriptsize
\caption{CIFAR-10 ResNet-18 repeat. The same boundary-pressure run improves test accuracy, hard/kernel boundary estimates, expected calibration error (ECE), and low-label probes over standard and weight-decay training.}
\label{tab:cifar10_resnet_repeat}
\begin{tabular}{lccccc}
\toprule
Method & Test Acc.\upgood & $\hat\psi_{\rm hard}$\downgood & $\hat\psi_{\rm kern}$\downgood & ECE\downgood & 20/50-spc probe\upgood\\
\midrule
Standard & $84.36\pm0.17\%$ & $1.96\pm0.07$ & $1.00$ & $0.051$ & $81.78/82.32$\\
Weight decay & $85.08\pm0.23\%$ & $2.12\pm0.11$ & $1.03$ & $0.047$ & $82.09/83.10$\\
\psimethod{$\psieff$-reg} & \psimethod{$86.13\pm0.16\%$} & \psimethod{$1.54\pm0.07$} & \psimethod{$0.76$} & \psimethod{$0.042$} & \psimethod{$83.88/84.39$}\\
\bottomrule
\end{tabular}
\end{table}

\subsubsection{Interface-counted multi-class variant}
\label{app:mc_interface}
The interface-counted partition metric is defined once in Appendix~\ref{app:multi}.
The frozen-feature intervention applies the classwise Smooth surrogate. The end-to-end experiment optimizes the top-2 pressure proxy and reports class-averaged OvR boundary readouts. The interface-counted construction measures pairwise competition in a recovered multiclass partition. In the Tiny-ImageNet
5-way rerun, for class indices $a<b$ and gate temperature $\tau_{\rm gate}>0$, we tested the soft interface gate
\[
\gamma_{ab}(z)=\sigma\!\left(\frac{\min(g_a(z),g_b(z))-\max_{c\notin\{a,b\}}g_c(z)}{\tau_{\rm gate}}\right),
\quad
s_{ab}(z)=\frac{g_a(z)-g_b(z)}{\|w_a-w_b\|},
\]
and summed gated pairwise tube estimates over $a<b$. The resulting policy rows
are reported in Tables~\ref{tab:multiclass_stress_app}
and~\ref{tab:multiclass_policy_supplement_app}, alongside the OvR score and
median-gap variants. These rows are included as multi-class diagnostics for the
same cached-feature protocol used in the binary experiments.

\clearpage
\section*{NeurIPS Paper Checklist}

\begin{enumerate}

\item \textbf{Claims}

\textbf{Question:} Do the main claims made in the abstract and introduction accurately reflect the paper's contributions and scope?

\textbf{Answer:} [Yes]

\textbf{Justification:} The abstract and introduction present boundary mass as a diagnostic for fixed representations and linear heads. The claims reflect the pushforward law, finite-tube estimator, controlled interventions, multiclass estimator agreement, and acquisition-policy results.

\item \textbf{Limitations}

\textbf{Question:} Does the paper discuss the limitations of the work performed by the authors?

\textbf{Answer:} [Yes]

\textbf{Justification:} The theoretical statements specify regularity, invertibility, and class-balance assumptions. Appendix~\ref{app:compute} describes finite-width, smoothing, and sampling effects; Appendices~\ref{app:multi} and~\ref{app:acquisition} report the multiclass and acquisition settings, and Appendix~\ref{app:ext_surrogate} explains evaluation during adaptation.

\item \textbf{Theory assumptions and proofs}

\textbf{Question:} For each theoretical result, does the paper provide the full set of assumptions and a complete and correct proof?

\textbf{Answer:} [Yes]

\textbf{Justification:} The theoretical statements specify the required regularity, smoothness, balance, fixed-head, and representation-map assumptions. The main text gives the core tube and pushforward arguments, while the extended appendix provides complete derivations and marks supplementary extensions with their additional assumptions.

\item \textbf{Experimental result reproducibility}

\textbf{Question:} Does the paper fully disclose all the information needed to reproduce the main experimental results of the paper to the extent that it affects the main claims and/or conclusions of the paper, regardless of whether the code and data are provided?

\textbf{Answer:} [Yes]

\textbf{Justification:} The experimental sections disclose the datasets, class-pair counts, frozen backbones, diagnostic-heldout protocol, label-budget target, tube-width choice, comparison diagnostics, and statistical tests used for the main claims. The appendix reports estimator reliability, sensitivity and cost analyses, pilot-label policy details, transferability variants, negative controls, and end-to-end check hyperparameters. The public repository provides a README, experiment scripts, data-preparation instructions, and commands for computing the diagnostics, tables, and figures.

\item \textbf{Open access to data and code}

\textbf{Question:} Does the paper provide open access to the data and code, with sufficient instructions to faithfully reproduce the main experimental results, as described in supplemental material?

\textbf{Answer:} [Yes]

\textbf{Justification:} The experiments use public datasets and pretrained backbones. The repository at \url{https://github.com/qqaqq-code/boundary_mass} contains the estimator, experiment scripts, result-processing utilities, plotting code, and usage instructions. Public source links provide access to upstream assets, and the documented workflow produces the feature caches.

\item \textbf{Experimental setting/details}

\textbf{Question:} Does the paper specify all the training and test details necessary to understand the results, such as data splits, hyperparameters, how they were chosen, and optimizer type?

\textbf{Answer:} [Yes]

\textbf{Justification:} The paper specifies the synthetic transformations, CIFAR-100 and Tiny-ImageNet class-pair experiments, frozen backbones, train/diagnostic/test separation, linear-probe protocol, label-budget definition, and default estimator hyperparameters. The supplementary end-to-end section reports architecture, optimizer, learning-rate schedule, batch size, seeds, augmentation, and regularizer settings.

\item \textbf{Experiment statistical significance}

\textbf{Question:} Does the paper report error bars suitably and correctly defined, or other appropriate information about the statistical significance of the experiments?

\textbf{Answer:} [Yes]

\textbf{Justification:} The main and appendix tables report permutation-test $p$-values for incremental predictive gains and mean$\pm$standard deviation over training seeds for the end-to-end experiments. The appendix also reports split-half estimator reliability and variability across independently refitted pilot heads.

\item \textbf{Experiments compute resources}

\textbf{Question:} For each experiment, does the paper provide sufficient information on the computer resources needed to reproduce the experiments, such as compute-worker type, memory, and execution time?

\textbf{Answer:} [Yes]

\textbf{Justification:} The appendix reports surrogate-training costs from saved logs and analyzes the marginal time and memory cost of cached-feature estimation. The execution environment uses Python 3.10.19, PyTorch 2.1.0+cu118, four NVIDIA L20 46GB GPUs, dual Xeon Gold 6530 CPUs with 128 logical cores, and 755 GiB RAM.

\item \textbf{Code of ethics}

\textbf{Question:} Does the research conducted in the paper conform, in every respect, with the NeurIPS Code of Ethics?

\textbf{Answer:} [Yes]

\textbf{Justification:} The research uses public vision benchmarks, cached representations, and offline supervised evaluation. Its intended use is label-budget analysis, accompanied by accuracy, robustness, privacy, and fairness evaluation as appropriate to the application.

\item \textbf{Broader impacts}

\textbf{Question:} Does the paper discuss both potential positive societal impacts and negative societal impacts of the work performed?

\textbf{Answer:} [Yes]

\textbf{Justification:} The broader-impacts paragraph discusses annotation-effort allocation, the effects of overweighting an aggregate score, and joint reporting of annotation cost, classwise accuracy, calibration, and robustness.

\item \textbf{Safeguards}

\textbf{Question:} Does the paper describe safeguards that have been put in place for responsible release of data or models that have a high risk for misuse, such as pretrained language models, image generators, or scraped datasets?

\textbf{Answer:} [N/A]

\textbf{Justification:} The released assets consist of diagnostic code, experiment scripts, and illustrative figures for public vision benchmarks. Upstream datasets and pretrained models are accessed from their original providers under the listed terms.

\item \textbf{Licenses for existing assets}

\textbf{Question:} Are the creators or original owners of assets used in the paper, such as code, data, or models, properly credited, and are the license and terms of use explicitly mentioned and properly respected?

\textbf{Answer:} [Yes]

\textbf{Justification:} The paper credits datasets and pretrained backbones through names and citations. Table~\ref{tab:assets_app} lists these assets, their uses, and the original references. The code supplement documents how to obtain upstream assets from their original providers.

\item \textbf{New assets}

\textbf{Question:} Are new assets introduced in the paper well documented, and is the documentation provided alongside the assets?

\textbf{Answer:} [Yes]

\textbf{Justification:} The new assets are code for the boundary-mass estimator, experiment scripts and plotting utilities. Documentation includes a README, expected inputs, upstream sources, and usage commands.

\item \textbf{Crowdsourcing and research with human subjects}

\textbf{Question:} For crowdsourcing experiments and research with human subjects, does the paper include the full text of instructions given to participants and screenshots, if applicable, as well as details about compensation, if any?

\textbf{Answer:} [N/A]

\textbf{Justification:} All empirical analyses use existing public vision datasets and offline model features. The research protocol consists of computational experiments on these existing assets.

\item \textbf{Institutional review board approvals or equivalent for research with human subjects}

\textbf{Question:} Does the paper describe potential risks incurred by study participants, whether such risks were disclosed to the subjects, and whether Institutional Review Board approvals or an equivalent approval/review were obtained?

\textbf{Answer:} [N/A]

\textbf{Justification:} The study consists of offline computational analysis of existing public benchmark datasets and model features; its experimental protocol is a computational benchmark evaluation.

\item \textbf{Declaration of LLM usage}

\textbf{Question:} Does the paper describe the usage of LLMs if it is an important, original, or non-standard component of the core methods in this research?

\textbf{Answer:} [N/A]

\textbf{Justification:} The core research method consists of mathematical analysis, feature extraction, linear-probe fitting, and numerical diagnostics. LLM assistance supported manuscript revision and proof checking; the paper states the mathematical assumptions, derivations, and experimental protocols explicitly.

\end{enumerate}

\end{document}